\documentclass[5p,sort&compress]{elsarticle}
\usepackage{amsmath}
\usepackage{amssymb}
\usepackage{color}
\usepackage{makecell}
\usepackage{mathtools}
\usepackage{mathrsfs}
\usepackage{mfirstuc}
\usepackage{multirow}
\usepackage{subcaption}
\usepackage{tikz}
\usepackage{booktabs}

\renewcommand{\thesubparagraph}{\alph{subparagraph}}

\usepackage{titlesec}
\titleformat{\paragraph}[runin]{\normalfont\normalsize\itshape}{}{0em}{}[.]
\titlespacing*{\paragraph}{0pt}{3.25ex plus 1ex minus .2ex}{1em}

\titleformat{\subparagraph}[runin]{\normalfont\normalsize\itshape}{\normalfont\normalsize (\thesubparagraph)}{0.5em}{}
\titlespacing*{\subparagraph}{\parindent}{0ex plus 1ex minus .2ex}{1em}

\newcommand{\project}{\textbf{Nothing Stands Still}}
\newcommand{\abbrev}{\textbf{NSS}}
\newcommand{\onepc}{fragment}
\newcommand{\allpc}{scan}

\newcommand{\ie}[1]{{{i.e., }{#1}}}
\newcommand{\eg}[1]{{{e.g., }{#1}}}

\definecolor{ForestGreen}{RGB}{34,139,34}
\newcommand{\red}[1]{\textcolor{red}{#1}}
\newcommand{\green}[1]{\textcolor{ForestGreen}{#1}}

\definecolor{A_curve}{RGB}{148,103,189}
\definecolor{B_curve}{RGB}{100, 149, 237}
\definecolor{C_curve}{RGB}{255, 105, 97}
\definecolor{D_curve}{RGB}{166, 219, 160}
\definecolor{E_curve}{RGB}{188, 189, 34}
\definecolor{F_curve}{RGB}{255, 200, 0}

\DeclareMathOperator*{\argmin}{arg\,min}

\usepackage{lineno,hyperref}
\modulolinenumbers[5]

\journal{Journal of \LaTeX\ Templates}

\let\linenumbers\nolinenumbers\nolinenumbers

\begin{document}

\begin{frontmatter}

\title{\project{}: A Spatiotemporal Benchmark on 3D Point Cloud Registration Under Large Geometric and Temporal Change}

\author[Stanford]{Tao Sun}
\author[ETH]{Yan Hao}
\author[ETH]{Shengyu Huang}
\author[Stanford]{Silvio Savarese}
\author[ETH]{Konrad Schindler}
\author[ETH,MS]{Marc Pollefeys}
\author[Stanford]{Iro Armeni\corref{mycorrespondingauthor}}

\cortext[mycorrespondingauthor]{Corresponding author}
\ead{iarmeni@stanford.edu}

\address[Stanford]{Stanford University}
\address[ETH]{ETH Zurich}
\address[MS]{Microsoft Mixed Reality \& AI Lab, Zurich}

\begin{abstract}
    Building 3D geometric maps of man-made spaces is a well-established and active field that is fundamental to numerous computer vision and robotics applications. However, considering the continuously evolving nature of built environments, it is essential to question the capabilities of current mapping efforts in handling temporal changes. In addition to the above, the ability to create spatiotemporal maps holds significant potential for achieving sustainability and circularity goals. Existing mapping approaches focus on small changes, such as object relocation within common living spaces or self-driving car operation in outdoor spaces; all cases where the main structure of the scene remains fixed. Consequently, these approaches fail to address more radical change in the structure of the built environment, such as on the geometry and topology of it. To promote advancements on this front, we introduce the \textbf{\project{} (\abbrev)} benchmark, which focuses on the spatiotemporal registration of 3D scenes undergoing large spatial and temporal change, ultimately creating one coherent spatiotemporal map. Specifically, the benchmark involves registering within the same coordinate system two or more partial 3D point clouds (fragments) originating from the same scene but captured from different spatiotemporal views. In addition to the standard task of \textit{pairwise} registration, we assess \textit{multi-way} registration of multiple \onepc{s} that belong to the same indoor environment and any temporal stage. As part of \abbrev{}, we introduce a dataset of 3D point clouds recurrently captured in large-scale building indoor environments that are under construction or renovation. The \abbrev{} benchmark presents three scenarios of increasing difficulty, with the goal to quantify the generalization ability of point cloud registration methods over space (within one building and across buildings) and time.  We conduct extensive evaluations of state-of-the-art methods on \abbrev{} over all tasks and scenarios. The results demonstrate the necessity for novel methods specifically designed to handle large spatiotemporal changes. The homepage of our benchmark is at \url{http://nothing-stands-still.com}

\end{abstract}


\end{frontmatter}

\linenumbers

\section{Introduction}

“Everything flows, nothing stands still" -- as Heraclitus~\footnote{An ancient Greek philosopher, 501 B.C.} advocated, a critical property of the world around us is that it changes over time. The temporal dimension and its impact on the built environment have not been ignored by the field of computer vision, photogrammetry, and robotics. Its study appears in different tasks, such as those related to video understanding \cite{simonyan2014two,purushwalkam2020aligning,haresh2021learning,kwon2022context,deng2024vg4d}, self-driving cars~\cite{qi2021offboard,huang2022dynamic,zhang2023towards}, change detection in 2D images acquired at scene-level  ~\cite{sakurada2020weakly,lei2020hierarchical,ru2020multi,prabhakar2020cdnet++,wang2021transcd,li2022scene} or in satellite imagery~\cite{bourdis2011constrained,peng2019end,chen2020dasnet,zheng2021change,cheng2024change}, change detection in 3D scans~\cite{xiao2015street, qin20163d, kharroubi2022three,gehrung2022change,huang2022semantics,de2023siamese,stilla2023change,lopez20243d}, object relocalization in recaptured 3D indoor scenes~\cite{wald20193rscan,wald2020beyond,halber2019rescan,zhu2023living}, and robot navigation in dynamic environments~\cite{droeschel2017continuous, wang2020mobile, wang2021navigation}. However, the examined change, especially in the 3D indoor domain, often focuses on small spatial (\eg that of a room~\cite{straub2019replica,halber2019rescan, wald20193rscan, wald2020beyond}) and temporal (\eg that of a few minutes~\cite{droeschel2017continuous,prabhakar2020cdnet++,wang2021transcd,huang2022dynamic,kwon2022context,deng2024vg4d}) scales, and mainly limited to object relocation (movement of objects)~\cite{straub2019replica,halber2019rescan, wald20193rscan, wald2020beyond, li2022scene}.

The built environment undergoes various changes throughout its lifecycle, starting from construction, through operation, and finally reaching the end-of-life phase. These changes go beyond simple relocation and involve differences in geometry, appearance, and topology of the building elements. Examples of such changes include the installation of pipes on ceilings, the transformation of floors before and after carpeting, and the gradual development of walls from a group of studs to their final structure visible to users. Among the different lifecycle phases, the most significant differences can be observed during construction and before/after renovation. 

Understanding and addressing these dynamic changes opens up new research directions, shifting the predominant static perspective of scene understanding. An instance of these directions is evident in robotics. Robots are frequently required to localize themselves within pre-mapped 3D structures that may have experienced alterations over time. Proficiently identifying and adapting to these changes improves not only the robot's navigation but also its interaction with the environment~\cite{tsamis2021lifelongmapping, stefanini2023safe}. Moreover, acquiring a spatiotemporal understanding of how buildings evolve over time is crucial for achieving sustainability and circular economy goals in the built environment~\cite{munaro2020towards}. For instance, it enables quantitative monitoring and quality control of construction progress, leading to a reduction in out-of-estimate construction costs associated with rework. Currently, progress monitoring is often assessed in a rough manner by project managers, with rework accounting for 52\% of the total out-of-estimate costs~\cite{love2002using}. Furthermore, a spatiotemporal understanding of building changes can play a significant role in establishing workflows for material reuse. It is estimated that 95\% of non-hazardous construction and demolition waste is reusable or recyclable~\cite{ma2020challenges}. However, a large amount of this material ends up in landfills due to a lack of information about materials within buildings. This is due to raw materials getting hidden behind surfaces or paint as construction progresses without proper documentation.

To this end, we propose \textbf{\project{} (\abbrev)}, a novel spatiotemporal benchmark utilizing 3D point cloud captures of indoor environments in the aforementioned lifecycle phases. These captures encompass a large spatial and temporal scale and contain changes that extend beyond object relocation. As part of the benchmark, we introduce a spatiotemporal point cloud dataset comprising 6 large-scale building areas (\ie, distinct buildings or large sections of them referred to as \textit{areas}) in multiple temporal stages (referred to as \textit{stages}) spanning several months (Section~\ref{sec:dataset}). We focus on the problem of spatiotemporal point cloud registration and design a series of experiments to demonstrate the inherent challenges of this setup and highlight the limitations of existing methods in addressing them. Notably, spaces under construction are commonly of low-texture and highly repetitive geometry, posing challenges for local feature-based algorithms in computer vision. These algorithms struggle in the spatiotemporal registration task, where similar local features may not correspond to the same location over time.

To evaluate the generalization ability of methods across space and time, we define three scenarios that involve different spatial and temporal data splits (Section~\ref{sec:benchmark_tasks}). Unlike typical spatial registration setups, our training and testing process includes not only point cloud pairs from the same \textit{stage} and \textit{area}, but also pairs from different \textit{stages} and \textit{areas}. When referring to `pairs from different areas' in a data split, it means point cloud pairs originating from distinct buildings. For instance, one point cloud pair may come from \textit{area} A, while another may come from \textit{area} B. Within a pair, the two point clouds can represent either the same or different \textit{stages} of their respective \textit{area} of origin. Given the large-scale nature of each \textit{area}, the input pairs represent partial observations, namely \textit{\onepc{s}}, of the complete area. Consequently, the pairwise registration task is constrained to achieving local alignment between the input pair. To achieve global alignment in the context of entire \textit{areas}, we incorporate the task of multi-way registration that considers all input \textit{\onepc{s}} belonging to the same \textit{area} at any \textit{stage}.

We evaluate several state-of-the-art algorithms~\cite{rusu2009fast,bai2020d3feat,Choy2019FCGF,predator,qin2022geometric} on the \textbf{\abbrev{}} benchmark.
We also evaluate these algorithms on a state-of-the-art spatiotemporal 3D point cloud dataset \cite{wald2020beyond}, which captures changes in inhabited indoor scenes related to furniture addition, removal, or relocation within rooms. This comparative analysis showcases the need for more challenging setups when addressing the problem of understanding and operating in dynamic environments.

The contributions of this paper can be summarized as:
\begin{itemize}
\item We introduce a new spatiotemporal dataset that captures large spatial and temporal changes in the geometry, appearance, and topology of building elements. The dataset comprises 6 indoor areas undergoing construction and renovation with recurrent captures spaced months apart (2-6 per area).
\item We propose a novel benchmark, \textbf{\abbrev}, for spatiotemporal 3D point cloud registration, which includes both pairwise and multi-way registration. The evaluation employs diverse data splits, where training and testing pairs originate from across areas \textit{and} stages.
\item We provide extensive experimental analysis and insights into the performance of state-of-the-art registration algorithms on the \abbrev{} benchmark. We also provide evaluation results on a state-of-the-art spatiotemporal 3D point cloud dataset~\cite{wald2020beyond}, following the same evaluation protocol.
\end{itemize}

We also provide the community with a server for evaluating their algorithms on the test sets, which we keep hidden. A leaderboard showcases the latest results and progress on the benchmark. For more details, please visit \href{http://nothing-stands-still.com/}{nothing-stands-still.com}.
\section{Related Work}
We first review works on spatiotemporal reasoning from visual data and the employed datasets before proceeding to pairwise and multi-way registration methods. Finally, we briefly discuss synthetic point cloud generation since our point cloud registration benchmark was created from 3D mesh data of real-world captures.

\begin{figure*}
    \includegraphics[keepaspectratio,width=0.99\linewidth]{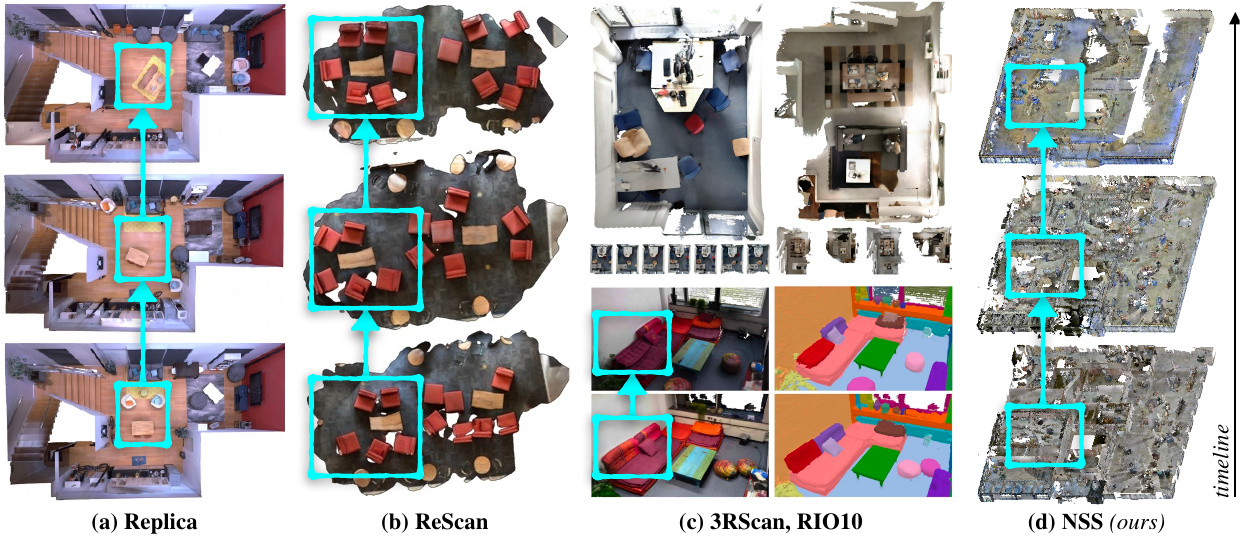}
    \caption{\textbf{Qualitative examples of existing indoor spatiotemporal datasets and \project{} (\abbrev{}).} As shown, existing datasets focus on small and daily changes in living environments, whereas \abbrev{} exhibits drastic changes over time. Examples of such changes over different \textit{stages} are highlighted with a \textcolor{cyan}{\textit{cyan box}}.}
    \label{fig:related_datasets}
\end{figure*}

\subsection{Spatiotemporal Reasoning}
Spatiotemporal reasoning from visual data is a fundamental problem in computer vision, photogrammetry, and robotics, and can be examined at various levels of detail. Change detection methods categorize scenes into stationary and changed regions, through 2D pixel~\cite{hussain2013change}, 3D point~\cite{yew2021city, de2023siamese}, or 3D voxel~\cite{pollard2007change,xiao2015street} classification tasks. Motion segmentation approaches~\cite{chen2021moving,accumulation} segment scenes into static and dynamic parts. Optical flow estimation techniques~\cite{horn1981determining,black1993framework,brox2009large} model fine-grained motion information by associating pixels across frames, typically formulated as optimization tasks \cite{black1993framework}. Modern learning-based methods~\cite{ilg2017flownet,hui2018liteflownet,fischer2015flownet,teed2020raft} directly learn flow prediction with enhanced accuracy and efficiency from large datasets. Scene flow estimation~\cite{vedula1999three,sun2010secrets} additionally provides depth information of objects in a 3D scene. Traditional methods~\cite{vogel20113d,vogel2013piecewise,vogel20153d} leverage motion smoothness priors within optimization frameworks, while learning-based methods~\cite{liu2019flownet3d,puy2020flot} learn directly from large-scale datasets.

Spatiotemporal reasoning encompasses various downstream tasks, including 3D change detection~\cite{kharroubi2022three,stilla2023change}, action recognition from videos\cite{simonyan2014two,8454294}, multi-object visual tracking~\cite{milan2016mot16,xiang2015learning}, dynamic reconstruction~\cite{rempe2020caspr,huang2022multiway}, and novel view synthesis~\cite{pumarola2021d,martin2021nerf}. Most tasks address spatiotemporal changes at video frame rate. Going beyond the temporal change induced in milliseconds, D$^{4}$R~\cite{golparvar2009d4ar} focuses on aligning images captured in construction sites on a daily basis for the purpose of construction progress monitoring. In a non-built setting, Dong et al. \cite{dong20174d} perform 4D reconstruction of agricultural crops for monitoring growth. Furthermore, works like~\cite{griffith2020transforming,schindler2010probabilistic,schindler2007inferring} aim to sequence online image collections spanning a year or decades based on visibility and temporal occupancy. Scene chronology~\cite{matzen2014scene} scales this sequencing operation to millions of photos and reasons about finer appearance changes due to denser sampling. While \cite{matzen2014scene} is limited to rendering planar regions, it was improved upon by~\cite{martin20153d}, which represents scene geometry using time-varying depth maps, enabling the generation of high-quality time-lapse videos. Recently, \cite{matzen2014scene} has been extended by employing neural fields \cite{lin2023neural} to achieve photo-realistic renderings with higher fidelity. 

Few works specifically handle point cloud sequences as inputs. Spatiotemporal reasoning from point cloud data is predominantly achieved by propagating temporal information via flow estimations~\cite{liu2019meteornet,choy20194d,fan2020pstnet}. ``Objects can move"~\cite{adam2022objects} has similarities to our benchmark and addresses indoor 3D change detection via geometric transformation consistency. However, it is limited to object relocation and cannot handle changes in the structure of the scene.

\subsection{Spatiotemporal 3D Datasets}
In recent years, several 3D datasets~\cite{Armeni_2016_CVPR, armeni2017joint, chang2017matterport3d, hua2016scenenn, dai2017scannet, straub2019replica,wald20193rscan,halber2019rescan,wald2020beyond} have emerged for indoor scene understanding, with some of them also considering the temporal aspect~\cite{straub2019replica,wald20193rscan,halber2019rescan,wald2020beyond}. However, not all of these datasets contain real-world scenes~\cite{park2021changesim}. In~\cite{Qiu2020IndoorData}, the authors generate a change dataset by leveraging an existing real-world static 3D dataset~\cite{chang2017matterport3d}. In order to generate change, they add synthetic models of small objects in the scenes (e.g., a cup or a car toy). There are three main datasets capturing real-world change in inhabited indoor spaces, namely Replica~\cite{straub2019replica}, ReScan~\cite{halber2019rescan}, and 3RScan~\cite{wald20193rscan}, that focus on the relocation, addition, or removal of furniture. RIO10~\cite{wald2020beyond} is a smaller version of 3RScan. These datasets capture aspects of daily human interaction with the built environment and are limited in spatial scale compared to \textbf{\abbrev} (one room versus one building floor with multiple rooms). For more details see Table~\ref{tab:datasets} and for visual samples Figure~\ref{fig:related_datasets}.

\begin{table}[h]
    \centering
    \resizebox{\linewidth}{!}{
    \small
    \begin{tabular}{lccccc}
            \hline
            \multirow{2}{*}{\textbf{Dataset}} & \multicolumn{2}{c}{\textbf{Area}} & \multicolumn{2}{c}{\textbf{Temporal Stage}} & \multirow{2}{*}{\shortstack{\textbf{Change}\\\textbf{Scale}}} \\
            \cmidrule(lr){2-3} \cmidrule(lr){4-5}
             & Num & Scale (\textit{Type}) & Total & Per Scene &  \\ \hline \hline
            \multirow{2}{*}{\textbf{Replica}~\cite{straub2019replica}} & \multirow{2}{*}{1} & room & \multirow{2}{*}{6} & \multirow{2}{*}{6} & \multirow{2}{*}{small}\\ 
             & & \textit{(typical living)} & & & \\ 
            \multirow{2}{*}{\textbf{ReScan}~\cite{halber2019rescan}} & \multirow{2}{*}{13} & room & \multirow{2}{*}{45} & \multirow{2}{*}{3-5} & \multirow{2}{*}{small}\\ 
             &  & \textit{(typical living)} & & & \\ 
            \multirow{2}{*}{\textbf{3RScan}~\cite{wald20193rscan}} & \multirow{2}{*}{478} & room & \multirow{2}{*}{1482} & \multirow{2}{*}{2-12} & \multirow{2}{*}{small}\\ 
             & & \textit{(typical living)} & & & \\ 
            \multirow{2}{*}{\textbf{RIO10}~\cite{wald2020beyond}} & \multirow{2}{*}{10} & room & \multirow{2}{*}{74} & \multirow{2}{*}{5-12} & \multirow{2}{*}{small}\\ 
             & & \textit{(typical living)} & & & \\ 
            \multirow{2}{*}{\textbf{\abbrev{}} \textit{(ours)}} & \multirow{2}{*}{6} & building & \multirow{2}{*}{27} & \multirow{2}{*}{2-6} & \multirow{2}{*}{large}\\ 
             & & \textit{(construction)} & & & \\ \hline
    \end{tabular}
    }
    \caption{\textbf{Comparison of existing indoor spatiotemporal 3D point cloud datasets.} \abbrev{} focuses on scenes that demonstrate large changes. Since the scale is on the building level, the number of areas (and temporal stages) is less than that of the other datasets but a single area contains numerous scenes on the scale of them.}
    \label{tab:datasets}
\end{table}

LAMAR~\cite{sarlin2022lamar} is similar to our benchmark. It is a large-scale dataset captured in diverse environments over an extended temporal horizon. However, it focuses on the task of visual localization from images and radios, while \textbf{\abbrev} concentrates on registration using point clouds. Additionally, the scenes captured in LAMAR do not exhibit significant changes in the environment's geometry and are more related to relocation scenarios. Other datasets focus on outdoor scenes and capture seasonal changes in real-world data~\cite{wenzel2020fourseasons} or are tailored to self-driving cars~\cite{Ros_2016_CVPR,HernandezBMVC17,bengarICCVW19,geiger2013vision,cordts2016cityscapes}. However, their review is outside the scope of this paper.

\subsection{3D Point Cloud Registration}
The field of 3D point cloud registration is well-established and active. Here, we discuss both pairwise registration and multi-way registration methods.

\textit{Pairwise registration}. Approaches here can be mainly grouped as \textit{feature-based} and \textit{end-to-end} registration. 

\subparagraph{Feature-based methods} typically involve two steps: local feature extraction and pose estimation. The pose estimation step uses either a robust estimator such as RANSAC~\cite{martin1981random} or globally optimal estimators~\cite{li20073d, hartley2009global, cai2019practical}. For local feature extraction, traditional methods use hand-crafted features~\cite{johnson1999, rusu2008PFH,rusu2009FPFH,tombari2010SHOT,tombari2010USC,theiler2014keypoint} to capture local geometry and, while having good generalization abilities across scenes, they often lack robustness against occlusions. In contrast, learned local features have taken over in the past few years, and, instead of using heuristics, they rely on deep models and metric learning~\cite{hermans2017defense,sun2020circle} to extract dataset-specific discriminative local descriptors. Depending on the input to models, these learned descriptors can be divided into patch-based and fully convolutional methods. Patch-based methods~\cite{gojcic20193DSmoothNet,ao2021spinnet} treat each point independently, while fully convolutional methods~\cite{Choy2019FCGF,bai2020d3feat} can extract all local descriptors for the whole scene in a single forward pass. \textsc{Predator}~\cite{predator} is the first work that pays special attention to low-overlap pairs and proposes an overlap-attention module to robustify registration by learning to sample interest points in the overlap region only. \cite{qin2022geometric,yu2021cofinet} improved \textsc{Predator} by operating in a coarse-to-fine manner. DPFM~\cite{attaiki2021dpfm} adopts this idea to non-rigid registration via an overlap attention mechanism in function space. 

\subparagraph{End-to-end methods} integrate differentiable pose estimators into the feature extraction pipeline~\cite{wang2019dcp,wang2019prnet,yew2020rpm,aoki2019pointnetlk, wei2023generalized}, providing an alternative to the feature-based methods. With the weighted Kabsch solver~\cite{kabsch} or the generalized differentiable RANSAC~\cite{wei2023generalized}, training can be directly supervised by ground truth poses. However, they mostly work on synthetic datasets~\cite{wu2015ModelNet} due to weak feature extractors.

In our experiments, we evaluate the performance on \abbrev{} of traditional~\cite{rusu2008PFH}, fully convolutional~\cite{bai2020d3feat,Choy2019FCGF}, attention-based~\cite{predator}, and coarse-to-fine~\cite{qin2022geometric} methods. 

\textit{Multi-way registration}
This task aims to resolve the ambiguities in pairwise registration by leveraging multi-view constraints. Traditional methods~\cite{huber2003fully,fantoni2012accurate} simply refine the initial pose estimations by extending ICP to the multi-way setting. However, this approach becomes computationally intractable due to the quadratically increased number of pairwise registrations as the views increase. Modern methods~\cite{torsello2011multiview,Choi2015robust,theiler2015globally,bernard2015solution,zhou2016fast,bhattacharya2019efficient} optimize the initial relative poses by incorporating global cycle consistency. This optimization is typically achieved by applying \textit{synchronization} techniques within the pose graphs using methods such as Iterative Reweighted Least Square (IRLS), Triggs' correction~\cite{triggs2000bundle}, and other lifting-based approaches~\cite{zach2018descending}. 
Recent studies have introduced learning-based methods to improve synchronization. For instance, ~\cite{huang2019learning} and ~\cite{gojcic2020learning} propose to learn the edge weights for the transformation synchronization, with~\cite{gojcic2020learning} additionally learning the pairwise registration~\cite{Choy2019FCGF} and outlier rejection~\cite{zhang2019learning}.
\cite{wang2023robust} propose constructing a sparse but reliable pose graph by first estimating the pairwise overlap ratios. They further improve the robustness of outlier edges by incorporating a history reweighting function in the IRLS scheme. 

In our experiments, we explore a widely-used pose graph synchronization method by Choi et al.\cite{Choi2015robust}, which is known for its efficiency in avoiding local minima by leveraging noisy-free odometry poses for initializing the nodes in the pose graph. Additionally, we investigate a recent synchronization method by Yew et al.\cite{yew2021learning}, PoseGraphNet, that does not assume prior knowledge of pairwise registration. Instead, it utilizes a recurrent Graph Neural Network (GNN) to progressively refine poses, beginning with node initializations as identity matrices. Note that for both methods, the edges in the graph are initialized using the results from pairwise registration, a common ground for constructing the pose graph.

\begin{figure*}[!t]
    \centering
    \includegraphics[width=0.99\linewidth,keepaspectratio]{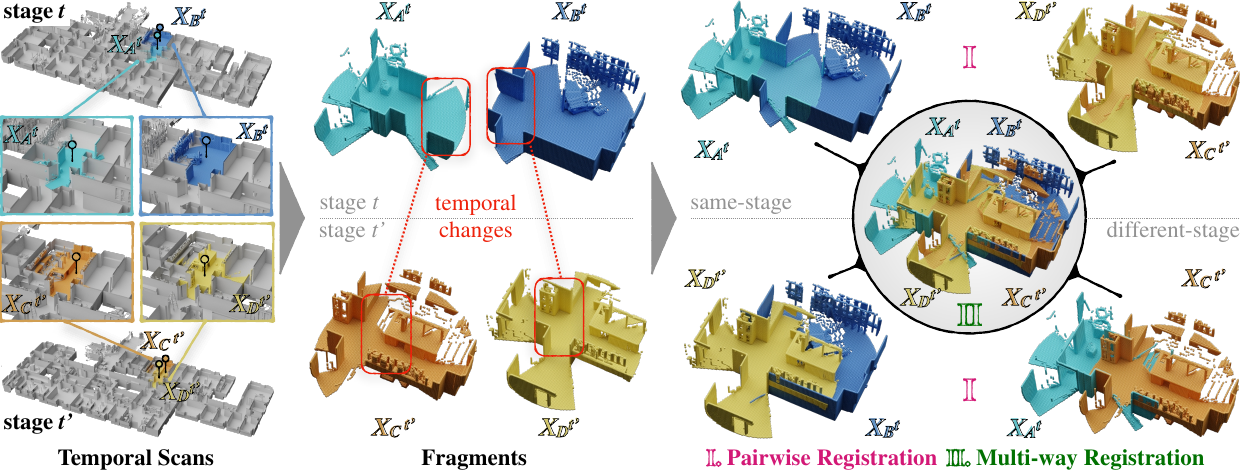}
    \caption{\textbf{Overview of the \project{} (\abbrev{}) benchmark: fragments (${X_A}^t$, ${X_B}^t$, ${X_C}^{t'}$, ${X_D}^{t'}$) captured in a construction site are spatiotemporally registered.} First, a pairwise registration step registers individually the pairs of fragments belonging to the same ((${X_A}^t$,${X_B}^t$) and (${X_C}^{t'}$,${X_D}^{t'}$)) or different stages ((${X_A}^t$,${X_C}^{t'}$) and (${X_B}^t$,${X_D}^{t'}$)). Then, a multi-way registration step creates a single and coherent spatiotemporal map of all fragments. Given current methods, this step is initialized by the results of the pairwise one. In this example, we assume overlap occurs for pairs (${X_A}^t$, ${X_B}^t$), (${X_B}^t$, ${X_D}^{t'}$), (${X_C}^{t'}$, ${X_D}^{t'}$), and (${X_C}^{t'}$, ${X_A}^t$). We define the overlapping pairs for entire areas using spatiotemporal graphs, as detailed in Sec.~\ref{sec:multiway_results}.}
    \label{fig:teaser}
\end{figure*}

\subsection{Synthetic Generation of Point Clouds} 
Generating synthetic visual data is a largely explored field. Here, we limit the scope to generating 3D point cloud data and depth images from two perspectives: sensor pose definition and sensor noise simulation. 

\paragraph{Sensor Pose Definition} Three main approaches are identified: (i) \textbf{Manual definition:} The user manually specifies the location of the sensor either by playing a video game~\cite{richter2016playing, shafaei2016play, hu2021sail} or via a graphical interface~\cite{handa2015scenenet, noichl2021bim}; (ii) \textbf{Real-world trajectory inserted in simulation:} A trajectory captured in the real world is inserted and transformed to simulate a trajectory in a synthetic scene~\cite{handa2012real, handa2014benchmark}; and (iii) \textbf{Random sampling in the synthetic scene:} Here, sensor locations and poses are randomly sampled in the simulation environment to address certain criteria. Methods use physics-based simulation of sensor trajectories~\cite{mccormac2017scenenet, roberto2017procedural}, react to dynamic movement of other objects in the scene~\cite{Ros_2016_CVPR, HernandezBMVC17, bengarICCVW19}, or densely sample in the free 3D space~\cite{kundu2020virtual, qiu2016unrealcv}. Closer to the latter and to our approach is a group of methods that define sensor positions with the use of a 2D occupancy map of the scene~\cite{song2017semantic, zhang2017, wang2020tartanair, biswasa2015planning, diaz2018scan, frias2019bim}. These methods commonly employ a set of constraints, criteria, and heuristics to exclude non-informative views from the final selection. Similar to them, we sample locations on a 2D occupancy grid of the scene and constrain the sensor location sampling based on the properties and way of use of a real-world sensor (\ie height position and distance between locations). However, instead of setting heuristics to exclude non-informative views, we define a probability map that favors more realistic locations (\eg further away from obstacles). In our final set of generated data, we include all possible scenarios from more to less informative. 

\paragraph{Sensor Noise Simulation} 
A well-known property of simulation environments is the possible mismatch in the distribution of the noise characteristics in the data. In practice, multiple scans of the same fragment may exhibit different noises. When simulating point clouds from any underlying geometry (\eg, reconstructed mesh, geometric primitives), to replicate this and prevent models from exploiting the consistent scanning artifacts during registration, works have been creating statistical noise models of sensors to utilize in simulation~\cite{noichl2021bim, handa2015scenenet, handa2014benchmark, barron2013intrinsic}. We follow the implementation in \cite{handa2015scenenet, gschwandtner2011blensor} to simulate the sensor noise during our synthetic data generation.

\section{Spatiotemporal Point Cloud Registration}
Before introducing the dataset and benchmark, we clarify all used terminology here: \textit{area} refers to a building's (large) indoor space that was recurrently captured over time; \textit{individual 3D point clouds} or else \textit{\onepc{s}} refer to partial 3D observations of the area in the form of point clouds; \textit{entire area point clouds} or else \textit{\allpc{s}} refer to the entire captured area reconstructed in the form of a 3D point cloud at one point in time; and \textit{stages} denote discrete points in time when an area was captured.

Given multiple \textit{\onepc{s}} of an \textit{area} that are captured at different \textit{stages} and 3D locations, the goal is to spatiotemporally align them and achieve a 3D \textit{\allpc} of the \textit{area} over time (\ie a 4D scan). This includes two tasks\footnote{State-of-the-art multi-way registration algorithms depend on initialization of the alignment between \onepc{s}, which can be acquired from the pairwise registration task. In the future, methods can solve the two spatiotemporal tasks independently without jeopardizing the structure of the benchmark.} (Figure~\ref{fig:teaser}): pairwise registration of \textit{\onepc{s}} that can belong to the same or different \textit{stages} (Figure~\ref{fig:teaser}(I)), followed by multi-way registration of all \textit{\onepc{s}} to result in the final spatiotemporal alignment of them (Figure~\ref{fig:teaser}(II)).

\paragraph{Pairwise registration} Consider source fragment $\mathbf{X}^{(S,t)} = \{\mathbf{x}_i^{(S,t)} \in \mathbb{R}^3\}^n_{i=1}$ and target fragment $\mathbf{X}^{(T,t^{'})} = \{\mathbf{x}_j^{(T,t^{'})} \in \mathbb{R}^3\}^m_{j=1}$ captured at $t$ and $t^{'}$, respectively. The spatiotemporal pairwise registration task is to recover a rigid transformation $\mathbf{M}^{*} = [\mathbf{R}^{*}, \mathbf{v}^{*}]$ where rotation matrix $\mathbf{R}^{*} \in \text{SO}(3)$ and translation vector $\mathbf{v}^{*} \in \mathbb{R}^3$, such that:

\begin{equation}
     \mathbf{M}^* = \argmin_{\mathbf{M}} \ \sum_{i=1}^{n} \left\| \mathbf{M}(\mathbf{x}^{(S, t)}_i) - \mathsf{NN}(\mathbf{M}(\mathbf{x}^{(S, t)}_i), \mathbf{X}^{(T,t')}) \right\|_2
\end{equation}

\noindent where $\mathbf{M}(\mathbf{x}) := \mathbf{R} \mathbf{x} + \mathbf{v}$ is the rigid transformation applied on point $\mathbf{x}$ and $\mathsf{NN}(\mathbf{x},\mathbf{X})$ represents the nearest neighbor of point $\mathbf{x}$ in point cloud $\mathbf{X}$ in Euclidean space.

\paragraph{Multi-way registration} 
Consider a set of \onepc{s} $\{X_i^{t}\}$ where each \onepc{} could be captured at any stage $t$. The spatiotemporal multi-way registration task is to recover a set of rigid transformations $\{\mathbf{M}_{i}^{t}\}$ between each \onepc{} in $\{X_i^{t}\} \backslash \{X_{i=1}^{t=1}\}$  and $X_{i=1}^{t=1}$, such that all \onepc{s} achieve a globally optimal alignment in the same reference system. Different from existing settings~\cite{Choi2015robust,huang2019learning,gojcic2020learning} that consider \onepc{s} from the same stage only, the proposed spatiotemporal multi-way registration contains \onepc{s} from different stages and is thus a more challenging optimization task. This step results in the spatiotemporal 3D reconstruction of the area.
\section{\xcapitalisewords{\project{} Dataset}}
\label{sec:dataset}

\begin{figure*}[bht!]
    \centering\includegraphics[width=0.88\linewidth,keepaspectratio]{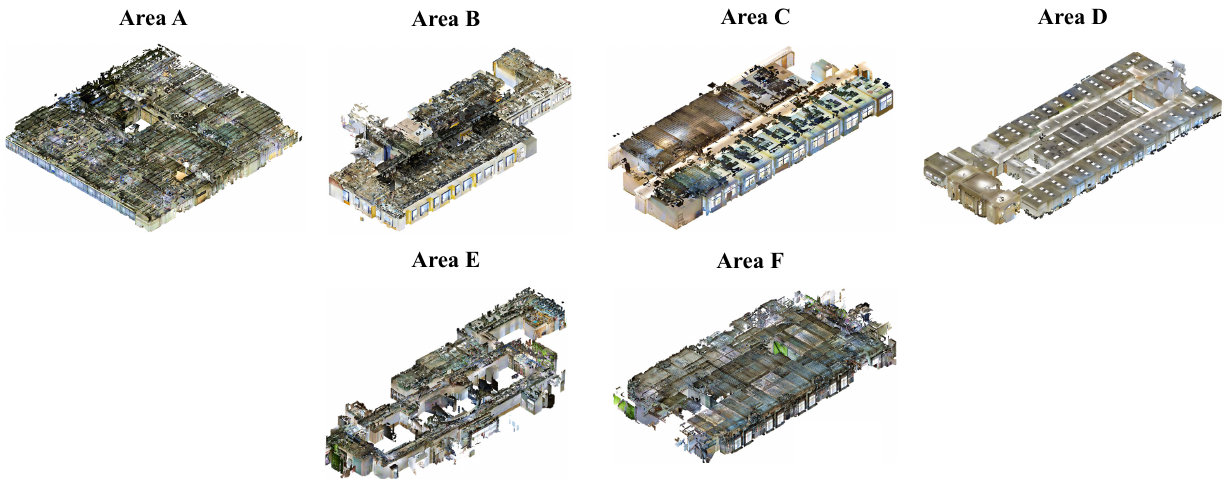}
    \caption{\textbf{Areas in the \textbf{\project{} dataset} at first temporal stage.} The building layout and size ranges across areas.}
    \label{fig:dataset}
\end{figure*}

\begin{figure*}[hbt!]
    \centering
    \includegraphics[width=0.99\linewidth,keepaspectratio]{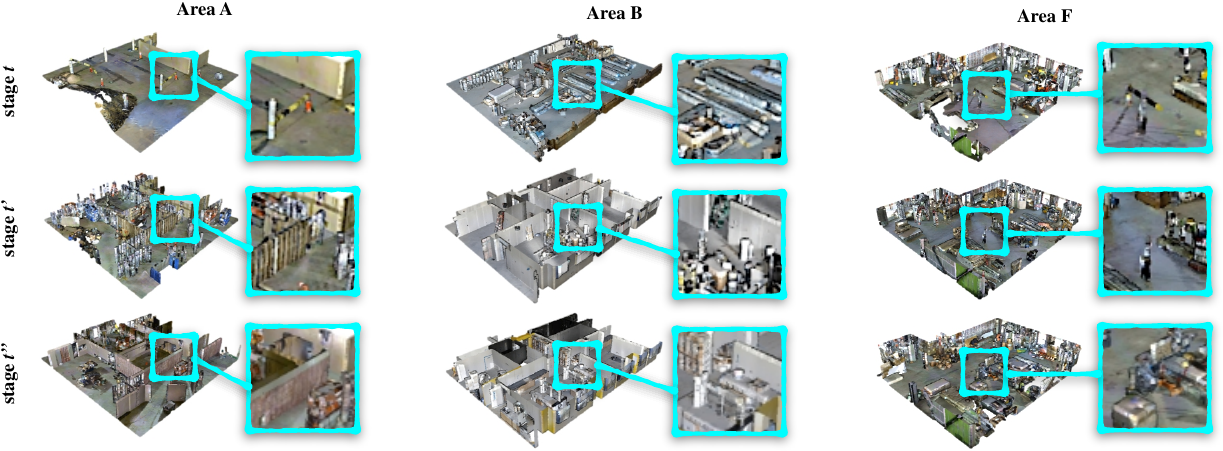}
    \caption{\textbf{Sample close-up snapshots of areas in the \textbf{\project{} dataset.}} Significant changes are occurring per area, starting from an empty scene and reaching the construction of rooms.}
    \label{fig:dataset_sample}
\end{figure*}

The \project{} (\abbrev{}) dataset consists of 3D \onepc{s} captured over time in 6 large-scale indoor areas, along with their corresponding \allpc{s}. The dataset focuses on the construction of interior layouts, where the exterior shell of the areas has been erected, and the interior space is empty. The captures chronicle the progression of various construction activities, including the creation of walls, installation of mechanical, electrical, and plumbing elements, movement of materials, temporary structures, machinery, and more. Figure~\ref{fig:dataset} provides a snapshot of all 6 areas at a single stage. Five of the areas in the dataset depict stages under construction, while one area (Area D) includes only before-and-after renovation stages with no visible construction. Still, the renovation stages involve significant structural changes like wall removal, functional changes like transforming a conference room into an office, and furniture replacement like desks and carpeting.

It is important to note that \onepc{s} belonging to different areas but annotated as being in the same stage (\eg stage $t'$) do not depict identical changes, as construction progresses differently across areas. Similarly, construction progress may vary even across \onepc{s} within the same area and stage. Figure~\ref{fig:dataset_sample} illustrates examples of areas and their stages. For all fragments in an area, the \abbrev{} dataset provides ground truth pose annotations that describe their spatial and temporal information.

\begin{figure*}[hbt!]
    \centering
    \begin{subfigure}{0.483\textwidth}
        \includegraphics[width=\columnwidth,keepaspectratio]{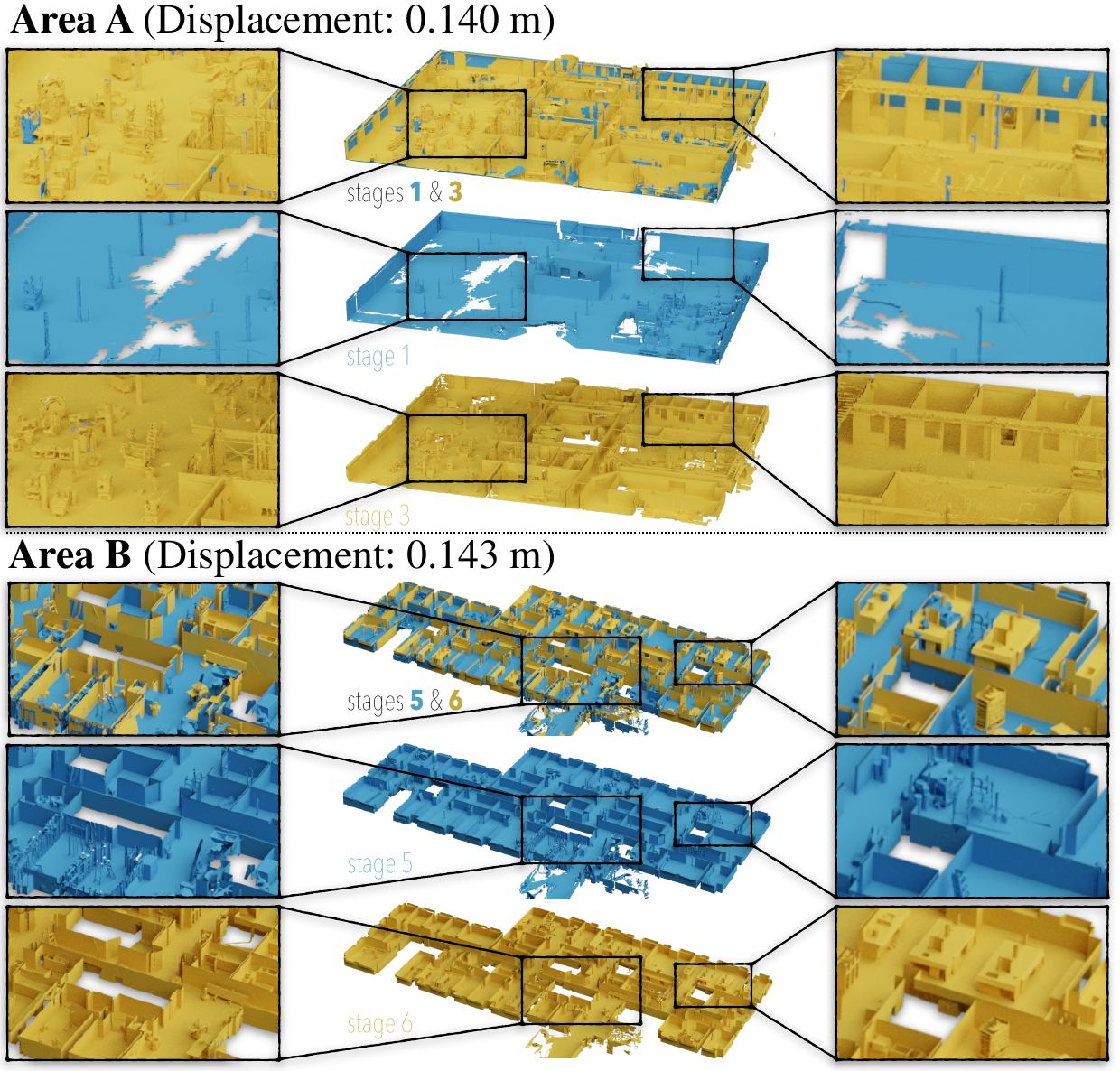}
    \end{subfigure}
    \hfill
    \begin{subfigure}{0.485\textwidth}
        \includegraphics[width=\columnwidth,keepaspectratio]{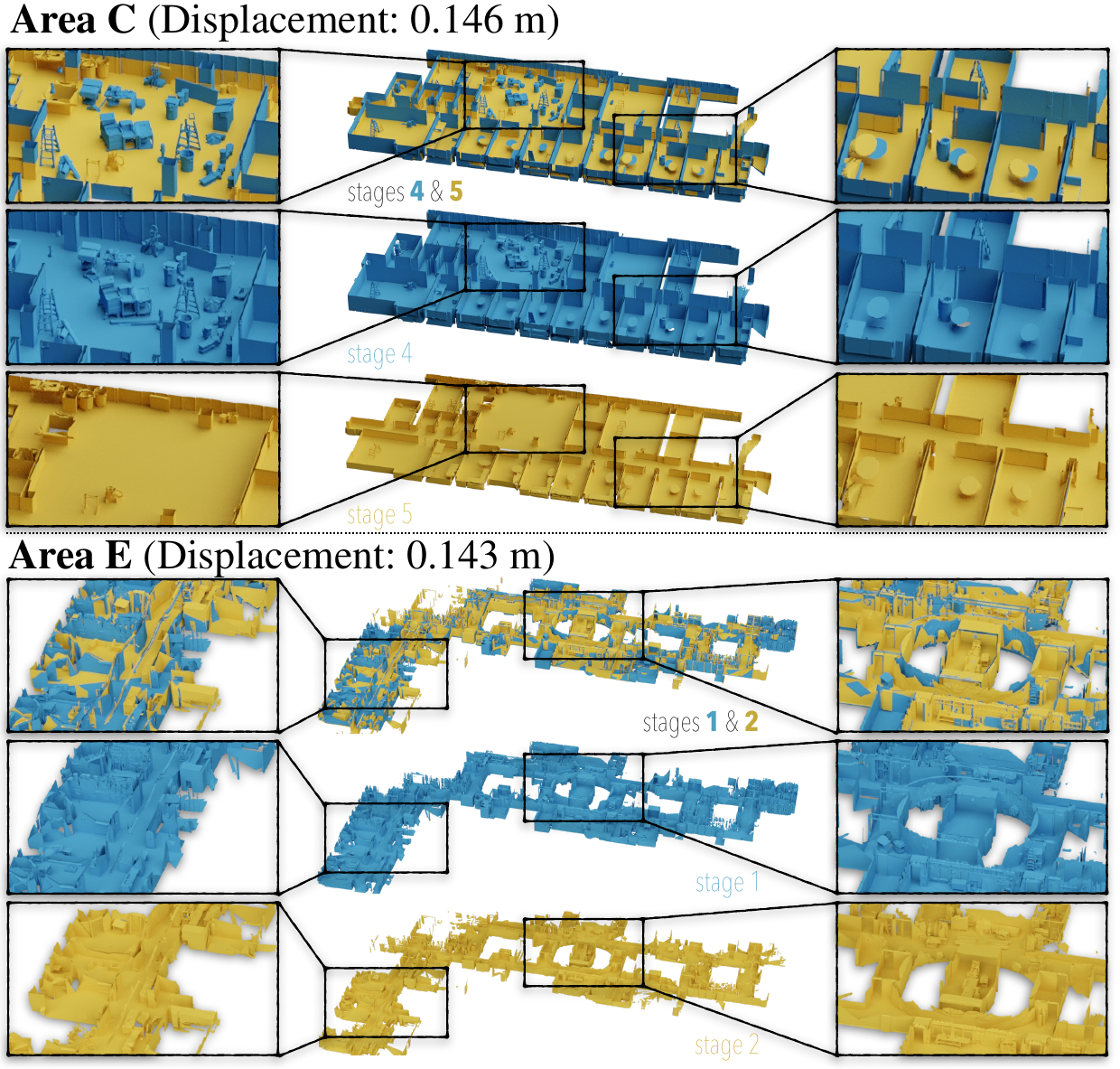}   
    \end{subfigure}
    \caption{\textbf{Global registration ground truth, for example, scans in four areas in the \project{} dataset}. Details about the alignment method for the global ground truth are provided in Sec. \ref{sec:scan_alignment}.}
    \label{fig:global_reg_gt}
\end{figure*}

\subsection{Dataset Acquisition}
Each area in the \abbrev{} dataset covers on average 2,500 $m^{2}$ and consists of 2--6 stages. The time intervals between stages can range from weeks to months, since the data collection is not based on a fixed schedule but rather follows the completion of significant construction tasks. The timing of data collection is determined in consultation with the project manager to ensure access to the construction site at appropriate and safe times. The objective is to capture the data just before crucial building information, such as pipes and structural elements, becomes inaccessible once covered by surfaces. Table~\ref{tab:data_split} provides details on the floorplan coverage for each area in $m^2$.

The dataset was collected using the Matterport Camera v1~\cite{matterport}. The Matterport Camera is a tripod-based reality capture system that acquires $360^{\circ}$ \onepc{s} from static locations. These \onepc{s} are subsequently registered together to create the final 3D \allpc{} of the captured area using proprietary software. The proprietary software is not accessible by or disclosed to users. Matterport3D automatically performs the registration upon uploading the data and provides to the user the final result. According to specifications, Matterport3D has a geometry error of around 1 inch from reality. Hence, while users have access to the 3D \allpc{s}, they do not have access to the individual \onepc{s}.
Furthermore, the 3D \allpc{s} from different stages depicting the same area are not spatiotemporally aligned in the same coordinate system because there is no geolocalization information available. Therefore, two important steps are undertaken in creating the dataset: (a) \textit{alignment of 3D \allpc{s}}: different-stage \allpc{s} of the same area are aligned in the same coordinate system to acquire all ground truth poses; and (b) \textit{\onepc{} generation}: as the original \onepc{s} are not accessible, novel ones are generated based on the provided \allpc{s}.

\subsection{Dataset Generation}
In this section, we describe the process of generating spatiotemporal ground truth pose information for all \onepc{s} that will serve the pairwise and multi-way registration tasks. Additionally, we explain how to generate  \onepc{s} and create pairs, as well as the formulation of the final dataset.

\subsubsection{Alignment of 3D \xcapitalisewords{\allpc{s}}}
\label{sec:scan_alignment}
To establish a single, global spatiotemporal coordinate system for all \allpc{s} that correspond to the same area, a manual rough alignment of them is initially performed. This involves using the ``Align (point pairs picking)'' tool in CloudCompare~\cite{cloudcompare}, a point cloud processing software. By manually selecting 10-15 correspondences between \allpc{s}, we obtain an initial alignment. In cases where an area consists of more than two stages, an \textit{anchor} stage is selected based on the highest area coverage across all stages. The remaining stages are then aligned with respect to this anchor stage. To refine the above alignment results, we programmatically employ the iterative closest point (ICP)~\cite{icp} algorithm. ICP aims to minimize the root mean square error (RMSE) between the input stages, ensuring a more accurate result. Examples of the global registration results are shown in Figure~\ref{fig:global_reg_gt}.

\begin{figure*}[t]
    \centering
    \begin{subfigure}{0.25\textwidth}
        \includegraphics[width=\columnwidth,keepaspectratio]{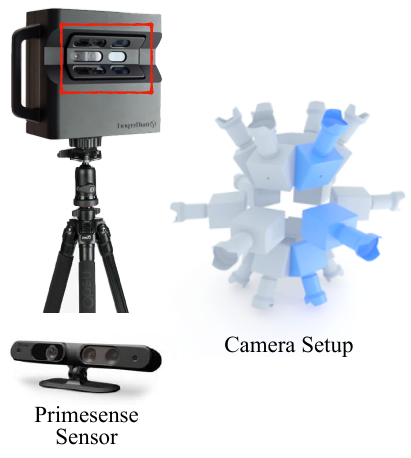}
        \caption{Matterport Camera Setup}
        \label{fig:matterport}
    \end{subfigure}
    \hfill
    \begin{subfigure}{0.7\textwidth}
        \includegraphics[width=\columnwidth]{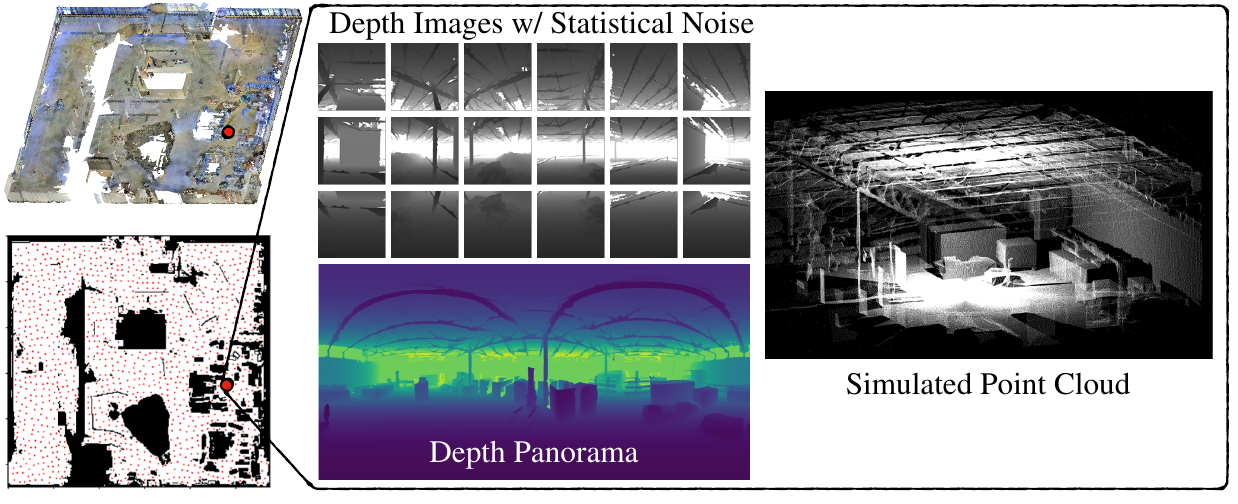}
        \caption{An example of generating synthetic fragments.}
        \label{fig:synthetic_frags}
    \end{subfigure}
    \caption{\textbf{Settings of sensor utilized for generating \onepc{s}.} (a) Given the used sensor settings, (b) we simulate 3 depth sensors with different pitch angles and statistical noise to capture the individual \onepc{s} per location. (\textit{Best viewed in color})}
    \label{fig:camera_setup}
\end{figure*}

\paragraph{Why ICP} The use of ICP for this change-depicting data is not ideal, since one of ICP's main assumptions is that the scene is static. A better circumstance would be to identify changed points and exclude them from the optimization process. However, this is non-trivial because determining the changed parts requires aligning the data first, leading to a circular dependency problem.
As a compromise, we choose ICP to refine the rough initial alignment and assume that non-changing points dominate the optimization process. The median cross-stage displacements (\ie the distance between a point in a non-anchor stage and its corresponding point in the anchor stage) after applying ICP for each area are as follows: $0.127$, $0.135$, $0.141$, $0.130$, $0.127$, and $0.117$ m, respectively. The effect of ICP can also be qualitatively evaluated in Figure~\ref{fig:global_reg_gt}.

\subsubsection{\xcapitalisewords{\onepc{} generation}}
We utilize the available 3D \allpc{s} to generate synthetic \onepc{s} that mimic real-world conditions, such as sensor settings and the capturing process. 

\paragraph{Sensor Settings} The Matterport Camera consists of three RGBD sensors \cite{primesense}. The configuration of the sensors is optimized to achieve maximum vertical coverage of the scene from a single viewpoint, with a pitch range of $\pm30^{\circ}$. As the sensors rotate around the gravity axis, the system captures data at intervals of $60^{\circ}$ (Figure~\ref{fig:matterport}). This process generates 18 RGBD images, which are then stitched together to form an equirectangular RGBD image. Further projecting the equirectangular image in the 3D space results in the \onepc{} captured at that location (Figure~\ref{fig:synthetic_frags}). 

To simulate this sensor setup, we use the Blender~\cite{blender} software and model it according to the described configuration. We incorporate statistical noise models~\cite{handa2015scenenet, gschwandtner2011blensor} for the Primesense sensor to mimic the real-world characteristics of the captured data. The depth images are sampled based on the reconstructed mesh of the scene, allowing us to closely simulate the raw output of the actual sensor at a specific location. In our simulation setup, we focus on simulating the depth sensor only, as accurately simulating realistic textures from the reconstructed 3D mesh is a challenging task. This does not affect the benchmark, since pairwise and multi-way registration tasks rely on geometric information rather than color.

\begin{figure*}[hbt!]
    \centering
    \includegraphics[width=0.9\linewidth]{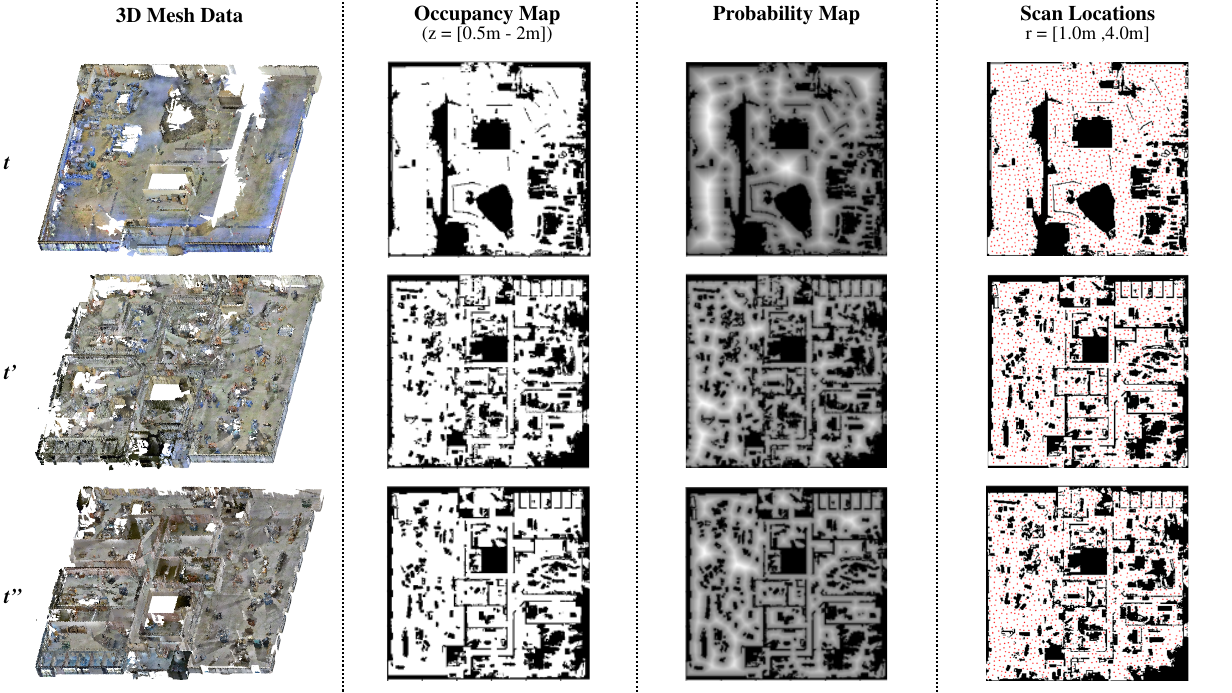}
    \caption{\textbf{Sampling \onepc{} locations on scans.} Locations are selected on the basis of a probabilistic 2D occupancy map, taking into account the employed sensor characteristics and real-world settings. (\textit{Best viewed in color})}
    \label{fig:occupancy}
\end{figure*}

\paragraph{Finding \onepc{} locations} The next step is to sample possible 3D locations of the sensor in each area and stage, so as to achieve maximum coverage. We compute these locations on a 2D occupancy probabilistic map of each stage (Figure~\ref{fig:occupancy}), by taking into account constraints imposed by the sensor system. First, we calculate a 2D occupancy map for each stage by taking into account the obstacle information in the vertical space. We exclude any data outside the height range of $[0.5, 2]$ m to remove occupancy resulting from floor or ceiling points. The maximum sensor height is set at $1.75$ m, so any location below $2$ m should be unobstructed for it to be considered valid. To create the occupancy map, we densely sample 3D points from the underlying mesh in a uniform manner.
The map is defined by a grid with a cell size of $0.10$ m $\times$ $0.10$ m, and if a point falls within a cell, the cell is marked as occupied.

Next, we enrich the occupancy map with probabilistic information that prioritizes free cells that are further away from occupied ones. This ensures that the sensor locations are preferentially placed further away from obstacles, as would occur in a real-world setting. The probabilistic occupancy map is then used to densely sample sensor locations. The sampling process starts by randomly selecting the first point, and subsequent sensor locations are placed within a 2D Euclidean distance uniformly distributed in the range of $[1, 4]$ m. The sensor height is varied in the range of $[1.5, 1.75]$ m, also uniformly distributed. The objective is to achieve maximum coverage of the 2D map, taking into account that the maximum depth sensing range of the employed sensor is $4.5$ m.

\subsubsection{\xcapitalisewords{pairwise registration dataset generation}}
\label{sec:pairwise_dataset} 
\begin{figure}[hbt!]
    \centering
    \includegraphics[width=\columnwidth]{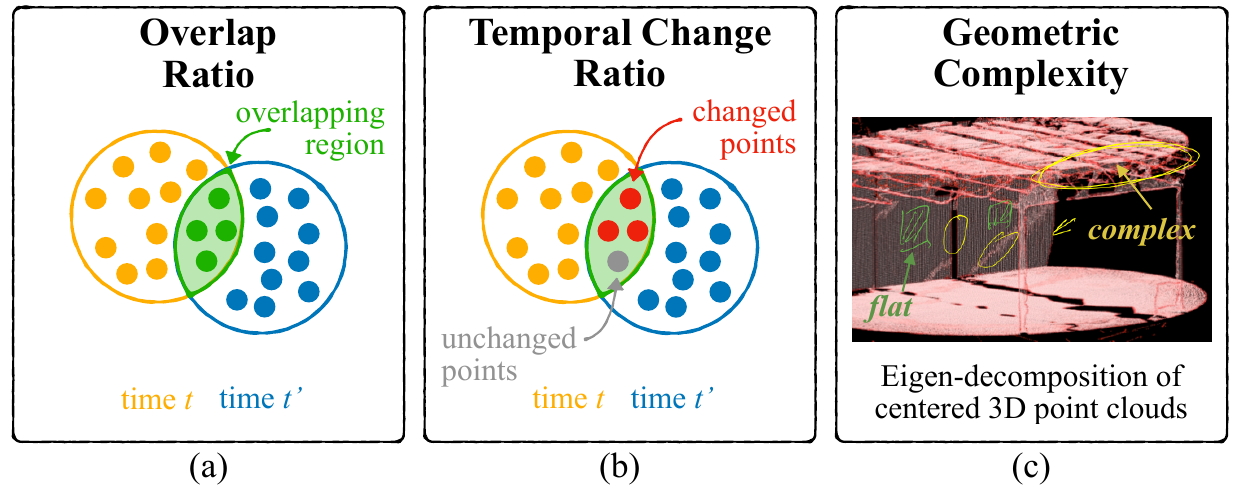}
    \caption{\textbf{Metrics for selecting \onepc{} pairs.} We employ three metrics that evaluate (a) spatial, (b) temporal, and (c) geometric characteristics of fragment pairs. (\textit{Best viewed in color})}
    \label{fig:dataset_metrics}
\end{figure}

To select \onepc{} pairs for the pairwise registration task, we aim to create a diverse and balanced dataset that represents various scenarios of overlap. We employ three metrics to guide the selection process:

\paragraph{Overlap Ratio (OR)} This is an existing metric in the spatial registration domain~\cite{predator} and refers to the ratio of spatially overlapping points between two \onepc{s}, regardless of whether they belong to the same or different stage. Given a pair of registered \onepc{s}, it measures the ratio of overlapping points over the whole point cloud (Figure~\ref{fig:dataset_metrics} (a)). Specifically, given source $\mathbf{X}^{(S)}$ and target $\mathbf{X}^{(T)}$ \onepc{s}, the overlapping part between them $\mathbf{O}(\mathbf{X}^{(S)}, \mathbf{X}^{(T)})$ is calculated as:
\begin{equation}
    \mathbf{O}(\mathbf{X}^{(S)}, \mathbf{X}^{(T)}; \tau) := \{\mathbf{x} \in \mathbf{X}^{(S)} \mid \mathsf{NN}(\mathbf{x}, \mathbf{X}^{(T)}) \leq \tau \}
\end{equation}
Then, the overlap ratio is defined as:
\begin{equation}
    \mathrm{Overlap\ Ratio} := \frac{|\mathbf{O}(\mathbf{X}^{(S)}, \mathbf{X}^{(T)}; \tau)|}{|\mathbf{X}^{(S)}|}
\end{equation}
Note that, in the case of different stage registration, the overlap ratio reflects the ratio of no-change points under the threshold $\tau$. For all evaluations, we set $\tau=0.2$ m, a threshold commonly used in scanning-based 3D datasets to determine a sufficiently close to the ground-truth transformation~\cite{zeng20163dmatch}.

\paragraph{Temporal Change Ratio (TCR)} 
\textit{OR} falls short in providing information about possible temporal changes that might have occurred in the overlapping region between two \onepc{s} that originate from different stages. To counter this limitation, we define and introduce the concept of the \textit{temporal change ratio}. This ratio denotes the proportion of points that have undergone changes within the overlap region encapsulated by a 3D convex hull (Figure~\ref{fig:dataset_metrics} (b)). Following the same threshold used in OR, we consider a point as changed if it lacks neighbors within the $\tau=0.2$ m Euclidean range in the other stage.

More specifically, given source \onepc{} $\mathbf{X}^{(S, t)}$ from stage $t$ and target \onepc{} $\mathbf{X}^{(T, t')}$ from stage $t'$, the temporal change ratio is defined as:
\begin{equation}
    \mathrm{TCR} := 1 - \frac{|\mathbf{O}(\mathbf{X}^{(S, t)}, \mathbf{X}^{(T, t')}; \tau)|}{|\mathbf{H}(\mathbf{X}^{(S, t)}, \mathbf{X}^{(T, t')})| }
\end{equation}
Here, the convex envelope $\mathbf{H}$ represents the boundary of the overlap region between the two \onepc{s}, and is defined as:
\begin{equation}
    \begin{aligned}
        \mathbf{H}(\mathbf{X}^{(S, t)}, \mathbf{X}^{(T, t')}) := \{ \mathbf{x} \in \mathbf{X}^{(S, t)} \mid \mathrm{hull}(\mathbf{X}^{(T, t')}) = \\ \mathrm{hull}(\mathbf{X}^{(T, t')} \cup \mathbf{x})\}
    \end{aligned}
\end{equation}
where $\mathrm{hull}(\cdot)$ is the convex hull of a given \onepc{}. 

\begin{figure}[t!]
    \centering
    \includegraphics[width=\columnwidth]{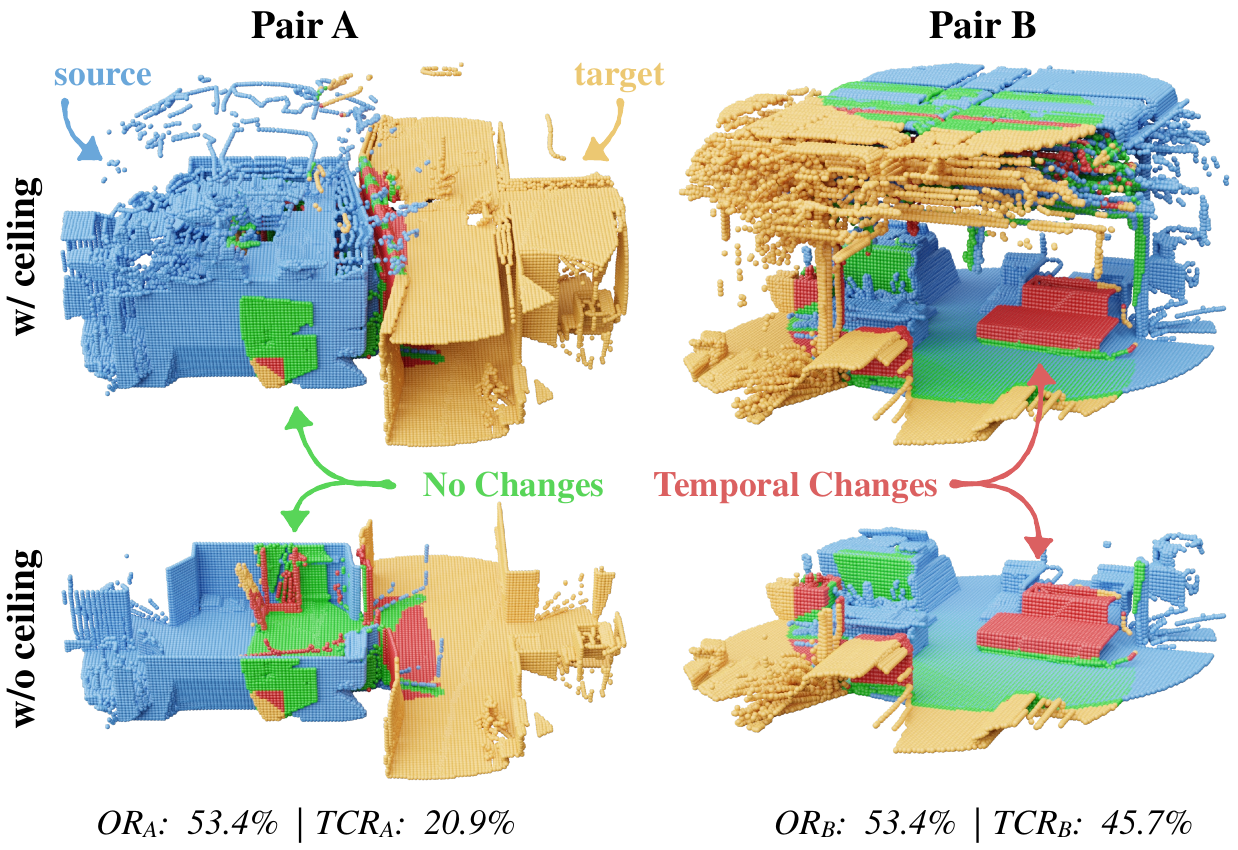}
    \caption{\textbf{Significance of overlap (OR) and temporal change (TCR) ratios in \abbrev{}.} Although the OR is the same in both pairs, TCR is substantially higher in pair B with almost half of the points in the overlapping area having changed. (\textit{Best viewed in color})}
    \label{fig:oc_tcr}
\end{figure}

Figure~\ref{fig:oc_tcr} showcases two examples of \onepc{} pairs, along with their respective overlap and temporal change ratios. While the overlap ratio remains consistent in both cases, a notable difference can be observed in the temporal change ratio. This discrepancy indicates that registering (b) is more difficult than (a) due to the scarcity of static points available for deriving correspondences within the overlapping region. Adding to the challenge is the fact that the majority of static points are associated with flat surfaces, which further restricts finding correspondences.

\paragraph{Geometric Complexity} The amount of geometric complexity of points in the overlapping region between two \onepc{s} plays an important role in defining easier versus more challenging registration pairs. To assess it, we calculate the surface variation or else curvature~\cite{weinmann2013feature}, which provides information about the local shape, using  eigendecomposition (Figure~\ref{fig:dataset_metrics} (c)):
\begin{equation}
    {C_\lambda}({\bf{x}}) = \frac{\lambda_3}{\lambda_1 + \lambda_2 + \lambda_3} 
\end{equation}
where $\lambda_i$ is the $i$-th eigenvalue for the 3D Structure Tensor~\cite{bigun1987optimal} over the points within a sphere of radius $r=0.5$ m centered at ${\bf{x}}$. This radius is empirically chosen to include enough points to reduce noise while maintaining sensitivity to local features. For every pair of \onepc{s}, we provide the averaged curvature value for all points located in their overlapping part, i.e.,
\begin{equation}
{C_\lambda}(\mathbf{X}^{(S)}, \mathbf{X}^{(T)}) = \frac{1}{|O|}\sum_{{\bf{x}} \in O}{ {C_\lambda}({\bf{x}}) }, \\ \text{where} \ O := \mathbf{O}(\mathbf{X}^{(S)}, \mathbf{X}^{(T)}).
\end{equation}
Regions with higher curvature values indicate more intricate and complex geometry and are generally easier to align. Regions with flatter geometry, characterized by lower curvature values, make the registration task more challenging since they exhibit simpler geometric shapes with less variability. 

\begin{figure*}[t!]
    \centering
    \begin{subfigure}{\linewidth}
    \centering
        \includegraphics[width=1\linewidth]{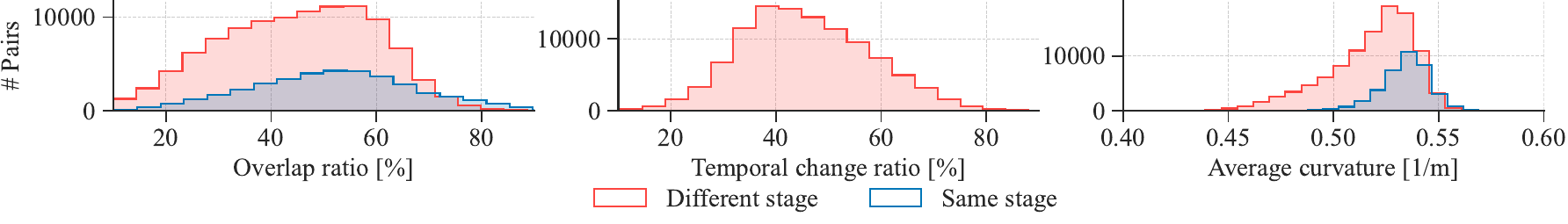}
        \caption{All generated \onepc{} pairs.}
        \label{fig:all_stats}
    \end{subfigure}
    \\
    \begin{subfigure}{\linewidth}
    \centering
        \includegraphics[width=1\linewidth]{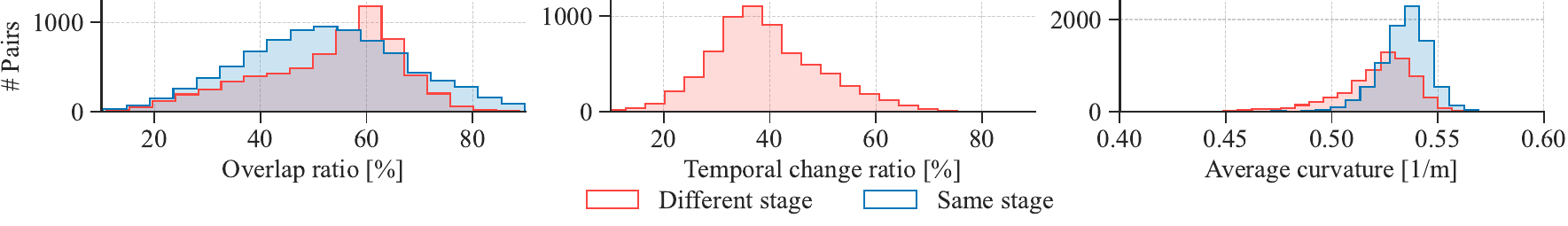}
        \caption{Final selected \onepc{} pairs.}
        \label{fig:dataset_stats}
    \end{subfigure}
    \caption{\textbf{\abbrev{} dataset statistics}. The histograms showcase the distribution of \onepc{} pairs with respect to spatial and temporal characteristics. These are the overlap ratio, temporal change ratio, and average curvature.}
    \label{fig:basic_dataset_stats}
    \vspace{-3pt}
\end{figure*}

\paragraph{Final Fragment Pairs} To determine the final set of fragment pairs for the \project{} dataset, we compute the metrics described above for all possible pairs within and across stages. After computing them, we create the distribution curves which provide insights into the data characteristics (Figure~\ref{fig:all_stats}). To ensure a diverse and balanced dataset, we sample data in a uniform manner from them (Figure~\ref{fig:dataset_stats}), \ie we select \onepc{} pairs so that they represent a range of overlap ratios, temporal change ratios, and geometric complexities. Details on the final number of \onepc{} pairs for the \textbf{\project{}} dataset are shown in Table~\ref{tab:data_split}.

\begin{table}[t!]
    \centering
    \footnotesize
    \begin{tabular}{ccccccc}
    \hline
    \multirow{2}{*}{\textbf{Area}}  & \multirow{2}{*}{\textbf{Stages}} & \multicolumn{2}{c}{\textbf{Area} [$m^2$]} & \multicolumn{2}{c}{\textbf{Pairs}}\\
    \cmidrule(lr){3-4} \cmidrule(lr){5-6}
      &  & Min & Max &  {Non-temporal} & {Temporal} \\ \hline\hline
    A & 3 & 3159.8 & 3342.0 & 9604 & 3825 \\
    B & 6 & 1482.4 & 2191.9 & 6801 & 5876\\
    C & 5 & 652.9 & 812.9 & 2185 & 1682\\
    D & 2 & 1112.1 & 1129.1 & 1557 & 844\\
    E & 4 & 944.3 & 5019.5 & 5224 & 2226 \\
    F & 5 & 1322.4 & 2661.1 & 12060 & 14359\\ \hline
    \end{tabular}
    \caption{\textbf{Details on area coverage and the total number of \onepc{} pairs in the \abbrev{} dataset.} The coverage of each area in the dataset may vary slightly at each stage due to construction activities, since certain parts may be inaccessible or obstructed.}
    \label{tab:data_split}
\end{table}

\subsubsection{\xcapitalisewords{\onepc{} alignment}}
\label{sec:local_alignment}
While the global alignment achieved in Section~\ref{sec:scan_alignment} provides a globally minimum registration error among \allpc{s}, it does not guarantee a locally optimal registration between \onepc{} pairs. This is illustrated in Figure~\ref{fig:global_reg_gt}. Using this initial and imperfect alignment as ground truth to systems will result in a very noisy learning process and ultimately to gross errors in registration. To address this, individual refinement of transformations is performed for each pair to achieve a locally minimum solution. 

However, directly refining the fragment pairs alone does not always yield the optimal solution for that pair. Context, particularly overlapping regions, plays a crucial role in registration. More context can provide more static anchors, but excessive context can hinder the process if there is significant temporal change. To determine the optimal context size around a \onepc{} location, cylindrical chunks are cropped from the \allpc{s} centered around the sensor locations of each fragment. Cylinders of different radii in the range of $[2, 18]$ m are used, with a step size of $2$ m. The range is selected empirically to allow for sufficient hyper-parameter tuning, with the upper limit encompassing the entire building in most cases. This results in 9 different-sized cylinders per sensor location, with the height of each cylinder being the total height of the \allpc{} at that location.

\begin{figure}[hbt!]
    \centering
    \includegraphics[width=\columnwidth,keepaspectratio]{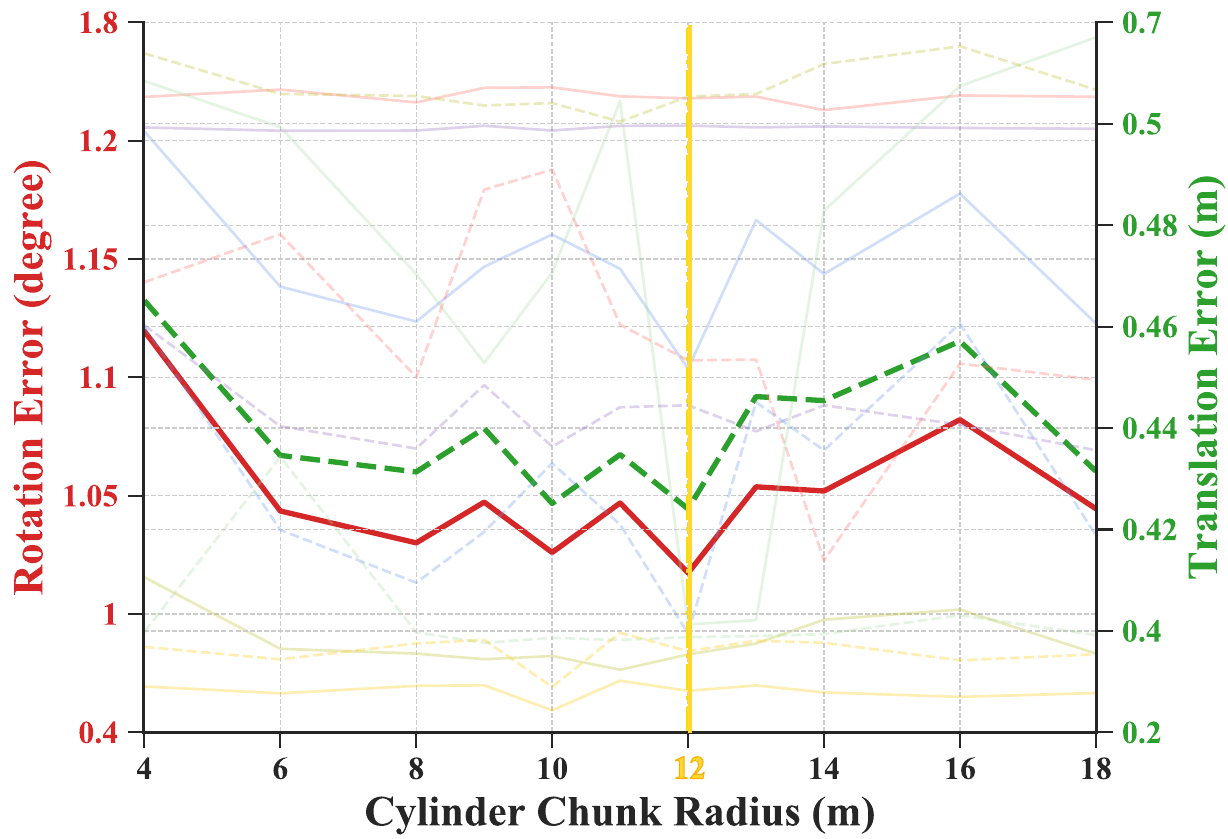}
    \caption{\textbf{Refinement of \onepc{} pair ground truth registration using cylinders of different radius.} The $12$ m radius is the one that provides the smallest error correction with respect to the initial global alignment. We plot the average error curves in \textcolor{red}{\textbf{red}} and \green{\textbf{green}}. Error curves per area are colorized as: \textcolor{A_curve}{A: purple} | \textcolor{B_curve}{B: blue} | \textcolor{C_curve}{C: pink} | \textcolor{D_curve}{D : green} | \textcolor{E_curve}{E : yellow} | \textcolor{F_curve}{F :orange}}
    \vspace{-5pt}
    \label{fig:cylinder_registr_plots}
\end{figure}

\begin{figure*}[t!]
    \centering\includegraphics[width=1.0\linewidth,keepaspectratio]{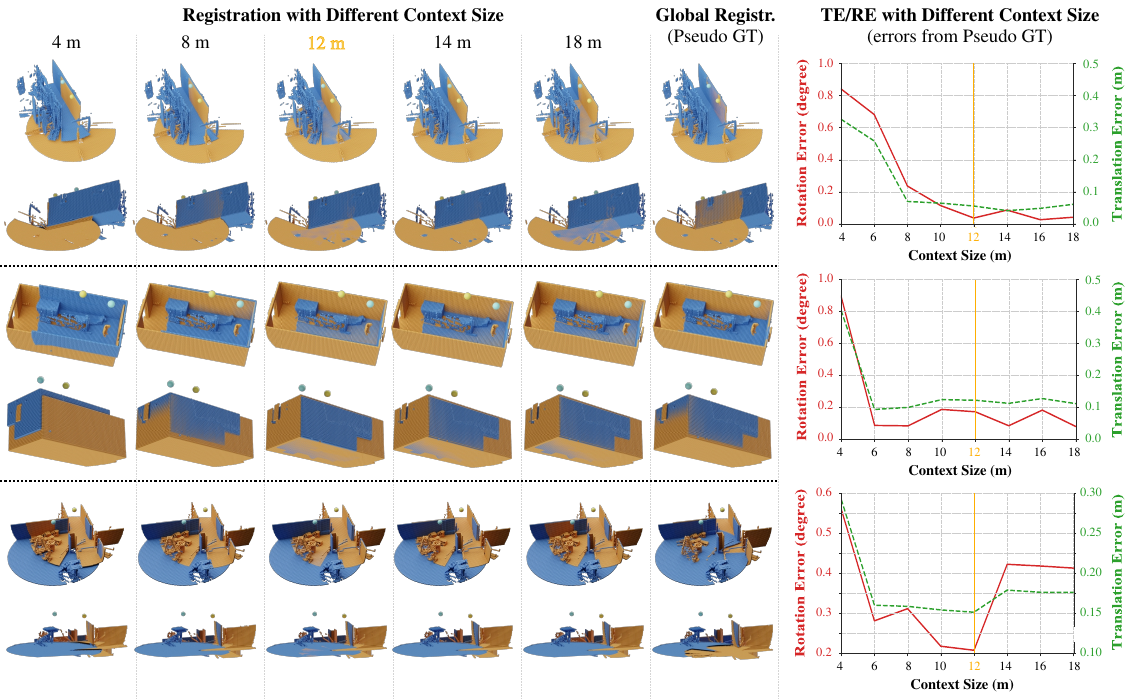}
    \caption{\textbf{Qualitative examples of the fine-tuned pairwise alignment using different context sizes.} The $12$ m radius provides on average the lowest translation (TE) and rotation (RE) errors between the pseudo ground truth and the fine-tuned result. This is particularly noticeable in the alignment of floors and walls. Even in the last row, where the TE is higher, the $12$ m radius still achieves a correct alignment. Two viewpoints are shown for a better understanding of the results. The ceiling has been removed for visualization purposes.}
    \label{fig:cylinder_registr_quals}
\end{figure*}

The alignment of the cylinder pairs is refined using the global alignment computed in Section~\ref{sec:scan_alignment} as the initialization (pseudo ground truth). The relative reconstruction error (translation error (TE) and rotation error (RE)) is computed with respect to the pseudo ground truth. The assumption is that the optimal radius will have a minimal deviation in terms of reconstruction error from it. Based on the results (Figure~\ref{fig:cylinder_registr_plots}), it is determined that a radius of $12$ m provides a balance of context for refined \onepc{} registration. This choice is confirmed by visualizing various random samples of refined \onepc{} pairs (Figure~\ref{fig:cylinder_registr_quals}). Even when there is larger-than-average displacement from the pseudo ground truth, the final registration results are improved. Finally, the pairwise registration ground truth is created using the refined local transformations on the $12$ m cylinders for all \onepc{} pairs.

\subsubsection{Multi-way Registration Dataset Generation}
For the multi-way registration task, the ground truth transformation across \onepc{s} is obtained from the global alignment achieved in Section~\ref{sec:scan_alignment}.
It is worth noting that the train and test sets for the multi-way registration are a subset of those used in the pairwise registration. This is due to certain \onepc{} pairs in the pairwise registration task not having sufficient overlap with other pairs (\ie at least 10\%). This results in many disjoint components per area instead of a coherent and connected global spatiotemporal map. While the goal is to have high intra-\onepc{} overlap, we keep disjoint components that contain enough fragments (at least 50). As a result, in certain areas, there may be more than one disjoint spatiotemporal map. During the simulation process, it is possible to generate fragments that create a single map per area, however, this scenario is not always realistic. We note that, in the multi-way registration experiments in Section~\ref{sec:multiway_results}, we use the subset test data but not the subset training data. State-of-the-art methods for this task aim to globally optimize the pairwise registration results, which does not require new training. The multi-way training annotations are provided as part of the \abbrev{} dataset for future work that may address this task independently of a prior pairwise registration step.

\section{\xcapitalisewords{\project{} Benchmark}}
\label{sec:benchmark_tasks}
The \textbf{\xcapitalisewords{\project{} (\abbrev{})} benchmark} consists of the tasks of pairwise and multi-way registration. \vspace{-15pt}

\paragraph{Data Splits} To evaluate the generalization ability of the methods, we define three different data splits: 
\begin{itemize}
\item \textit{Original}: This is the standard data split in the spatial registration domain. The training and testing are performed on \onepc{} pairs from all areas and stages. Although the train-test fragments are not duplicates, they originate from the same area and stage, allowing methods to learn about the composition of an area during training.
\item \textit{Cross-area}: In this split, the training is done on fragments from three areas (all stages), and the testing is performed on the remaining areas. This evaluates a method's generalization ability to unseen areas.
\item \textit{Cross-stage}: Here, the training is conducted on the first 50\% of stages of each area, and the testing is performed on the remaining ones. This split aims to assess the domain gap across stages.
\end{itemize}

Table~\ref{tab:benchmark_tasks} provides an overview of the splits, and Table~\ref{tab:area_split} offers more detailed information about them. Both the pairwise registration and multi-way registration tasks are evaluated on all three data splits.
\begin{table}[hbt!]
    \footnotesize
    \centering
    \begin{tabular}{p{3.5cm}|cc}
    \hline
     \textbf{Data Split} & \textbf{Unseen Stage} & \textbf{Unseen Area} \\ \hline \hline
    Original & \includegraphics[width=6pt]{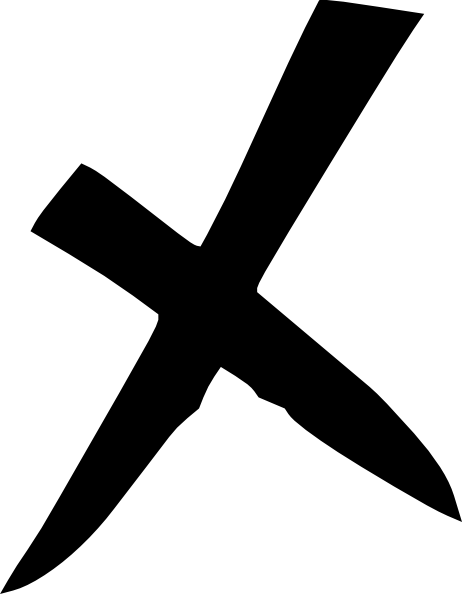} & \includegraphics[width=6pt]{figs/x.png}\\
    Cross-Area & \includegraphics[width=6pt]{figs/x.png} & \includegraphics[width=6pt]{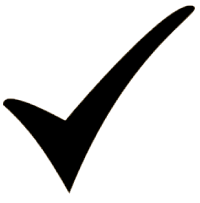}\\
    Cross-Stage & \includegraphics[width=6pt]{figs/tick.png} &
    \includegraphics[width=6pt]{figs/x.png}\\ \hline
    \end{tabular}
    \caption{\textbf{Generalization type in \abbrev{} data splits}. Each split receives during testing, with respect to training, data from \textit{unseen areas}, and \textit{unseen stages} from the same area.}
    \label{tab:benchmark_tasks}
\end{table}

\begin{table}[hbt!]
    \centering
    \resizebox{\columnwidth}{!}{
    \begin{tabular}{l|ccc|ccccc|c}
    \hline
     & \multicolumn{3}{c|}{\textbf{Cross-Area}} & \multicolumn{5}{c|}{\textbf{Cross-Stage}} & \textbf{Original} \\ \hline \hline
    \multicolumn{10}{c}{\textit{Training}} \\ \hline
    Area  & A   & B   & F   & A   & B   & C   & E   & F   & all \\
    Stage & all & all & all & 1-2 & 1-3 & 1-3 & 1-2 & 1-3 & all \\ \hline
    \multicolumn{10}{c}{\textit{Testing}} \\ \hline
    Area  & C   & D   & E   & A & B   & C   & E   & F   & all \\
    Stage & all & all & all & 3 & 4-6 & 4-5 & 3-4 & 4-5 & all \\ \hline
    \end{tabular}
    }
    \caption{\textbf{Area split per evaluation task on the \abbrev{} dataset.}}
    \label{tab:area_split}
\end{table}

\begin{table*}[ht!]
    \centering
    \setlength\tabcolsep{3pt}
    \setlength\extrarowheight{2pt}
    \resizebox{\linewidth}{!}{
    \begin{tabular}{lcccccccccccc}
            \toprule
            \multirow{3}{*}{\textbf{Method}} & \multicolumn{4}{c}{\textbf{Cross-Area}} & \multicolumn{4}{c}{\textbf{Cross-Stage}} & \multicolumn{4}{c}{\textbf{Original}}  \\  \cmidrule(lr){2-5}  \cmidrule(lr){6-9}  \cmidrule(lr){10-13}
             & \textbf{Recall} & \textbf{RMSE}  & \textbf{TE} & \textbf{RE}   & \textbf{Recall} & \textbf{RMSE} & \textbf{TE} & \textbf{RE} & \textbf{Recall} & \textbf{RMSE} & \textbf{TE} & \textbf{RE} \\ 
             & [\% $\uparrow$] & [m $\downarrow$] & [m $\downarrow$] & [$^\circ$ $\downarrow$] & [\% $\uparrow$] & [m $\downarrow$] &  [m $\downarrow$] &[$^\circ$ $\downarrow$] & [\% $\uparrow$] & [m $\downarrow$] & [m $\downarrow$] & [$^\circ$ $\downarrow$] \\ \midrule \hline
            \multicolumn{13}{c}{\textit{All spatiotemporal pairs}} \\
            \hline
            FPFH~\cite{rusu2009FPFH} & 22.83 &  3.30  & 0.66 / 3.08 & \textbf{0.20} / 43.21 & 18.73 & 2.53   & 0.80 / 2.43  & \textbf{0.16} / 45.87 & 11.70 &  2.52  & 0.44 / 2.43 & \textbf{0.10} / 45.32  \\
            FCGF~\cite{Choy2019FCGF} & 28.22 &  2.07  & 1.83 / 2.01 & \underline{0.62} / 29.25 & 37.70 & 1.81   & 1.29 / 1.78 & 0.52 / 41.04 & 24.43 & 2.24   & 1.09 / 2.04 & 0.76 / 39.89    \\
            D3Feat~\cite{bai2020d3feat} & 31.77 & 1.98   & \underline{0.08} / 1.95 & 1.44 / \textbf{24.22} & \underline{51.37} & 1.62   & \underline{0.07} / 1.57 & 1.19 / 32.09 & 22.73 & 2.37   & \underline{0.09} / 2.26 & 1.45 / 33.09   \\
            \textsc{Predator}~\cite{predator} & \textbf{55.53} & \textbf{1.09}   & \textbf{0.05} / \textbf{1.08} & 0.98 / \underline{25.05} & \textbf{76.73} &  \textbf{0.77}  & \textbf{0.04} / \textbf{0.68} & 0.74 / \textbf{15.27} & \textbf{64.97} & \textbf{0.71}  & \textbf{0.06} / \textbf{0.65} & 0.79 / \textbf{13.52}   \\
            GeoTransformer~\cite{qin2022geometric} & \underline{38.13} & \underline{1.24}   & 0.14 / \underline{1.28} & 0.64 / 27.90 & 47.78 & \underline{0.98}   & 0.14 / \underline{0.98} & \underline{0.39} / \underline{22.27} & \underline{39.07} &  \underline{0.96}  & 0.14 / \underline{0.99} & \underline{0.41} / \underline{22.93}   \\
            \hline
            \multicolumn{13}{c}{\textit{Only same-stage pairs}} \\
            \hline
            FPFH~\cite{rusu2009FPFH} &32.86 &  2.46  & 1.57/ 2.34 & \textbf{0.28} / 34.41 & 46.40 & 1.94   & 1.12 / 1.90 & \underline{0.38} / 33.42 & 30.82 & 2.58   & 1.13 / 2.42 & \textbf{0.27} / 29.35    \\
            FCGF~\cite{Choy2019FCGF} & 39.32 & 1.88   & 1.78 / 1.84 & 0.55 / 28.01 & 44.65 & 1.77   & 0.98 / 1.76 &  0.41 / 30.47 & 42.86 &  2.24  & 0.56 / 2.23 & {0.44} / 32.12    \\
            D3Feat~\cite{bai2020d3feat} & 43.62 & 1.91   & \underline{0.08} / 1.93 & 1.31 / 24.05 & \underline{58.47} & 1.48   & \underline{0.07} / 1.48 & 1.10 / 28.39 & 36.51 & 2.09   & \underline{0.08} / 2.05 & 1.36 / 27.22   \\
            \textsc{Predator}~\cite{predator} & \textbf{76.80} &  \textbf{0.81}  & \textbf{0.05} / \textbf{0.83} & 0.86 / \textbf{18.41} & \textbf{87.49} & \textbf{0.44}   & \textbf{0.04} / \textbf{0.48} & 0.69 / \textbf{9.89} & \textbf{92.99} &  \textbf{0.27}  & \textbf{0.04} / \textbf{0.27} & 0.67 / \textbf{4.83}   \\
            GeoTransformer~\cite{qin2022geometric} & \underline{50.88} & \underline{1.07}   & 0.13 / \underline{1.13} & \underline{0.54} / \underline{23.73} & 54.07 & \underline{0.79}   & 0.14 / \underline{0.83} & \textbf{0.37} / \underline{17.26} & \underline{55.59} & \underline{0.69}   & 0.14 / \underline{0.73} & \underline{0.35} / \underline{17.02}  \\
            \hline
            \multicolumn{13}{c}{\textit{Only different-stage pairs}} \\
            \hline
            FPFH~\cite{rusu2009FPFH} & 1.06 &  4.88  & \textbf{0.07} / 4.32 & \textbf{0.03} / 65.89 & 0.82 & 4.23   & \textbf{0.09} / 4.06 & \textbf{0.02} / 72.43 & 0.42 &  4.21  & 0.03 / 4.06  & \textbf{0.00} / 78.01  \\
            FCGF~\cite{Choy2019FCGF} & 5.21 &  3.22  & 2.13 / 3.21 & 2.17 / 45.61 & {14.06} & 4.15   & 2.40 / 4.02 & \underline{0.93} / 62.15 & 10.52 &   3.28 & 2.75 / 3.23 & 1.74 / 53.24   \\
            D3Feat~\cite{bai2020d3feat} & 6.12 & 2.01 & 0.16 / 2.01 & 3.48 / \textbf{24.57} & 12.85 & 2.40   & 0.13 / 2.03 & 3.56 / 52.18 & 4.76 &  2.75  & \textbf{0.12} / 2.53 & 2.43 / 40.76   \\
            \textsc{Predator}~\cite{predator} & \underline{9.49} &  \underline{1.71}  & 0.16 / \underline{1.62} & 3.08 / 39.42 & \textbf{18.42} &  \textbf{2.03}  & \underline{0.10} / \textbf{1.77} & 2.08 / \textbf{44.46} & \textbf{28.42} & \textbf{1.28}   & \underline{0.13} / \textbf{1.16} & 1.29 / \textbf{24.85}   \\
            GeoTransformer~\cite{qin2022geometric} & \textbf{10.55} &  \textbf{1.62}  & \underline{0.15} / \textbf{1.59} & \underline{1.63} / \underline{36.91} & \underline{13.39} & \underline{2.25}   & 0.16 / \underline{1.81} & 0.96 / \underline{49.66} & \underline{17.51} & \underline{1.31}   & \underline{0.13} / \underline{1.34} & \underline{0.66} / \underline{30.62}   \\
            \bottomrule
    \end{tabular}
    }
    \caption{\textbf{Pairwise registration results of existing 3D point cloud registration methods on \project{}}. We report registration recall (Recall) and translation (TE) and rotation (RE) errors. For TE and RE, we report the average measurements among: [successfully registered pairs] / [all pairs]. The first value is the standard evaluation setting.}
    \label{tab:pairwise}
\end{table*}

\paragraph{Evaluation Metrics} 
To evaluate both registration tasks, we follow the same evaluation metrics as in the spatial registration domain~\cite{predator} and use: registration recall (Recall), relative translation error (RTE), and relative rotation error (RRE). For the RTE and RRE metrics, their formal definitions are, 
\begin{equation}
    \text{RRE} = \angle(\mathbf{R}_{GT}^{-1} \hat{\mathbf{R}}), \quad \text{RTE} = \| \mathbf{v}_{GT} - \hat{\mathbf{v}} \|_2,
\end{equation}
where $\{\mathbf{R}_{GT}, \mathbf{v}_{GT}\}$ and $\{\hat{\mathbf{R}}, \hat{\mathbf{v}}\}$ denote the groundtruth and estimated rigid transformation, respectively. Here,
\begin{equation}
    \angle(X) = \arccos\left(\frac{\text{trace}(X) - 1}{2}\right),
\end{equation} returns the angle of rotation matrix $X$ in degrees. 

The registration recall is defined as the ratio of the number of successfully registered point cloud pairs to the total number of point cloud pairs. In our benchmark, a pair is considered successfully registered if it satisfies two criteria: the relative rotation error (RRE) is less than 10 degrees, and the relative translation error (RTE) is less than 0.2 meters. The thresholds are common values for indoor point cloud registration~\cite{zeng20163dmatch}. This metric provides a comprehensive measure of the registration algorithm's accuracy with varying degrees of spatial displacement.

To measure the performance of methods in spatiotemporal registration, we employ overlap ratio and temporal change ratio as ablation metrics. Please refer to Section~\ref{sec:pairwise_dataset} for their definitions.

\section{Experiments}
\label{sec:experiments}
In the pairwise registration task of the Nothing Stands Still benchmark, we evaluate state-of-the-art approaches that include both hand-crafted and learned features: FPFH~\cite{rusu2009FPFH}, D3Feat~\cite{bai2020d3feat}, FCGF~\cite{Choy2019FCGF}, \textsc{Predator}~\cite{predator}, and GeoTransformer~\cite{qin2022geometric}\footnote{We follow the original training protocol per method, and integrate all the evaluated methods in our point cloud registration codebase. The codebase is open-sourced together with the benchmark.}. In the multi-way registration task, we evaluate the state-of-the-art methods in \cite{Choi2015robust} and \cite{yew2021learning}, which we initialize with the registration results of the two best performing methods on the pairwise registration task.

\subsection{Pairwise Spatiotemporal Registration}
\label{sec:pairwise_results}
We report the results in Table~\ref{tab:pairwise}. Overall, \textsc{Predator} performs the best in the majority of the metrics for all data splits and temporal ablations. However, different-stage pairs pose significant challenges for all registration methods. The average performance drops between same-stage and different-stage pairs over all methods by $40.8$ p.p., $41.5$ p.p., $37.4$ p.p. for the cross-area, cross-stage, and original splits respectively. We can also observe that learning methods achieve better performance than the hand-engineered FPFH, especially for temporal registration. FPFH only successfully registers about 1\% of the different-stage pairs, significantly lagging behind other learning-based methods. Although FPFH's RE for successfully registered pairs is very low, the low registration success rate suggests that these results are unclear. Indeed, when computing the RE over all pairs in the dataset, we see that the orders are higher in magnitude. Hence, RE calculation on only successfully registered pairs, does not fully showcase the robustness of a method. We also observe that the modeling of interactions between \onepc{s} of different stages may play a key role in temporal registration. For example, compared to D3Feat, \textsc{Predator} and GeoTransformer show a large margin of 23.7 and 12.8 p.p. in different-stage pairs in the Original split. We hypothesize that the attention mechanism they utilize between the inputs enables them to capture more temporal-related patterns, while other methods treat the \onepc{} input pairs independently. 

\begin{figure*}[h!]
    \centering
    \includegraphics[width=\linewidth]{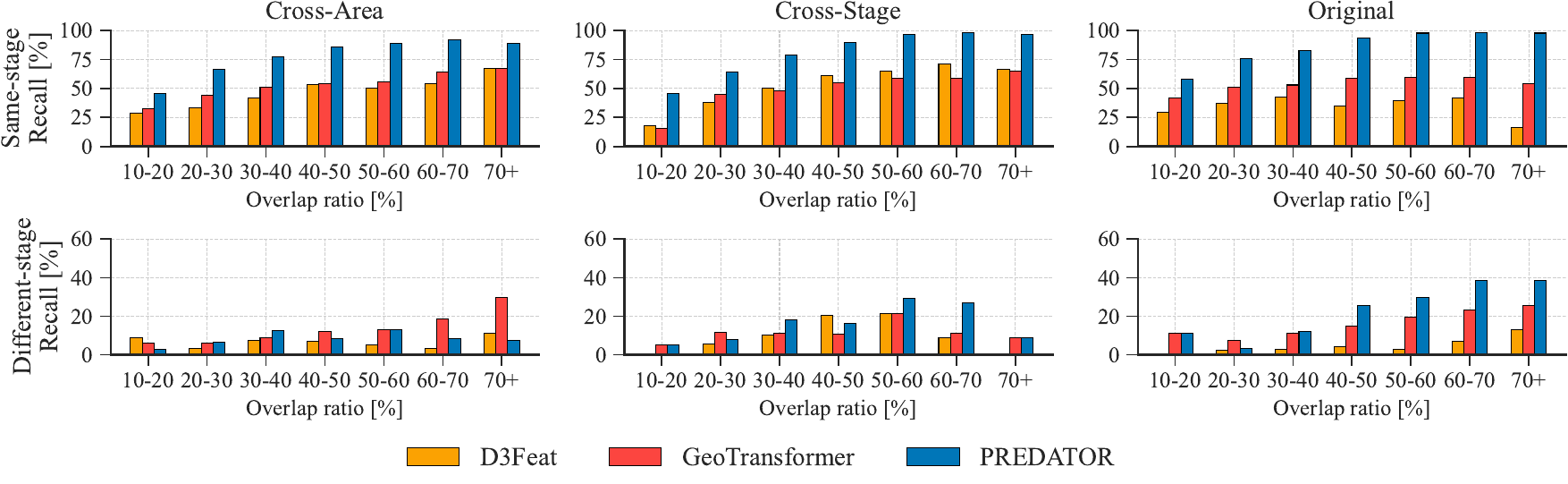}
    \caption{\textbf{Registration recall [\%] per overlap ratio (OR) bin for existing 3D point cloud registration methods.} A clear performance gap is visible between same-stage pairs (top row) and different-stage pairs (bottom row) for these methods.}
    \label{fig:recall_per_bin}
\end{figure*}
\begin{figure*}[h!]
    \centering
    \includegraphics[width=\linewidth]{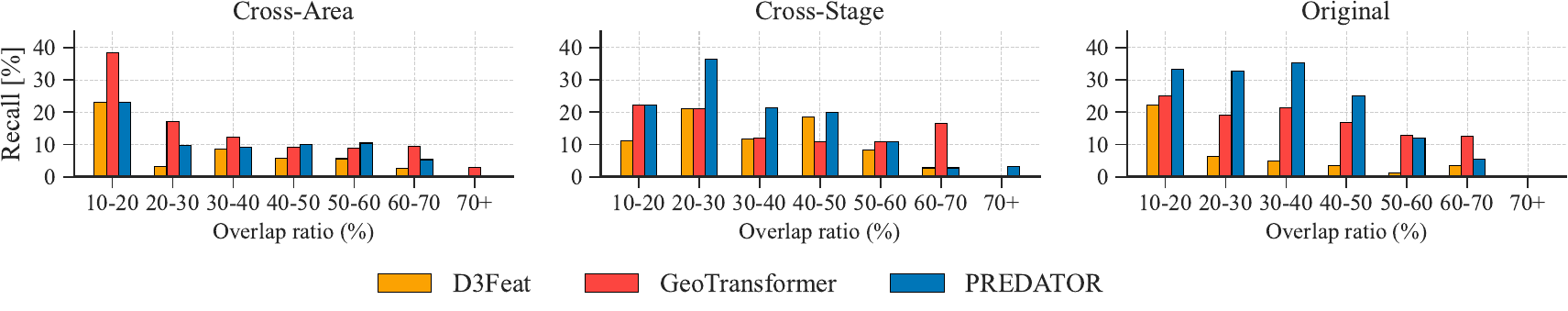}
    \caption{\textbf{Registration recall [\%] per temporal change ratio (TCR) bin for existing 3D point cloud registration methods on different-stage pairs.} It is evident that larger temporal changes pose greater challenges for these methods, and that the cross-area split is a setting with increased challenges.
    \label{fig:recall_per_tcr_bin}}
\end{figure*}

When comparing the results of different data splits, we notice that methods perform the best on the cross-stage split and worst on the cross-area split. This behavior is expected for both cases. In the cross-stage split, methods learn the general structure and characteristics of the area during training and are able to make predictions on unseen stages more accurately. This has practical applications in industries such as construction or building management, where a small initial annotation effort can lead to significant future gains. In the cross-area split, methods struggle to generalize to unseen environments, which is a common challenge in various computer vision tasks.
As mentioned above, the registration of fragment pairs from different stages poses difficulties, which is further emphasized in the cross-area split.

Figure~\ref{fig:recall_per_bin} provides a histogram of registration recall for all data splits based on the overlap ratio of \onepc{} pairs. The three best-performing methods in the pairwise registration task, namely D3Feat, GeoTransformer, and \textsc{Predator}, are included in the evaluation. The results show a clear trend where higher overlap ratios correspond to higher recall values across all splits.
Figure~\ref{fig:recall_per_tcr_bin} presents the histogram analysis based on the temporal change ratio. It demonstrates that as the temporal change increases, the registration problem becomes more difficult. Among the three dataset splits, the cross-area split exhibits the least robustness against large temporal changes. It is noteworthy that these methods perform exceptionally well in the prominent spatial registration benchmarks of 3DMatch~\cite{zeng20163dmatch} and 3DLoMatch~\cite{predator}, achieving high accuracy rates in the range of 80\% to 90\% (Table~\ref{tab:3Dmatchcomparison}). However, their performance drops by around 50\% in the \abbrev{} benchmark. 
While the difference in performance between 3DLoMatch and \abbrev{} is less pronounced for pairs with low overlap (10-30\%), for different-stage pairs, regardless of the overlap percentage, it is significantly lower.

\begin{table}[t!]
    \centering
    \resizebox{\columnwidth}{!}{
    \footnotesize
    \begin{tabular}{lccc}
    \hline
    \textbf{Benchmark}  & \textbf{D3Feat} & \textbf{PREDATOR} & \textbf{GeoTrans.}\\ \hline\hline
    \multicolumn{4}{c}{\textit{Standard Overlap [30\%+]}}\\ \hline
    3DMatch~\cite{zeng20163dmatch}               & \textbf{82.2} & \textbf{89.0} & \textbf{92.0} \\ 
    \textbf{NSS} \textit{(all)}                  & 34.1 & 58.8 & 40.3 \\
    \textbf{NSS} \textit{(same-stage only)}      & \underline{47.9} & \underline{83.1} & \underline{54.3} \\
    \textbf{NSS} \textit{(different-stage only)} &  6.6 & 10.6 & 12.5 \\ \hline
    \multicolumn{4}{c}{\textit{Low Overlap [10-30\%]}}\\ \hline
    3DLoMatch~\cite{predator}                    & \textbf{37.2} & \underline{59.8} & \textbf{75.0} \\
    \textbf{NSS} \textit{(all)}                  & 25.2 & 47.6 & 32.6 \\
    \textbf{NSS} \textit{(same-stage only)}      & \underline{32.6} & \textbf{62.4} & \underline{42.2} \\
    \textbf{NSS} \textit{(different-stage only)} &  4.4 &  5.9 & 5.9  \\ \hline
    \end{tabular}}
    \caption{\textbf{Comparison of performance on \abbrev{} (original split) with that on 3DMatch and 3DLoMatch.} We compare the registration recall of the three best-performing methods on \abbrev{} and clearly observe that their results on the the standard spatial registration benchmarks are substantially higher on ours.}
    \label{tab:3Dmatchcomparison}
\end{table}

\begin{table*}[hbt!]
    \centering
    \footnotesize
    \begin{tabular}{p{4cm}ccccccccc}
            \hline
            \multirow{2}{*}{\textbf{Method}} & \multicolumn{3}{c}{\textbf{Cross-Area}} & \multicolumn{3}{c}{\textbf{Cross-Stage}} & \multicolumn{3}{c}{\textbf{Original}} \\
            \cmidrule(lr){2-4} \cmidrule(lr){5-7} \cmidrule(lr){8-10}
             &  {All} & {Same-only} & {Diff} &  {All} & {Same-only} & {Diff} & {All} & {Same-only} & {Diff}  \\ \hline \hline
            \multicolumn{10}{c}{\textit{All testing pairs}} \\ \hline
            D3Feat~\cite{bai2020d3feat} & 31.77 & \textbf{49.53} & \red{$-$17.76} & 51.37 & \textbf{66.07} & \red{$-$14.70} & 22.73 & \textbf{46.60} & \red{$-$23.87}   \\
            \textsc{Predator}~\cite{predator} & 55.90 & \textbf{58.50} & \red{$-$2.60} & 76.63 & \textbf{77.60} & \red{$-$0.97} & \textbf{64.67} & 62.77 & \green{$+$1.90}  \\
            GeoTransformer~\cite{qin2022geometric} & 38.30 & \textbf{38.87} & \red{$-$0.57} & 47.24 & \textbf{47.38} & \red{$-$0.14} & \textbf{40.37} & 36.08 & \green{$+$4.29} \\ \hline
            \multicolumn{10}{c}{\textit{Same-stage testing pairs}} \\ \hline
            D3Feat~\cite{bai2020d3feat} & 43.62 & \textbf{67.06} & \red{$-$23.44} & 58.47 & \textbf{74.62} & \red{$-$16.15} & 36.51 & \textbf{70.38} & \red{$-$33.87}   \\
            \textsc{Predator}~\cite{predator} & 76.80 & \textbf{80.51} & \red{$-$3.71} & 87.49 & \textbf{88.59} & \red{$-$1.10} & 92.99 & \textbf{93.23} & \red{$-$0.24}   \\
            GeoTransformer~\cite{qin2022geometric} & 50.88 & \textbf{51.71} & \red{$-$0.83} & \textbf{54.07} & \textbf{54.07} & \green{0.00} & 55.59 & \textbf{56.24} & \red{$-$0.65}   \\
            \hline
            \multicolumn{10}{c}{\textit{Different-stage testing pairs}} \\ \hline
            D3Feat~\cite{bai2020d3feat} & 6.12 & \textbf{11.60} & \red{$-$5.48} & 12.85 & \textbf{19.70} & \red{$-$6.85} & 4.76 & \textbf{15.59} & \red{$-$10.83}   \\
            \textsc{Predator}~\cite{predator} & 9.49 & \textbf{10.86} & \red{$-$1.37} & \textbf{18.42} & 17.99 & \green{$+$0.43} & \textbf{28.42} & 23.04 & \green{$+$5.38}   \\
            GeoTransformer~\cite{qin2022geometric} & \textbf{10.55} & 9.08 & \green{$+$1.47} & \textbf{13.39} & 10.80 & \green{$+$2.59} & \textbf{17.51} & 9.76 & \green{$+$7.75}   \\ \hline
    \end{tabular}
    \caption{\textbf{Effect of training with temporal data on registration recall [\%].} Methods are trained using either \textit{all} training data or \textit{same-only} stage pairs. Testing is evalauted on all pairs. Values in \red{red} denote a drop in performance, whereas values in \green{green} an increase.}
    \label{tab:same_stage_only}
\end{table*}

\begin{figure*}[h!]
    \centering
    \includegraphics[trim=2 0 2 0,clip,width=0.95\linewidth]{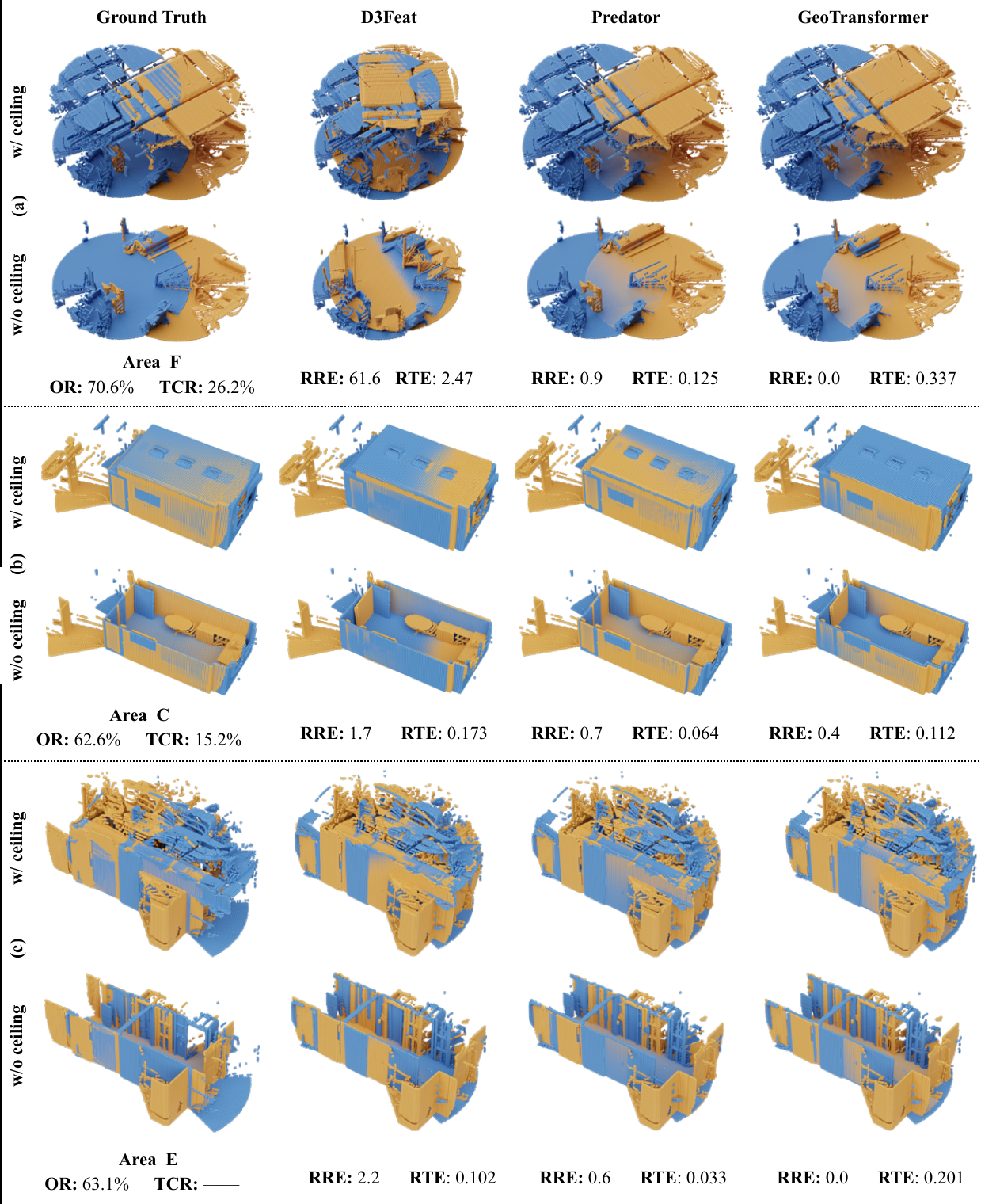}
    \caption{\textbf{Qualitative results on pairwise registration.} Results on a \onepc{} pair are showcased for D3Feat~\cite{bai2020d3feat}, \textsc{Predator}~\cite{predator}, and GeoTransformer~\cite{qin2022geometric}. The overlap (OR) and temporal change (TCR) ratios are reported per input pair, as well as the translation (TE) and rotation (RE) errors per method (in meters and degrees respectively).}
    \label{fig:pairwise_resultsa}
\end{figure*}

\begin{figure*}[h!]
    \centering
    \includegraphics[width=0.93\linewidth]{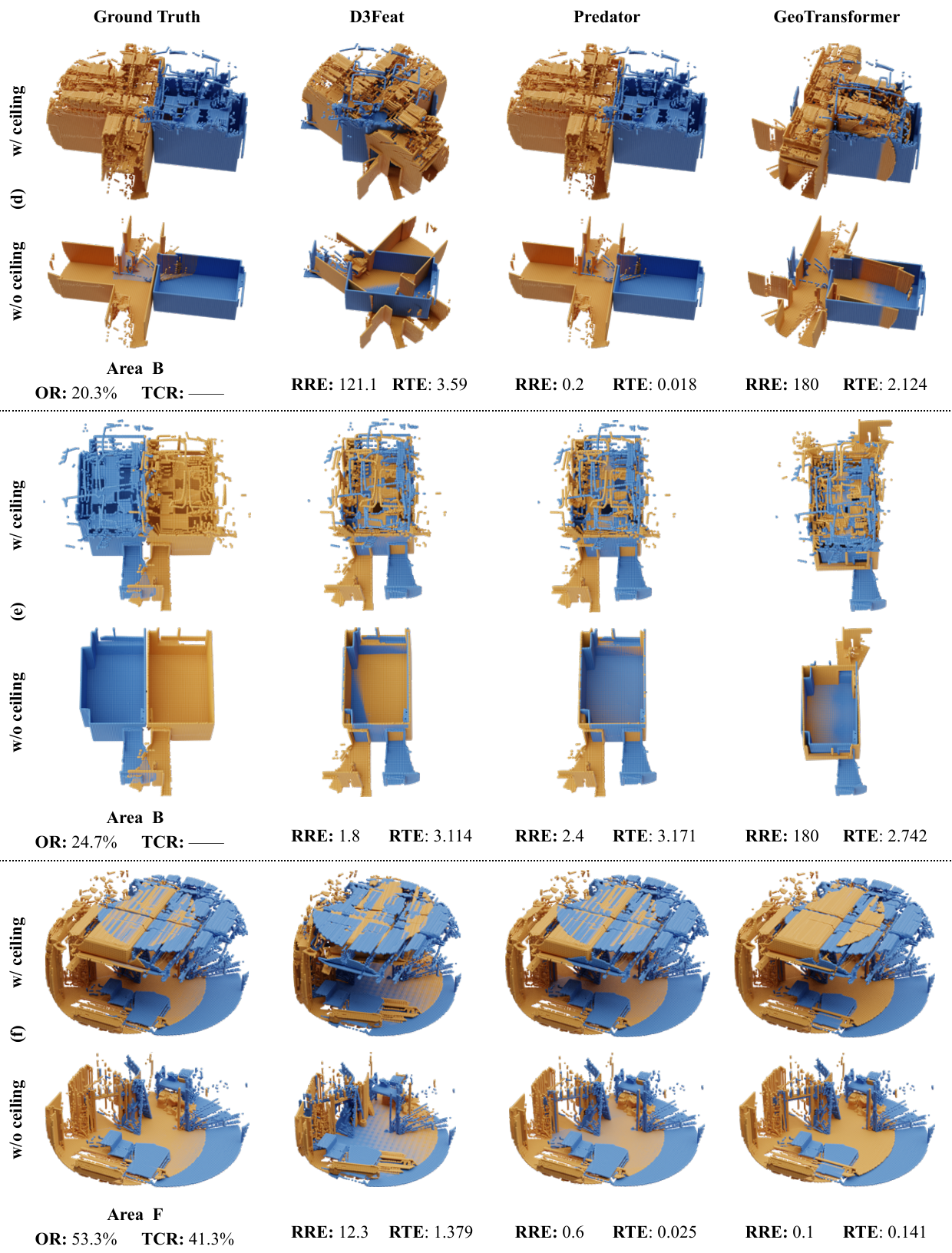}
    \caption{\textbf{Additional qualitative results on pairwise registration.} Results on a \onepc{} pair are showcased for D3Feat~\cite{bai2020d3feat}, \textsc{Predator}~\cite{predator}, and GeoTransformer~\cite{qin2022geometric}. The overlap (OR) and temporal change (TCR) ratios are reported per input pair, as well as the translation (TE) and rotation (RE) errors per method (in meters and degrees respectively).}
    \label{fig:pairwise_resultsb}
\end{figure*}

Figures~\ref{fig:pairwise_resultsa} and \ref{fig:pairwise_resultsb} provide example results of spatiotemporal pairwise registration for D3Feat, GeoTransformer, and \textsc{Predator}. Consistent with the quantitative results, \textsc{Predator} demonstrates more accurate registration compared to the other methods. In cases where the overlap ratio is very high and the temporal changes have minimal impact on the main structure of the scene (row $b$) or do not exist (row $c$), all three methods achieve similarly good results, which is an expected behavior. However, there are scenarios where D3Feat struggles to register pairs correctly, even with a high overlap ratio. This is particularly evident in cases where there are significant temporal changes and the scene geometry contains repetitive elements, such as studs (rows $a$ and $f$). This limitation is attributed to D3Feat's reliance on local geometry constraints and independent \onepc{} processing. In a scenario with low overlap and no temporal change (row $d$), both D3Feat and GeoTransformer fail to find the correct alignment, while \textsc{Predator} performs better. Lastly, in row $e$, all three methods encounter a failure case. Here, not only is the overlap low, but the two rooms are closely similar, making it a challenging scenario to solve. The main differences are the cut-out corners in the blue-colored fragment and the mirrored location of the doors.

\subsubsection{Effect of Temporal Data}
To further investigate the impact of training on both same-stage and different-stage pairs on the registration recall of D3Feat, GeoTransformer, and \textsc{Predator}, we conducted the following experiment:  we only trained on the same-stage pairs available in different data splits. The evaluation was still performed on the entire test set, which includes different-stage pairs. The results in Table~\ref{tab:same_stage_only} indicate that the presence of different-stage pairs hinders the training process for some methods. When trained exclusively with same-stage data, D3Feat demonstrates significantly better performance during testing. We hypothesize that this improvement is due to D3Feat relying on the local geometric assumption that similar local geometric structures are expected to be registered together. As expected, the recall for all methods and splits in same-stage registration is higher when trained solely on same-stage pairs. This suggests that the methods struggle to effectively distinguish between the spatial and temporal characteristics of the data, thereby affecting same-stage registration due to \textit{temporal noise}. However, in the case of different-stage registration, the methods benefit from the presence of such pairs in training, indicating that some learning is occurring. Figure~\ref{fig:recall_per_bin} illustrates that these benefits primarily stem from the highly overlapping \onepc{} pairs. Despite the advantages of utilizing all spatiotemporal data during training, there is still significant room for the methods to fully exploit the potential of different-stage pairs.

\begin{table*}[h!]
    \centering
    \footnotesize
    \begin{tabular}{p{3cm}cccccc}
            \hline
            \multirow{2}{*}{\textbf{Method}} & \multicolumn{3}{c}{\textbf{Cross-Stage}} & \multicolumn{3}{c}{\textbf{Original}} \\
            \cmidrule(lr){2-4} \cmidrule(lr){5-7}
             & \textbf{Recall} [\% $\uparrow$] & \textbf{TE}  [m $\downarrow$] & \textbf{RE}  [$^\circ$ $\downarrow$] & \textbf{Recall} [\% $\uparrow$] & \textbf{TE}  [m $\downarrow$] & \textbf{RE}  [$^\circ$ $\downarrow$] \\ \hline \hline
            \multicolumn{7}{c}{\textit{All testing pairs}} \\
            \hline
            D3Feat~\cite{bai2020d3feat} & 41.76 & \textbf{0.04} / \underline{0.68} & \underline{2.46} / 48.94 & 45.26 & \textbf{0.04} / 0.63 & \underline{2.39} / 44.35   \\
            \textsc{Predator}~\cite{predator} & \textbf{64.17} & \underline{0.05} / \textbf{0.45} & \textbf{2.40} / \textbf{27.86} & \textbf{73.70} & \textbf{0.04} / \textbf{0.34} & \textbf{2.11} / \textbf{19.62}   \\
            GeoTransformer~\cite{qin2022geometric} & \underline{46.04} & .09 / 0.70 & 4.10 / \underline{44.83} & \underline{47.21} & \underline{0.09} / \underline{0.61} & 4.16 / \underline{38.95} \\
            
            \hline
            \multicolumn{7}{c}{\textit{Same-stage testing pairs}} \\
            \hline
            D3Feat~\cite{bai2020d3feat} & \underline{66.33} & \underline{0.03} / \underline{0.51} & \underline{1.57} / \underline{35.21} & \underline{67.40} & \textbf{0.02} / \underline{0.50} & \textbf{1.24} / \underline{32.56}   \\
            \textsc{Predator}~\cite{predator} & \textbf{75.94} & \textbf{0.02} / \textbf{0.37} & \textbf{1.42} / \textbf{21.91} & \textbf{81.18} & \textbf{0.02} / \textbf{0.29} & \underline{1.40} / \textbf{15.93}   \\
            GeoTransformer~\cite{qin2022geometric} & 60.97 & 0.08 / 0.69 & 3.90 / 42.68  & 1.71 & \underline{0.08} / 0.66 & 4.02 / 40.69 \\
            
            \hline
            \multicolumn{7}{c}{\textit{Different-stage testing pairs}} \\
            \hline
            D3Feat~\cite{bai2020d3feat} & 21.68 & \underline{0.08} / 0.81 & 4.70 / 60.17 & 27.39 & \underline{0.09} / 0.75 & 4.67 / 53.87   \\
            \textsc{Predator}~\cite{predator} & \textbf{54.55} & \textbf{0.07} / \textbf{0.52} & \textbf{3.53} / \textbf{32.73} & \textbf{67.67} & \textbf{0.06 }/ \textbf{0.37} & \textbf{2.79} / \textbf{22.60}  \\
            GeoTransformer~\cite{qin2022geometric} & \underline{33.84} & 0.10 / \underline{0.71} & \underline{4.32} / \underline{46.58} & \underline{35.51} & 0.10 / \underline{0.58} & \underline{4.28} / \underline{37.55}  \\
            \hline
    \end{tabular}
    
    \caption{\textbf{Registration performance on RIO10 dataset~\cite{wald2020beyond}.} We follow the same data generation and evaluation protocol as in \abbrev{} and report registration recall (Recall) and translation (TE) and rotation (RE) errors. For TE and RE, we report the average measurements among: [successfully registered pairs] / [all pairs]. The first value is the standard evaluation setting.}
    \label{tab:baseline}
\end{table*}

\begin{figure*}[t!]
\centering
\includegraphics[width=1\linewidth,keepaspectratio]{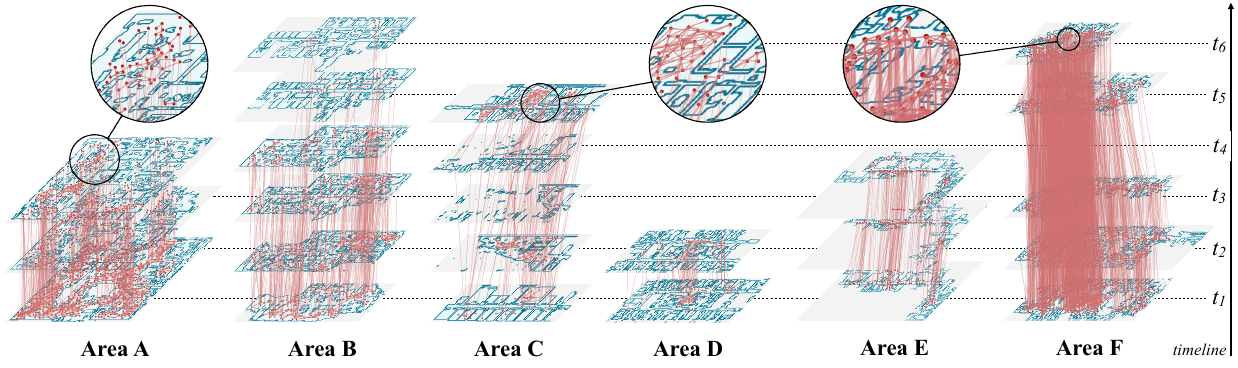}
    \caption{\textbf{Spatiotemporal graphs for the multi-way registration task.} Nodes represent \onepc{} locations and edges denote that these pairs are overlapping in the 3D spatiotemporal map of each area. The resulting graphs are dense both in nodes and edges.}
    \label{fig:posegraph}
\end{figure*}

\subsubsection{Comparison to RIO10 Spatiotemporal Dataset}
RIO10~\cite{wald2020beyond} is a recent indoor point cloud dataset specifically designed for camera re-localization tasks in changing environments. We follow the \textbf{\abbrev{}} benchmark protocol to evaluate the spatiotemporal registration performance for D3Feat, \textsc{Predator}, and GeoTransformer on this dataset. We consider only the \textit{original} and \textit{cross-stage} splits, since, due to the spatial scale of each area, the cross-area and original splits overlap. The results are presented in Table~\ref{tab:baseline}. In comparison to the performance on the \abbrev{} dataset (refer to Table~\ref{tab:pairwise}), we observe better registration performance across all metrics on average when evaluating on RIO10. Particularly, the drop in performance for different-stage pair registration is significantly smaller. This suggests that the domain gap between same-stage and different-stage pairs in this dataset is not as drastic, which is expected considering the smaller changes depicted in the scenes.

\subsection{Multi-way Spatiotemporal Registration}
\label{sec:multiway_results}
Both Choi et al. \cite{Choi2015robust} and PoseGraphNet \cite{yew2021learning} require the construction of a pose graph, where nodes represent \onepc{s} and edges denote the predicted transformation between two connected \onepc{s}. We define the pose graph including the \onepc{s} from all stages within an area. In both cases, the edges are initialized with the relative transformations from the pairwise registration results for the two best-performing methods, \textsc{Predator} and GeoTransformer. Suppose no edge exists between two \onepc{s} in the graph, it implies that they either do not overlap or overlap by less than 10\%, which is considered insufficient for meaningful registration for current methods. Figure~\ref{fig:posegraph} displays the dataset pose graphs for all areas in the original split, which comprise numerous nodes and edges, thereby making the optimization process for finding a globally optimal alignment challenging.

\begin{table*}[h!]
\resizebox{\textwidth}{!}{
\footnotesize
    \setlength\tabcolsep{3pt}
    \setlength\extrarowheight{1pt}
\begin{tabular}{lc|cccc|ccccc|cccc}
\toprule
\multirow{3}{*}{\textbf{Method}} & \multicolumn{4}{c}{\textbf{Pairwise Outputs}} & \multicolumn{5}{c}{\textbf{Choi~\cite{Choi2015robust}}} &  \multicolumn{5}{c}{\textbf{PoseGraphNet~\cite{yew2021learning}}} \\ 
\cmidrule(lr){2-5}  \cmidrule(lr){6-10}  \cmidrule(lr){11-15}
             & \textbf{RMSE}$^\text{P}$  & \textbf{Recall}$^\text{G}$ & \textbf{TE}$^\text{G}$ & \textbf{RE}$^\text{G}$  & \textbf{RMSE}$^\text{P}$ & \textbf{RMSE}$^\text{G}$ & \textbf{Recall}$^\text{G}$ & \textbf{TE}$^\text{G}$ & \textbf{RE}$^\text{G}$ & \textbf{RMSE}$^\text{P}$ & \textbf{RMSE}$^\text{G}$ & \textbf{Recall}$^\text{G}$ & \textbf{TE}$^\text{G}$ & \textbf{RE}$^\text{G}$ \\ 
             & [m $\downarrow$] & [\% $\uparrow$] & [m $\downarrow$] & [$^\circ$ $\downarrow$] & [m $\downarrow$] & [m $\downarrow$] & [\% $\uparrow$] &  [m $\downarrow$] &[$^\circ$ $\downarrow$] & [m $\downarrow$] & [m $\downarrow$] & [\% $\uparrow$] & [m $\downarrow$] & [$^\circ$ $\downarrow$] \\ \hline \hline
\multicolumn{15}{c}{\textit{Cross-Area}} \\ 
\hline
\textsc{Predator}~\cite{predator}               & \textbf{1.21} & 56.23   & \textbf{0.05} / \textbf{1.04} & 0.95 / 24.89 & 1.91 &  33.43 & 53.61 & \textbf{0.05} / 1.80 & 0.69 / 21.67 & 1.43 & 28.29 & \textbf{77.28} & \textbf{0.05} / 1.11 & \textbf{0.13} / \textbf{0.44}   \\
GeoTransformer~\cite{qin2022geometric} & 1.33 & 38.52   & 0.14 / 1.26 & 1.18 / 26.64 & 2.26 & 39.90 & 42.51 & 0.11 / 2.13 & 0.83 / 24.73 & 1.64 & \textbf{27.63} & 53.31 & 0.06 / 1.28 & 0.16 / 0.57    \\ \hline  
\multicolumn{15}{c}{\textit{Cross-Stage}} \\ \hline    
\textsc{Predator}~\cite{predator}               & \textbf{0.88} & 75.85 & 0.04 / \textbf{0.71} & 0.75 / 15.90 & 1.48 & 19.68 & 71.07 & \textbf{0.03} / 1.37 & 0.41 / 14.86 & 1.04 & \textbf{18.45} & \textbf{79.94}  & 0.05 / 0.79 & 0.08 / 0.24      \\
GeoTransformer~\cite{qin2022geometric} & 1.05 & 46.62 & 0.14 / 0.99 & 0.97 / 22.55 & 1.74 & 28.19 & 56.90 & 0.10 / 1.64 & 0.63 / 19.84 & 1.13 & 20.93 & 62.93  & 0.04 / 0.88 & \textbf{0.07} / \textbf{0.23}   \\  \hline
\multicolumn{15}{c}{\textit{Original}} \\ \hline    
\textsc{Predator}~\cite{predator}               & 0.83 & 66.65 & 0.06 / 0.65 & 0.81 / 13.57 & 1.84 & 20.72 & 62.35 &  0.06 / 1.73 & 0.69 / 18.24 & \textbf{0.53}  & 16.14 & \textbf{82.34} & \textbf{0.05} / \textbf{0.43} & 0.08 / \textbf{0.37}   \\
GeoTransformer~\cite{qin2022geometric} & 1.23 & 40.23 & 0.14 / 1.00 & 0.97 / 24.00 & 2.83 & 27.71 & 40.49 & 0.12 / 2.54 & 1.04 / 30.58 & 1.08 & \textbf{14.55} & 65.35  & 0.09 / 0.84  &  \textbf{0.07} / 0.60 \\ \bottomrule
\end{tabular}
}
\caption{\textbf{Multi-way registration results of existing 3D optimization methods on \project{}.} We report pairwise (RMSE$^P$) and global registration metrics (RMSE$^G$, Recall, TE, RE) on the testing pairs of this task and compare with the pairwise registration results per split since they correspond to the performance \textit{before} the multi-way pose graph optimization. Best values per metric and split are highlighted in \textbf{bold}.}
\label{tab:multiway2}
\end{table*}

In Choi et al.~\cite{Choi2015robust}, the authors distinguish between odometry and uncertainty edges. They acquire this distinction directly from the input data, which is an RGBD video sequence. Since this is not applicable to our case, we select odometry edges by constructing a minimum spanning tree from a randomly chosen node. We experimented with various edge selection methods but did not observe significant differences in the results. Next, we initialize the edge weights using the predicted average matchability scores from the two best-performing pairwise registration methods (GeoTransformer and \textsc{Predator}). We also select all non-temporal pairs as constrained pairs, \ie they will not be pruned during optimization. In each iteration, edges with weights below a threshold are considered non-valid and are pruned from the graph. For further details on the pose graph optimization, we refer the reader to~\cite{Choi2015robust}. The objective here is to achieve a set of consistent registrations that minimize the weighted average root mean square error (RMSE) of all nodes in the pose graph. 

PoseGraphNet removes the reliance on odometry data. It employs a GNN to learn to perform transformation synchronization. We train their method on the training set of the multi-way registration task with the objective of minimizing absolute rotation and translation errors. 
Unlike the pose graph setup for Choi et al.~\cite{Choi2015robust}, which initializes nodes with odometry information, PoseGraphNet starts with identity matrices for the nodes, and then refines their poses through incremental updates. All weights in this GNN are learned during training, ensuring robustness against outliers and noisy data. 
During inference, the trained model applies this learned synchronization recurrently for multi-way registration.

Table~\ref{tab:multiway2} presents the results for both methods with respect to global (RMSE$^G$, Recall$^G$, TE$^G$, and RE$^G$) and pairwise (RMSE$^P$) registration metrics for all dataset splits. We also compare to the pairwise registration outputs since they correspond to the performance \textit{before} the multi-way pose graph optimization. Before analyzing them, it is important to note the reasons that the \textit{pairwise} outputs showcase differences with respect to Table~\ref{tab:pairwise}. Firstly, different ground truth poses are used in the two tasks. As mentioned in Section~\ref{sec:dataset}, the discrepancy arises from the distinction in global and local alignment for \onepc{s} and \allpc{s}. Secondly, the multi-way registration test set is a subset of the pairwise one, as not all \onepc{} pairs could be connected into a larger pose graph. 

For both methods, the errors of global RMSE (RMSE$^G$) are a magnitude higher than in pairwise RMSE (RMSE$^P$), showcasing the complexity of the multi-way problem without excluding that a pairwise-to-global approach might not be the most effective one. Consistent to the pairwise evaluation trends: (i) the cross-area split is the hardest here too; (ii) evaluating TE$^G$ and RE$^G$ only on successfully registered pairs is not a sufficient indication of performance; and (iii) overall \textsc{Predator} leads to improved performance than GeoTransformer. In comparison to Choi et al., PoseGraphNet performs the best overall in global metrics. This can be attributed to (i) using learning and (ii) having an objective that is closer to registration recall than weighted RMSE minimization. Particularly in cases where the pairwise registration results are low, such as in the cross-area split or in the case of GeoTransformer, PoseGraphNet brings the most benefits. We note that even though the gap between \textsc{Predator} and GeoTransfomer decreases after multi-way registration, the difference is still substantial between the two methods, especially on registration recall. However, there is a connection between PoseGraphNet performance and the size of the pose graphs. As shown in Figure \ref{fig:pr-curve}, its performance is fine for graphs with fewer than 30 nodes but deteriorates with very large graphs. This is aligned with expectations since PoseGraphNet was primarily developed for datasets with smaller graphs (e.g., most scenes in ScanNet~\cite{dai2017scannet} have fewer than 30 nodes).

\begin{figure}[hbt]
    \centering
    \includegraphics[width=0.85\columnwidth]{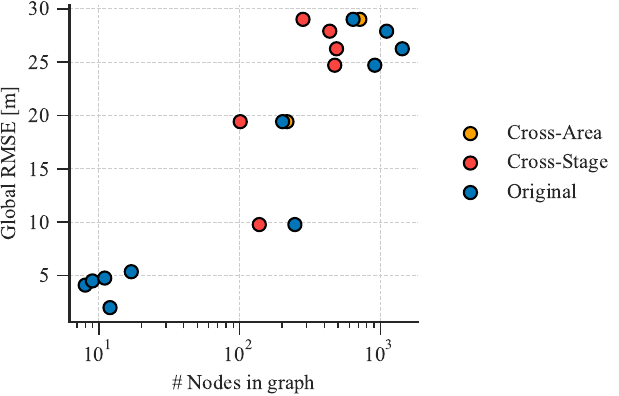}
    \caption{\textbf{The measured Global RMSE of the multiway spatiotemporal registration using PoseGraphNet.}}
    \label{fig:pr-curve}
\end{figure}

It is important to consider the challenge of selecting constrained pairs in Choi et al. for our specific setting. The current selection process results in a high number of non-valid edges in our complex spatiotemporal registration scenarios. Specifically, the number of valid edges per split is: \textit{Cross-Area}: 32.03\%; \textit{Cross-Stage}: 26.48\%; \textit{Original}: 28.30\%. The cross-area split is less affected by temporal changes and consequently yields a higher number of valid edges in this constrained optimization task.

\begin{figure*}[hbt]
    \centering
\includegraphics[width=0.99
\linewidth]{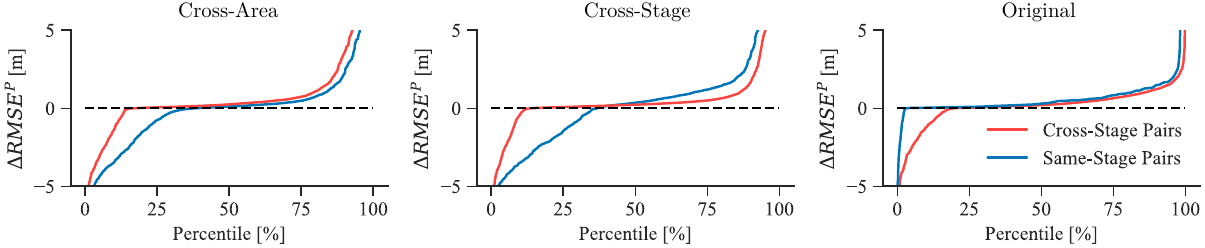}
    \caption{\textbf{Percentile distribution of the change in Pairwise RMSE ($\Delta RMSE^P$) across all graph edges using PoseGraphNet for multi-way registration, for all dataset splits}. The area under the dashed line demonstrates a decrease in RMSE and the area above demonstrates an increase. A curve's slope points to the rate of change. Areas created between the dashed line and a curve also demonstrate the amount of edges changed -- larger area means more edges changed.}
    \label{fig:multiway_change}
\end{figure*}

Figure~\ref{fig:multiway_change} ablates the percentile distribution of change in pairwise RMSE ($\Delta RMSE^P$) across cross- and same-stage graph edges for all dataset splits, when using PoseGraphNet for multi-way registration. While Table~\ref{tab:multiway2} shows that RMSE$^G$ is high and RMSE$^P$ only improves in the original split, we can still observe RMSE$^P$ improving over some the graph edges in all splits. Areas below the zero dashed line point to improvements after the multi-way registration. 
\begin{itemize}
\item \textbf{Decrease in RMSE$^P$:} In the cross-area and cross-stage splits, improvements mainly come from the same-stage edges. Cross-stage ones show a lower rate of improvement, i.e., these cases are more challenging to solve. In the original split, where the recall is already high after pairwise registration, the improvement of both same- and cross-stage pairs is smaller, with the former having minimal impact and the latter contributing the most.
\item \textbf{No change:} In all splits, most cross-stage edges undergo almost no improvement (the curve is close to parallel to the dashed line), pointing to the particular challenge in temporal registration. Same-stage edges showcase a similar behaviour mainly in the original split. In the cross-stage split, same-stage edges barely remain the same; they either improve or deteriorate.
\item \textbf{Increase in RMSE$^P$:} Cross-stage and same-stage edges have a similar rate of deterioration per split. In the cross-area split, cross-stage edges show the most deterioration versus the same-stage ones, in contrast to the cross-stage and original splits. However, for the latter, the difference between cross- and same-stage edges in terms of most deterioration is small.
\end{itemize}

\begin{figure*}[ht!]
    \centering
    \includegraphics[width=0.73\linewidth,keepaspectratio]{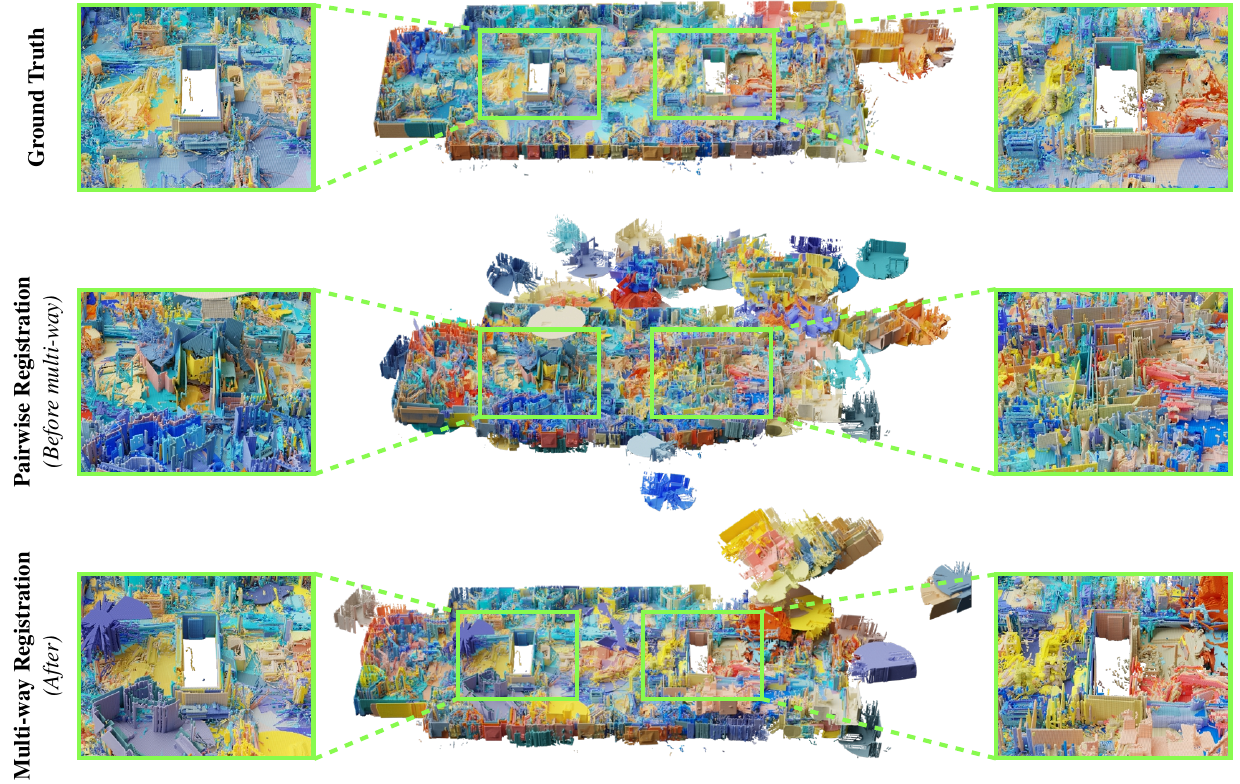}
    \caption{\textbf{Spatiotemporal registration after the pairwise~\cite{predator} and multi-way~\cite{Choi2015robust} tasks for Area F.} Different color hues represent different stages. Note the improved alignment in the elevator shafts after  multi-way registration.}
    \label{fig:multicolor}
\end{figure*}

\begin{figure*}[ht!]
    \centering
    \includegraphics[width=0.73\linewidth,keepaspectratio]{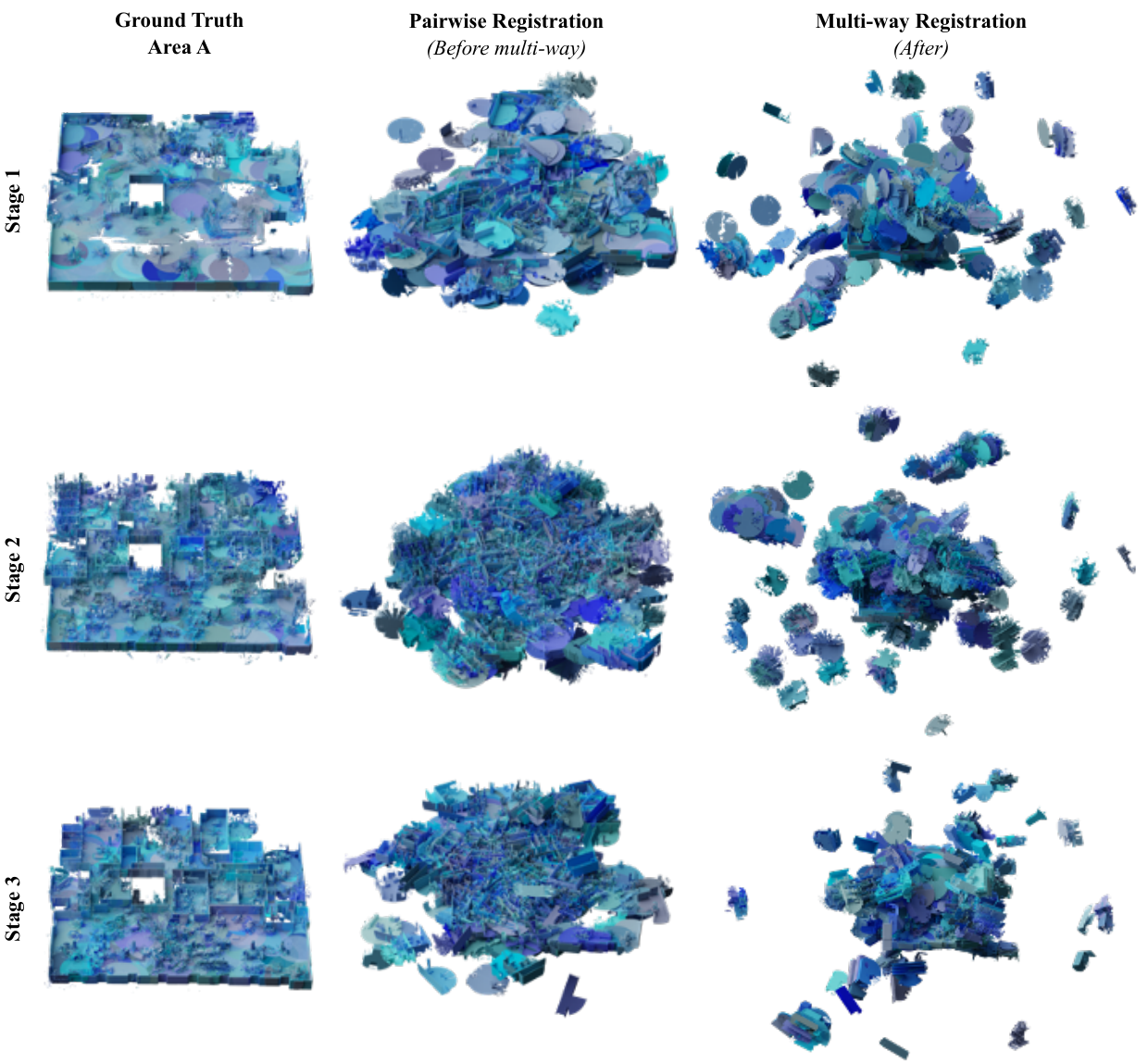}
    \caption{\textbf{Spatiotemporal registration results for Area A per temporal stage.} Different blue hues denote independent \onepc{} locations. When the pairwise registration from \textsc{Predator}~\cite{predator} fails to recover a rough initial alignment, the multi-way step~\cite{Choi2015robust} cannot recover it.}
    \label{fig:multiA}
\end{figure*}

\begin{figure*}[ht!]
    \centering
    \includegraphics[height=0.95\textheight,keepaspectratio]{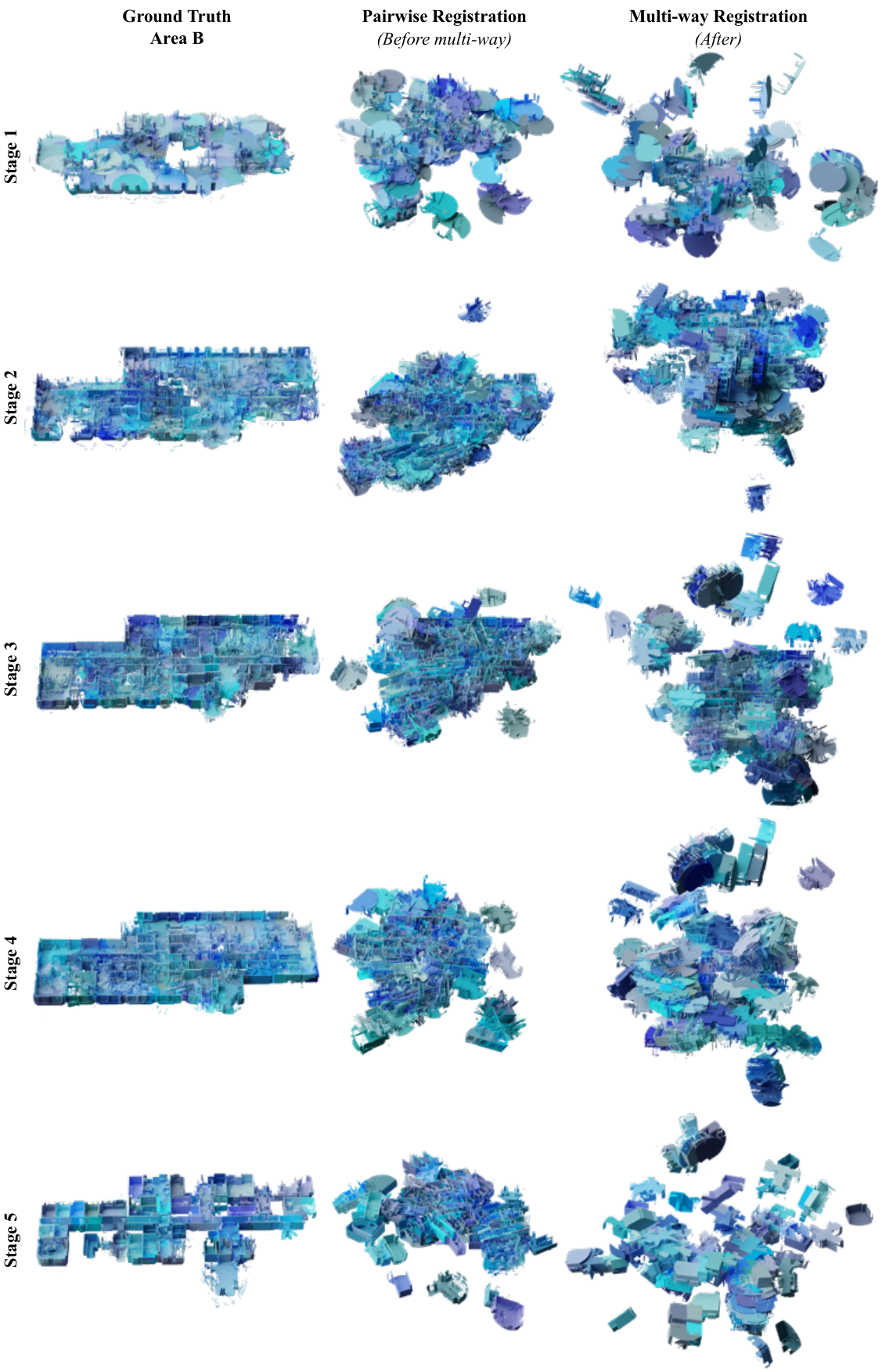}
    \caption{\textbf{Spatiotemporal registration results for Area B per temporal stage.} Different blue hues denote independent \onepc{} locations. Pairwise registration results from \textsc{Predator}~\cite{predator} and multi-way from~\cite{Choi2015robust}.}
    \label{fig:multiB}
\end{figure*}

\begin{figure*}[ht!]
    \centering
    \begin{subfigure}{\linewidth}
    \centering
        \includegraphics[width=0.83\linewidth]{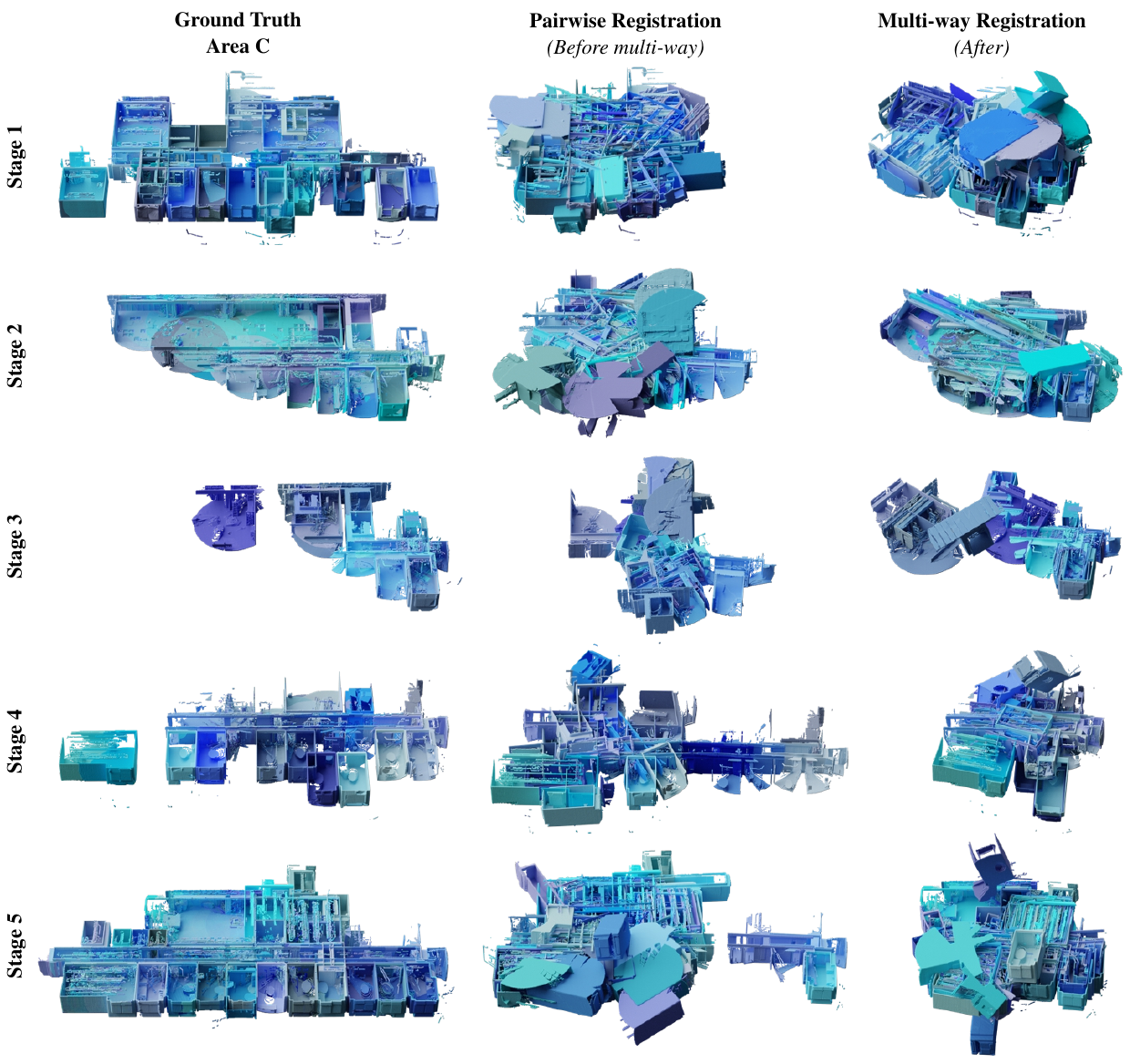}
        \caption{\textbf{Spatiotemporal registration results for Area C per temporal stage.}}
        \label{fig:multiCDa}
    \end{subfigure}
    \begin{subfigure}{\linewidth}
    \centering
        \includegraphics[width=0.83\linewidth]{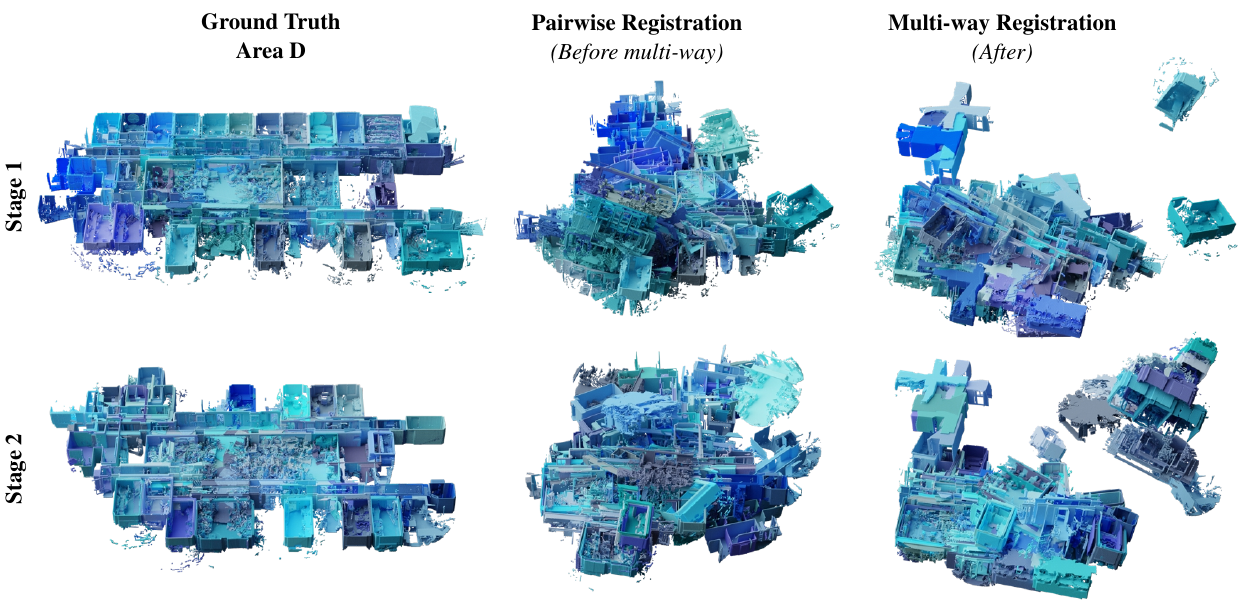}
        \caption{\textbf{Spatiotemporal registration results for Area D per temporal stage.}}
        \label{fig:multiCDb}
    \end{subfigure}
    \caption{\textbf{Spatiotemporal registration results for Area C and D per temporal stage.} Different blue hues denote independent \onepc{} locations. Pairwise registration results from \textsc{Predator}~\cite{predator} and multi-way from~\cite{Choi2015robust}.}
\end{figure*}

\begin{figure*}[ht!]
\includegraphics[width=0.95\linewidth,keepaspectratio]{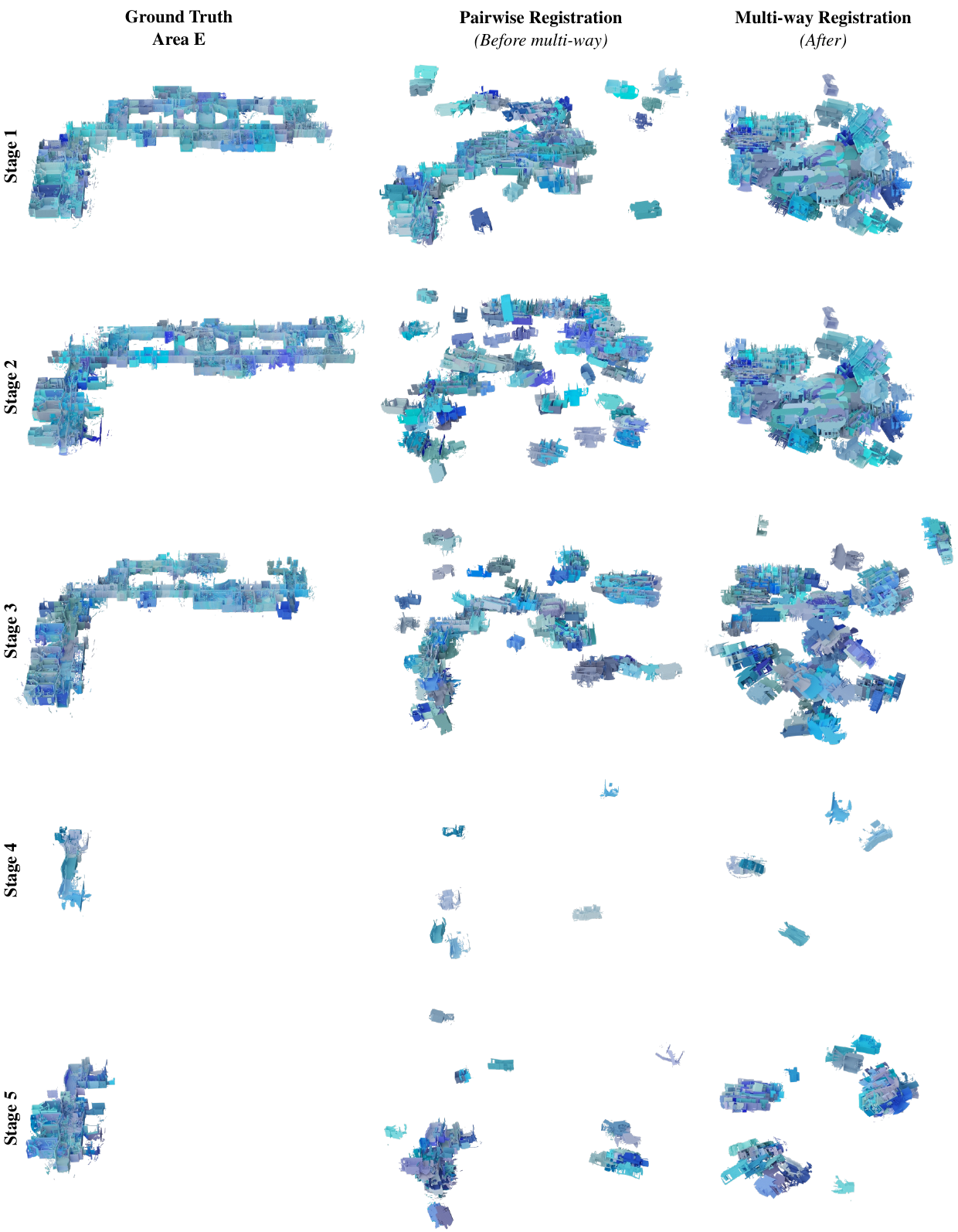}
    \caption{\textbf{Spatiotemporal registration results for Area E per temporal stage.} Different blue hues denote independent \onepc{} locations. Pairwise registration results from \textsc{Predator}~\cite{predator} and multi-way from~\cite{Choi2015robust}.}
    \label{fig:multiE}
\end{figure*}

\begin{figure*}[ht!]
\includegraphics[width=\linewidth,keepaspectratio]{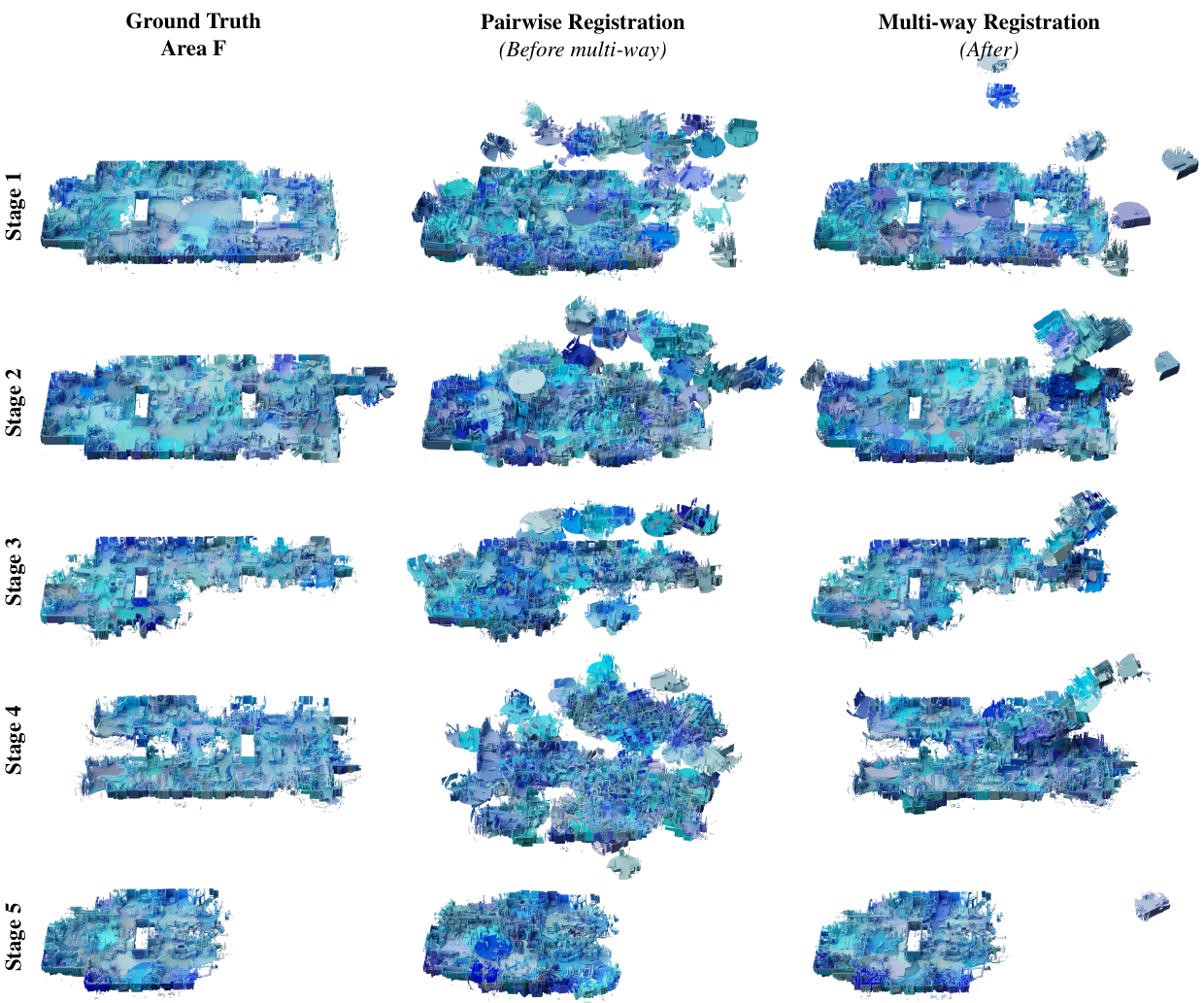}
    \caption{\textbf{Spatiotemporal registration results for Area F per temporal stage.} Different blue hues denote independent \onepc{} locations. When the pairwise registration results from \textsc{Predator}~\cite{predator} achieve better initialization, the multi-way registration from~\cite{Choi2015robust} can recover an improved alignment.}
    \label{fig:multiF}
    \vspace{-10pt}
\end{figure*}

Figure~\ref{fig:multicolor} illustrates the spatiotemporal registration after the pairwise and multi-way tasks for Area F, where different color hues represent different stages. As depicted, multi-way registration achieves superior alignment, particularly evident in the elevator shafts where pairwise registration fails to recover the alignment. However, not all areas exhibit such improvement. Figures~\ref{fig:multiA}-~\ref{fig:multiF} present the spatiotemporal registration results for all areas per temporal stage. We observe that when pairwise registration provides a relatively good initialization of the pose graph, the multi-way step can further improve the results. However, when the pairwise step fails, no further alignment improvement can be achieved (\eg Figure~\ref{fig:multiB}). This limitation persists even when the initialization is relatively good in a few stages. The failed alignment of other stages pulls the \onepc{s} away from their initial positions since the optimization goal is to attain a globally plausible solution (Figure~\ref{fig:multiCDa}).

\subsection{Overlap / Non-overlap Classification}
In order to investigate a method's behavior when dealing with non-overlapping \onepc{s} or \onepc{s} with extremely low overlap ratios (below 10\%), we conduct a binary classification task. In this task, a pair of \onepc{s} is considered overlapping if the overlap ratio exceeds a certain threshold $\theta$ (set to $0.1$ in our experiment). We employ \textsc{Predator}, the best-performing model on the \abbrev{} dataset, for this analysis. Instead of employing the average overlap score to perform the classification, which is available in \cite{predator}, we opted to use the average matching probability of all points in a \onepc{} pair as the output probability for classification.

We evaluate the classification performance using the mean Average Precision (mAP) and the Area under ROC Curve (AUROC) scores. The mAP score offers an averaged measure of precision at different recall levels, which provides a comprehensive assessment of the model's ability to correctly predict overlapping pairs even when these are unevenly distributed in the data. On the other hand, the AUROC provides a measure of the model's discriminative capacity, irrespective of the decision threshold of the matching probability. 
Table~\ref{tab:overlap_detection} demonstrates the potential of the method to classify overlapping/non-overlapping pairs and highlights areas for improvement, particularly in scenarios with significant temporal changes that occur in real-world \textit{in-the-wild} registration settings (Figure~\ref{fig:pr-curve}).

\begin{table}[h!]
    \centering
    \footnotesize
    \begin{tabular}{p{2.2cm}ccc}
            \hline
            \textbf{Metrics} & \textbf{Cross-Area} & \textbf{Cross-Stage} & \textbf{Original} \\ \hline \hline
             \multicolumn{4}{c}{\textit{All pairs}} \\ \hline
             mAP [$\uparrow$] &   0.577 & \textbf{0.774} & 0.612\\
             AUROC [$\uparrow$] & 0.582 & \textbf{0.759} & 0.642 \\ \hline
             \multicolumn{4}{c}{\textit{Same-stage pairs}} \\ \hline
             mAP [$\uparrow$] &   0.784 & \textbf{0.916} & 0.833\\
             AUROC [$\uparrow$] & 0.595 & \textbf{0.786} & 0.657 \\
             \hline
             \multicolumn{4}{c}{\textit{Different-stage pairs}} \\ \hline
            mAP [$\uparrow$] &   0.318 & 0.211 & \textbf{0.410}\\
             AUROC [$\uparrow$] & 0.541 & 0.575 & \textbf{0.616} \\
             \hline
    \end{tabular}
    \caption{\textbf{Overlap Classification on \abbrev{} with PREDATOR.} Pairs from the test set are overlapping samples, whereas pairs randomly selected from different locations are non-overlapping ones. }
    \label{tab:overlap_detection}
\end{table}

\begin{figure}[hbt!]
    \centering
    \includegraphics[width=1\columnwidth]{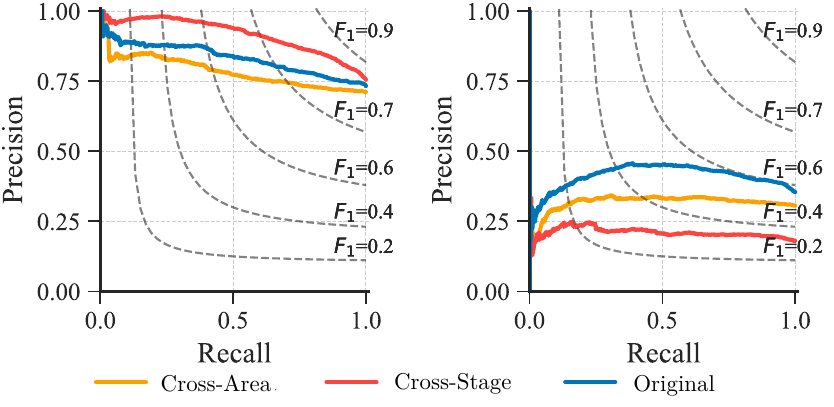}
    \caption{\textbf{Precision-recall curve on overlap classification with \textsc{Predator}.} Results are shown for same- (left) and different- (right) stage pairs. Temporal changes affect overlap detection accuracy greatly.}
    \label{fig:pr-curve}
\end{figure}

We observe that the cross-stage split exhibits the best overall and same-stage scores but encounters challenges with temporal cases. This indicates that temporal changes that were not encountered during training can impact overlap classification. It's important to note that the cross-stage split only includes the first half of temporal stages per area, meaning that the method lacks knowledge of how the construction may progress in the future\footnote{Stages across areas, although not identical, exhibit a certain pattern as construction tasks often follow a specific sequence.}. In this classification task, pairs from the test set are considered overlapping pairs, while pairs randomly selected from different locations are regarded as non-overlapping pairs. This task can be valuable when considering practical applications of a registration algorithm, where prior knowledge of \onepc{} overlap may not be available from the outset.

\section{Conclusion}

In this study, we introduced a new benchmark called \textbf{\project{}} for evaluating the performance of 3D point cloud registration in spatiotemporal scenarios. This benchmark assesses methods' capabilities across space, time, and generalization. To support this benchmark, we also presented a novel spatiotemporal dataset containing indoor areas captured over time, exhibiting significant geometric changes. Our findings, as discussed in Section~\ref{sec:experiments}, indicate that existing 3D registration methods have limited ability to handle temporal changes effectively. Moreover, the conflicting objectives between pairwise and multi-way registration tasks currently pose challenges in developing end-to-end algorithms. This paper highlights the substantial room for improvement in this field.
In addition, both the benchmark and dataset hold great potential for various applications such as in robotic navigation, virtual and augmented reality applications, construction progress monitoring, and learning and detecting change.

\bibliography{bibfile}

\begin{thebibliography}{100}
\expandafter\ifx\csname url\endcsname\relax
  \def\url#1{\texttt{#1}}\fi
\expandafter\ifx\csname urlprefix\endcsname\relax\def\urlprefix{URL }\fi
\expandafter\ifx\csname href\endcsname\relax
  \def\href#1#2{#2} \def\path#1{#1}\fi

\bibitem{simonyan2014two}
K.~Simonyan, A.~Zisserman, Two-stream convolutional networks for action
  recognition in videos, in: NeurIPS, 2014.

\bibitem{purushwalkam2020aligning}
S.~Purushwalkam, T.~Ye, S.~Gupta, A.~Gupta, Aligning videos in space and time,
  in: Computer Vision--ECCV 2020: 16th European Conference, Glasgow, UK, August
  23--28, 2020, Proceedings, Part XXVI 16, Springer, 2020, pp. 262--278.

\bibitem{haresh2021learning}
S.~Haresh, S.~Kumar, H.~Coskun, S.~N. Syed, A.~Konin, Z.~Zia, Q.-H. Tran,
  Learning by aligning videos in time, in: CVPR, 2021, pp. 5548--5558.

\bibitem{kwon2022context}
T.~Kwon, B.~Tekin, S.~Tang, M.~Pollefeys, Context-aware sequence alignment
  using 4d skeletal augmentation, in: CVPR, 2022, pp. 8172--8182.

\bibitem{deng2024vg4d}
Z.~Deng, X.~Li, X.~Li, Y.~Tong, S.~Zhao, M.~Liu, Vg4d: Vision-language model
  goes 4d video recognition, in: ICRA, 2024.

\bibitem{qi2021offboard}
C.~R. Qi, Y.~Zhou, M.~Najibi, P.~Sun, K.~Vo, B.~Deng, D.~Anguelov, Offboard 3d
  object detection from point cloud sequences, in: CVPR, 2021.

\bibitem{huang2022dynamic}
S.~Huang, Z.~Gojcic, J.~Huang, A.~Wieser, K.~Schindler, Dynamic 3d scene
  analysis by point cloud accumulation, in: ECCV, 2022.

\bibitem{zhang2023towards}
L.~Zhang, A.~J. Yang, Y.~Xiong, S.~Casas, B.~Yang, M.~Ren, R.~Urtasun, Towards
  unsupervised object detection from lidar point clouds, in: CVPR, 2023.

\bibitem{sakurada2020weakly}
K.~Sakurada, M.~Shibuya, W.~Wang, Weakly supervised silhouette-based semantic
  scene change detection, in: ICRA, 2020.

\bibitem{lei2020hierarchical}
Y.~Lei, D.~Peng, P.~Zhang, Q.~Ke, H.~Li, Hierarchical paired channel fusion
  network for street scene change detection, IEEE Transactions on Image
  Processing 30 (2020) 55--67.

\bibitem{ru2020multi}
L.~Ru, B.~Du, C.~Wu, Multi-temporal scene classification and scene change
  detection with correlation based fusion, IEEE Transactions on Image
  Processing 30 (2020) 1382--1394.

\bibitem{prabhakar2020cdnet++}
K.~R. Prabhakar, A.~Ramaswamy, S.~Bhambri, J.~Gubbi, R.~V. Babu,
  B.~Purushothaman, Cdnet++: Improved change detection with deep neural network
  feature correlation, in: 2020 International Joint Conference on Neural
  Networks (IJCNN), 2020.

\bibitem{wang2021transcd}
Z.~Wang, Y.~Zhang, L.~Luo, N.~Wang, Transcd: scene change detection via
  transformer-based architecture, Optics Express.

\bibitem{li2022scene}
J.~Li, P.~Tang, Y.~Wu, M.~Pan, Z.~Tang, G.~Hui, Scene change detection:
  semantic and depth information, Multimedia Tools and Applications.

\bibitem{bourdis2011constrained}
N.~Bourdis, D.~Marraud, H.~Sahbi, Constrained optical flow for aerial image
  change detection, in: 2011 IEEE International Geoscience and Remote Sensing
  Symposium, IEEE, 2011, pp. 4176--4179.

\bibitem{peng2019end}
D.~Peng, Y.~Zhang, H.~Guan, End-to-end change detection for high resolution
  satellite images using improved unet++, Remote Sensing 11~(11) (2019) 1382.

\bibitem{chen2020dasnet}
J.~Chen, Z.~Yuan, J.~Peng, L.~Chen, H.~Huang, J.~Zhu, Y.~Liu, H.~Li, Dasnet:
  Dual attentive fully convolutional siamese networks for change detection in
  high-resolution satellite images, IEEE Journal of Selected Topics in Applied
  Earth Observations and Remote Sensing 14 (2020) 1194--1206.

\bibitem{zheng2021change}
Z.~Zheng, A.~Ma, L.~Zhang, Y.~Zhong, Change is everywhere: Single-temporal
  supervised object change detection in remote sensing imagery, in: Proceedings
  of the IEEE/CVF international conference on computer vision, 2021, pp.
  15193--15202.

\bibitem{cheng2024change}
G.~Cheng, Y.~Huang, X.~Li, S.~Lyu, Z.~Xu, H.~Zhao, Q.~Zhao, S.~Xiang, Change
  detection methods for remote sensing in the last decade: A comprehensive
  review, Remote Sensing 16~(13) (2024) 2355.

\bibitem{xiao2015street}
W.~Xiao, B.~Vallet, M.~Br{\'e}dif, N.~Paparoditis, Street environment change
  detection from mobile laser scanning point clouds, ISPRS Journal of
  Photogrammetry and Remote Sensing 107 (2015) 38--49.

\bibitem{qin20163d}
R.~Qin, J.~Tian, P.~Reinartz, 3d change detection--approaches and applications,
  ISPRS Journal of Photogrammetry and Remote Sensing 122 (2016) 41--56.

\bibitem{kharroubi2022three}
A.~Kharroubi, F.~Poux, Z.~Ballouch, R.~Hajji, R.~Billen, Three dimensional
  change detection using point clouds: A review, Geomatics 2~(4) (2022)
  457--485.

\bibitem{gehrung2022change}
J.~Gehrung, M.~Hebel, M.~Arens, U.~Stilla, Change detection in street
  environments based on mobile laser scanning: A fuzzy spatial reasoning
  approach, ISPRS Open Journal of Photogrammetry and Remote Sensing 5 (2022)
  100019.

\bibitem{huang2022semantics}
R.~Huang, Y.~Xu, L.~Hoegner, U.~Stilla, Semantics-aided 3d change detection on
  construction sites using uav-based photogrammetric point clouds, Automation
  in Construction 134 (2022) 104057.

\bibitem{de2023siamese}
I.~de~G{\'e}lis, S.~Lef{\`e}vre, T.~Corpetti, Siamese kpconv: 3d multiple
  change detection from raw point clouds using deep learning, ISPRS Journal of
  Photogrammetry and Remote Sensing 197 (2023) 274--291.

\bibitem{stilla2023change}
U.~Stilla, Y.~Xu, Change detection of urban objects using 3d point clouds: A
  review, ISPRS Journal of Photogrammetry and Remote Sensing 197 (2023)
  228--255.

\bibitem{lopez20243d}
M.~F. L{\'o}pez-Armenta, R.~Nespeca, 3d change detection for cultural heritage
  monitoring: Two case studies of underground sculptural reliefs, Digital
  Applications in Archaeology and Cultural Heritage 33 (2024) e00328.

\bibitem{wald20193rscan}
J.~Wald, A.~Avetisyan, N.~Navab, F.~Tombari, M.~Niessner, Rio: 3d object
  instance re-localization in changing indoor environments, in: ICCV, 2019.

\bibitem{wald2020beyond}
J.~Wald, T.~Sattler, S.~Golodetz, T.~Cavallari, F.~Tombari, Beyond controlled
  environments: 3d camera re-localization in changing indoor scenes, in: ECCV,
  2020.

\bibitem{halber2019rescan}
M.~Halber, Y.~Shi, K.~Xu, T.~Funkhouser, Rescan: Inductive instance
  segmentation for indoor rgbd scans, in: ICCV, 2019.

\bibitem{zhu2023living}
L.~Zhu, S.~Huang, I.~A. Konrad~Schindler, Living scenes: Multi-object
  relocalization and reconstruction in changing 3d environments, in: The IEEE
  Conference on Computer Vision and Pattern Recognition (CVPR), 2024.

\bibitem{droeschel2017continuous}
D.~Droeschel, M.~Schwarz, S.~Behnke, Continuous mapping and localization for
  autonomous navigation in rough terrain using a 3d laser scanner, Robotics and
  Autonomous Systems 88 (2017) 104--115.

\bibitem{wang2020mobile}
B.~Wang, Z.~Liu, Q.~Li, A.~Prorok, Mobile robot path planning in dynamic
  environments through globally guided reinforcement learning, IEEE Robotics
  and Automation Letters 5~(4) (2020) 6932--6939.

\bibitem{wang2021navigation}
X.~Wang, Y.~Mizukami, M.~Tada, F.~Matsuno, Navigation of a mobile robot in a
  dynamic environment using a point cloud map, Artificial Life and Robotics 26
  (2021) 10--20.

\bibitem{straub2019replica}
J.~Straub, T.~Whelan, L.~Ma, Y.~Chen, E.~Wijmans, S.~Green, J.~J. Engel,
  R.~Mur-Artal, C.~Ren, S.~Verma, et~al., The replica dataset: A digital
  replica of indoor spaces, arXiv preprint arXiv:1906.05797.

\bibitem{tsamis2021lifelongmapping}
G.~Tsamis, I.~Kostavelis, D.~Giakoumis, D.~Tzovaras, Towards life-long mapping
  of dynamic environments using temporal persistence modeling, in: 2020 25th
  International Conference on Pattern Recognition (ICPR), 2021, pp.
  10480--10485.
\newblock \href {http://dx.doi.org/10.1109/ICPR48806.2021.9413161}
  {\path{doi:10.1109/ICPR48806.2021.9413161}}.

\bibitem{stefanini2023safe}
E.~Stefanini, E.~Ciancolini, A.~Settimi, L.~Pallottino,
  \href{https://www.mdpi.com/1424-8220/23/13/6066}{Safe and robust map updating
  for long-term operations in dynamic environments}, Sensors 23~(13).
\newblock \href {http://dx.doi.org/10.3390/s23136066}
  {\path{doi:10.3390/s23136066}}.
\newline\urlprefix\url{https://www.mdpi.com/1424-8220/23/13/6066}

\bibitem{munaro2020towards}
M.~R. Munaro, S.~F. Tavares, L.~Bragan{\c{c}}a, Towards circular and more
  sustainable buildings: A systematic literature review on the circular economy
  in the built environment, Journal of Cleaner Production 260 (2020) 121134.

\bibitem{love2002using}
P.~E. Love, G.~D. Holt, L.~Y. Shen, H.~Li, Z.~Irani, Using systems dynamics to
  better understand change and rework in construction project management
  systems, International journal of project management 20~(6) (2002) 425--436.

\bibitem{ma2020challenges}
M.~Ma, V.~W. Tam, K.~N. Le, W.~Li, Challenges in current construction and
  demolition waste recycling: A china study, Waste Management 118 (2020)
  610--625.

\bibitem{rusu2009fast}
R.~B. Rusu, N.~Blodow, M.~Beetz, Fast point feature histograms (fpfh) for 3d
  registration, in: ICRA, 2009.

\bibitem{bai2020d3feat}
X.~Bai, Z.~Luo, L.~Zhou, H.~Fu, L.~Quan, C.-L. Tai, D3feat: Joint learning of
  dense detection and description of 3d local features, in: CVPR, 2020.

\bibitem{Choy2019FCGF}
C.~Choy, J.~Park, V.~Koltun, Fully convolutional geometric features, in: ICCV,
  2019.

\bibitem{predator}
S.~Huang, Z.~Gojcic, M.~Usvyatsov, A.~Wieser, K.~Schindler, Predator:
  Registration of 3d point clouds with low overlap, in: CVPR, 2021.

\bibitem{qin2022geometric}
Z.~Qin, H.~Yu, C.~Wang, Y.~Guo, Y.~Peng, K.~Xu, Geometric transformer for fast
  and robust point cloud registration, in: CVPR, 2022.

\bibitem{hussain2013change}
M.~Hussain, D.~Chen, A.~Cheng, H.~Wei, D.~Stanley, Change detection from
  remotely sensed images: From pixel-based to object-based approaches, ISPRS
  Journal of photogrammetry and remote sensing.

\bibitem{yew2021city}
Z.~J. Yew, G.~H. Lee, City-scale scene change detection using point clouds, in:
  ICRA, 2021.

\bibitem{pollard2007change}
T.~Pollard, J.~L. Mundy, Change detection in a 3-d world, in: 2007 IEEE
  Conference on Computer Vision and Pattern Recognition, Ieee, 2007, pp. 1--6.

\bibitem{chen2021moving}
X.~Chen, S.~Li, B.~Mersch, L.~Wiesmann, J.~Gall, J.~Behley, C.~Stachniss,
  Moving object segmentation in 3d lidar data: A learning-based approach
  exploiting sequential data, IEEE Robotics and Automation Letters.

\bibitem{accumulation}
S.~Huang, Z.~Gojcic, J.~Huang, K.~S. Andreas~Wieser, Dynamic 3d scene analysis
  by point cloud accumulation, in: ECCV, 2022.

\bibitem{horn1981determining}
B.~K. Horn, B.~G. Schunck, Determining optical flow, Artificial Intelligence.

\bibitem{black1993framework}
M.~J. Black, P.~Anandan, A framework for the robust estimation of optical flow,
  in: ICCV, 1993.

\bibitem{brox2009large}
T.~Brox, C.~Bregler, J.~Malik, Large displacement optical flow, in: CVPR, 2009.

\bibitem{ilg2017flownet}
E.~Ilg, N.~Mayer, T.~Saikia, M.~Keuper, A.~Dosovitskiy, T.~Brox, Flownet 2.0:
  Evolution of optical flow estimation with deep networks, in: CVPR, 2017.

\bibitem{hui2018liteflownet}
T.-W. Hui, X.~Tang, C.~C. Loy, Liteflownet: A lightweight convolutional neural
  network for optical flow estimation, in: CVPR, 2018.

\bibitem{fischer2015flownet}
P.~Fischer, A.~Dosovitskiy, E.~Ilg, P.~H{\"a}usser, C.~Haz{\i}rbas, V.~Golkov,
  P.~Van~der Smagt, D.~Cremers, T.~Brox, Flownet: Learning optical flow with
  convolutional networks, arXiv preprint arXiv:1504.06852.

\bibitem{teed2020raft}
Z.~Teed, J.~Deng, Raft: Recurrent all-pairs field transforms for optical flow,
  in: ECCV, 2020.

\bibitem{vedula1999three}
S.~Vedula, S.~Baker, P.~Rander, R.~Collins, T.~Kanade, Three-dimensional scene
  flow, in: ICCV, 1999.

\bibitem{sun2010secrets}
D.~Sun, S.~Roth, M.~J. Black, Secrets of optical flow estimation and their
  principles, in: CVPR, 2010.

\bibitem{vogel20113d}
C.~Vogel, K.~Schindler, S.~Roth, 3d scene flow estimation with a rigid motion
  prior, in: ICCV, 2011.

\bibitem{vogel2013piecewise}
C.~Vogel, K.~Schindler, S.~Roth, Piecewise rigid scene flow, in: ICCV, 2013.

\bibitem{vogel20153d}
C.~Vogel, K.~Schindler, S.~Roth, 3d scene flow estimation with a piecewise
  rigid scene model, IJCV.

\bibitem{liu2019flownet3d}
X.~Liu, C.~R. Qi, L.~J. Guibas, Flownet3d: Learning scene flow in 3d point
  clouds, in: CVPR, 2019.

\bibitem{puy2020flot}
G.~Puy, A.~Boulch, R.~Marlet, Flot: Scene flow on point clouds guided by
  optimal transport, in: ECCV, 2020.

\bibitem{8454294}
L.~Wang, Y.~Xiong, Z.~Wang, Y.~Qiao, D.~Lin, X.~Tang, L.~Van~Gool, Temporal
  segment networks for action recognition in videos, T-PAMI.

\bibitem{milan2016mot16}
A.~Milan, L.~Leal-Taix{\'e}, I.~Reid, S.~Roth, K.~Schindler, Mot16: A benchmark
  for multi-object tracking, arXiv preprint arXiv:1603.00831.

\bibitem{xiang2015learning}
Y.~Xiang, A.~Alahi, S.~Savarese, Learning to track: Online multi-object
  tracking by decision making, in: ICCV, 2015.

\bibitem{rempe2020caspr}
D.~Rempe, T.~Birdal, Y.~Zhao, Z.~Gojcic, S.~Sridhar, L.~J. Guibas, Caspr:
  Learning canonical spatiotemporal point cloud representations, NeurIPS.

\bibitem{huang2022multiway}
J.~Huang, T.~Birdal, Z.~Gojcic, L.~J. Guibas, S.-M. Hu, Multiway non-rigid
  point cloud registration via learned functional map synchronization, T-PAMI.

\bibitem{pumarola2021d}
A.~Pumarola, E.~Corona, G.~Pons-Moll, F.~Moreno-Noguer, D-nerf: Neural radiance
  fields for dynamic scenes, in: CVPR, 2021.

\bibitem{martin2021nerf}
R.~Martin-Brualla, N.~Radwan, M.~S. Sajjadi, J.~T. Barron, A.~Dosovitskiy,
  D.~Duckworth, Nerf in the wild: Neural radiance fields for unconstrained
  photo collections, in: CVPR, 2021.

\bibitem{golparvar2009d4ar}
M.~Golparvar-Fard, F.~Pe{\~n}a-Mora, S.~Savarese, D4ar--a 4-dimensional
  augmented reality model for automating construction progress monitoring data
  collection, processing and communication, Journal of information technology
  in construction 14~(13) (2009) 129--153.

\bibitem{dong20174d}
J.~Dong, J.~G. Burnham, B.~Boots, G.~Rains, F.~Dellaert, 4d crop monitoring:
  Spatio-temporal reconstruction for agriculture, in: 2017 IEEE International
  Conference on Robotics and Automation (ICRA), IEEE, 2017, pp. 3878--3885.

\bibitem{griffith2020transforming}
S.~Griffith, F.~Dellaert, C.~Pradalier, Transforming multiple visual surveys of
  a natural environment into time-lapses, The International Journal of Robotics
  Research 39~(1) (2020) 100--126.

\bibitem{schindler2010probabilistic}
G.~Schindler, F.~Dellaert, Probabilistic temporal inference on reconstructed 3d
  scenes, in: CVPR, 2010.

\bibitem{schindler2007inferring}
G.~Schindler, F.~Dellaert, S.~B. Kang, Inferring temporal order of images from
  3d structure, in: CVPR, 2007.

\bibitem{matzen2014scene}
K.~Matzen, N.~Snavely, Scene chronology, in: ECCV, 2014.

\bibitem{martin20153d}
R.~Martin-Brualla, D.~Gallup, S.~M. Seitz, 3d time-lapse reconstruction from
  internet photos, in: ICCV, 2015.

\bibitem{lin2023neural}
H.~Lin, Q.~Wang, R.~Cai, S.~Peng, H.~Averbuch-Elor, X.~Zhou, N.~Snavely, Neural
  scene chronology, in: CVPR, 2023.

\bibitem{liu2019meteornet}
X.~Liu, M.~Yan, J.~Bohg, Meteornet: Deep learning on dynamic 3d point cloud
  sequences, in: ICCV, 2019.

\bibitem{choy20194d}
C.~Choy, J.~Gwak, S.~Savarese, 4d spatio-temporal convnets: Minkowski
  convolutional neural networks, in: CVPR, 2019.

\bibitem{fan2020pstnet}
H.~Fan, X.~Yu, Y.~Ding, Y.~Yang, M.~Kankanhalli, {PSTNet}: Point
  spatio-temporal convolution on point cloud sequences, in: ICLR, 2020.

\bibitem{adam2022objects}
A.~Adam, T.~Sattler, K.~Karantzalos, T.~Pajdla, Objects can move: 3d change
  detection by geometric transformation consistency, in: ECCV, 2022.

\bibitem{Armeni_2016_CVPR}
I.~Armeni, O.~Sener, A.~R. Zamir, H.~Jiang, I.~Brilakis, M.~Fischer,
  S.~Savarese, 3d semantic parsing of large-scale indoor spaces, in: CVPR,
  2016.

\bibitem{armeni2017joint}
I.~Armeni, S.~Sax, A.~R. Zamir, S.~Savarese, Joint 2d-3d-semantic data for
  indoor scene understanding, arXiv preprint arXiv:1702.01105.

\bibitem{chang2017matterport3d}
A.~Chang, A.~Dai, T.~Funkhouser, M.~Halber, M.~Niessner, M.~Savva, S.~Song,
  A.~Zeng, Y.~Zhang, Matterport3d: Learning from rgb-d data in indoor
  environments, arXiv preprint arXiv:1709.06158.

\bibitem{hua2016scenenn}
B.-S. Hua, Q.-H. Pham, D.~T. Nguyen, M.-K. Tran, L.-F. Yu, S.-K. Yeung,
  Scenenn: A scene meshes dataset with annotations, in: 3DV, 2016.

\bibitem{dai2017scannet}
A.~Dai, A.~X. Chang, M.~Savva, M.~Halber, T.~Funkhouser, M.~Nie{\ss}ner,
  Scannet: Richly-annotated 3d reconstructions of indoor scenes, in: CVPR,
  2017.

\bibitem{park2021changesim}
J.-M. Park, J.-H. Jang, S.-M. Yoo, S.-K. Lee, U.-H. Kim, J.-H. Kim, Changesim:
  towards end-to-end online scene change detection in industrial indoor
  environments, in: IROS, 2021.

\bibitem{Qiu2020IndoorData}
Y.~Qiu, Y.~Satoh, R.~Suzuki, K.~Iwata, H.~Kataoka, {Indoor scene change
  captioning based on multimodality data}, Sensors (Switzerland).

\bibitem{sarlin2022lamar}
P.-E. Sarlin, M.~Dusmanu, J.~L. Sch\"onberger, P.~Speciale, L.~Gruber,
  V.~Larsson, O.~Miksik, M.~Pollefeys, {LaMAR}: {B}enchmarking {L}ocalization
  and {M}apping for {A}ugmented {R}eality, in: ECCV, 2022.

\bibitem{wenzel2020fourseasons}
P.~Wenzel, R.~Wang, N.~Yang, Q.~Cheng, Q.~Khan, L.~von Stumberg, N.~Zeller,
  D.~Cremers, {4Seasons}: A cross-season dataset for multi-weather {SLAM} in
  autonomous driving, in: GCPR, 2020.

\bibitem{Ros_2016_CVPR}
G.~Ros, L.~Sellart, J.~Materzynska, D.~Vazquez, A.~M. Lopez, The synthia
  dataset: A large collection of synthetic images for semantic segmentation of
  urban scenes, in: CVPR, 2016.

\bibitem{HernandezBMVC17}
D.~Hernandez-Juarez, L.~Schneider, A.~Espinosa, D.~Vazquez, A.~M. Lopez,
  U.~Franke, M.~Pollefeys, J.~C. Moure, Slanted stixels: Representing san
  francisco’s steepest streets, in: BMVC, 2017.

\bibitem{bengarICCVW19}
J.~Zolfaghari~Bengar, A.~Gonzalez-Garcia, G.~Villalonga, B.~Raducanu, H.~H.
  Aghdam, M.~Mozerov, A.~M. Lopez, J.~van~de Weijer, Temporal coherence for
  active learning in videos, in: ICCV Workshops, 2019.

\bibitem{geiger2013vision}
A.~Geiger, P.~Lenz, C.~Stiller, R.~Urtasun, Vision meets robotics: The kitti
  dataset, The International Journal of Robotics Research.

\bibitem{cordts2016cityscapes}
M.~Cordts, M.~Omran, S.~Ramos, T.~Rehfeld, M.~Enzweiler, R.~Benenson,
  U.~Franke, S.~Roth, B.~Schiele, The cityscapes dataset for semantic urban
  scene understanding, in: CVPR, 2016.

\bibitem{martin1981random}
M.~A. Fischler, R.~C. Bolles, Random sample consensus: A paradigm for model
  fitting with applications to image analysis and automated cartography,
  Communication of ACM.

\bibitem{li20073d}
H.~Li, R.~Hartley, The 3d-3d registration problem revisited, in: 2007 IEEE 11th
  international conference on computer vision, IEEE, 2007, pp. 1--8.

\bibitem{hartley2009global}
R.~I. Hartley, F.~Kahl, Global optimization through rotation space search,
  International Journal of Computer Vision 82~(1) (2009) 64--79.

\bibitem{cai2019practical}
Z.~Cai, T.-J. Chin, A.~P. Bustos, K.~Schindler, Practical optimal registration
  of terrestrial lidar scan pairs, ISPRS journal of photogrammetry and remote
  sensing 147 (2019) 118--131.

\bibitem{johnson1999}
A.~Johnson, M.~Hebert, Using spin images for efficient object recognition in
  cluttered 3d scenes, IEEE TPAMI.

\bibitem{rusu2008PFH}
R.~B. Rusu, N.~Blodow, Z.~C. Marton, M.~Beetz, Aligning point cloud views using
  persistent feature histograms, in: IROS, 2008.

\bibitem{rusu2009FPFH}
R.~B. Rusu, N.~Blodow, M.~Beetz, Fast point feature histograms ({FPFH}) for
  3{D} registration, in: ICRA, 2009.

\bibitem{tombari2010SHOT}
F.~Tombari, S.~Salti, L.~Di~Stefano, Unique signatures of histograms for local
  surface description, in: ECCV, 2010.

\bibitem{tombari2010USC}
F.~Tombari, S.~Salti, L.~Di~Stefano, Unique shape context for 3{D} data
  description, in: ACM Workshop on 3D Object Retrieval, 2010.

\bibitem{theiler2014keypoint}
P.~W. Theiler, J.~D. Wegner, K.~Schindler,
  \href{https://www.sciencedirect.com/science/article/pii/S0924271614001701}{Keypoint-based
  4-points congruent sets -- automated marker-less registration of laser
  scans}, ISPRS Journal of Photogrammetry and Remote Sensing 96 (2014)
  149--163.
\newblock \href
  {http://dx.doi.org/https://doi.org/10.1016/j.isprsjprs.2014.06.015}
  {\path{doi:https://doi.org/10.1016/j.isprsjprs.2014.06.015}}.
\newline\urlprefix\url{https://www.sciencedirect.com/science/article/pii/S0924271614001701}

\bibitem{hermans2017defense}
A.~Hermans, L.~Beyer, B.~Leibe, In defense of the triplet loss for person
  re-identification, arXiv preprint arXiv:1703.07737.

\bibitem{sun2020circle}
Y.~Sun, C.~Cheng, Y.~Zhang, C.~Zhang, L.~Zheng, Z.~Wang, Y.~Wei, Circle loss: A
  unified perspective of pair similarity optimization, in: CVPR, 2020.

\bibitem{gojcic20193DSmoothNet}
Z.~Gojcic, C.~Zhou, J.~D. Wegner, A.~Wieser, The perfect match: 3d point cloud
  matching with smoothed densities, in: CVPR, 2019.

\bibitem{ao2021spinnet}
S.~Ao, Q.~Hu, B.~Yang, A.~Markham, Y.~Guo, Spinnet: Learning a general surface
  descriptor for 3d point cloud registration, in: CVPR, 2021.

\bibitem{yu2021cofinet}
H.~Yu, F.~Li, M.~Saleh, B.~Busam, S.~Ilic, Cofinet: Reliable coarse-to-fine
  correspondences for robust pointcloud registration, NeurIPS 34.

\bibitem{attaiki2021dpfm}
S.~Attaiki, G.~Pai, M.~Ovsjanikov, Dpfm: Deep partial functional maps, in: 3DV,
  2021.

\bibitem{wang2019dcp}
Y.~Wang, J.~M. Solomon, Deep closest point: Learning representations for point
  cloud registration, in: ICCV, 2019.

\bibitem{wang2019prnet}
Y.~Wang, J.~M. Solomon, Prnet: Self-supervised learning for partial-to-partial
  registration, Advances in neural information processing systems 32.

\bibitem{yew2020rpm}
Z.~J. Yew, G.~H. Lee, {RPM-Net}: Robust point matching using learned features,
  in: CVPR, 2020.

\bibitem{aoki2019pointnetlk}
Y.~Aoki, H.~Goforth, R.~A. Srivatsan, S.~Lucey, {PointnetLK}: Robust \&
  efficient point cloud registration using {Pointnet}, in: CVPR, 2019.

\bibitem{wei2023generalized}
T.~Wei, Y.~Patel, A.~Shekhovtsov, J.~Matas, D.~Barath, Generalized
  differentiable ransac, in: Proceedings of the IEEE/CVF International
  Conference on Computer Vision (ICCV), 2023, pp. 17649--17660.

\bibitem{kabsch}
K.~S. Arun, T.~S. Huang, S.~D. Blostein, Least-squares fitting of two 3-d point
  sets, IEEE TPAMI.

\bibitem{wu2015ModelNet}
Z.~Wu, S.~Song, A.~Khosla, F.~Yu, L.~Zhang, X.~Tang, J.~Xiao, 3d shapenets: A
  deep representation for volumetric shapes, in: CVPR, 2015.

\bibitem{huber2003fully}
D.~F. Huber, M.~Hebert, Fully automatic registration of multiple 3d data sets,
  Image and Vision Computing 21~(7) (2003) 637--650.

\bibitem{fantoni2012accurate}
S.~Fantoni, U.~Castellani, A.~Fusiello, Accurate and automatic alignment of
  range surfaces, in: International Conference on 3D Imaging, Modeling,
  Processing, Visualization \& Transmission, IEEE, 2012.

\bibitem{torsello2011multiview}
A.~Torsello, E.~Rodola, A.~Albarelli, Multiview registration via graph
  diffusion of dual quaternions, in: CVPR, 2011.

\bibitem{Choi2015robust}
S.~Choi, Q.-Y. Zhou, V.~Koltun, Robust reconstruction of indoor scenes, in:
  CVPR, 2015.

\bibitem{theiler2015globally}
P.~W. Theiler, J.~D. Wegner, K.~Schindler, Globally consistent registration of
  terrestrial laser scans via graph optimization, ISPRS journal of
  photogrammetry and remote sensing 109 (2015) 126--138.

\bibitem{bernard2015solution}
F.~Bernard, J.~Thunberg, P.~Gemmar, F.~Hertel, A.~Husch, J.~Goncalves, A
  solution for multi-alignment by transformation synchronisation, in: CVPR,
  2015.

\bibitem{zhou2016fast}
Q.-Y. Zhou, J.~Park, V.~Koltun, Fast global registration, in: ECCV, Springer,
  2016.

\bibitem{bhattacharya2019efficient}
U.~Bhattacharya, V.~M. Govindu, Efficient and robust registration on the 3d
  special euclidean group, in: ICCV, 2019.

\bibitem{triggs2000bundle}
B.~Triggs, P.~F. McLauchlan, R.~I. Hartley, A.~W. Fitzgibbon, Bundle
  adjustment—a modern synthesis, in: Vision Algorithms: Theory and Practice:
  International Workshop on Vision Algorithms Corfu, Greece, September 21--22,
  1999 Proceedings, Springer, 2000, pp. 298--372.

\bibitem{zach2018descending}
C.~Zach, G.~Bourmaud, Descending, lifting or smoothing: Secrets of robust cost
  optimization, in: Proceedings of the European Conference on Computer Vision
  (ECCV), 2018, pp. 547--562.

\bibitem{huang2019learning}
X.~Huang, Z.~Liang, X.~Zhou, Y.~Xie, L.~J. Guibas, Q.~Huang, Learning
  transformation synchronization, in: CVPR, 2019.

\bibitem{gojcic2020learning}
Z.~Gojcic, C.~Zhou, J.~D. Wegner, L.~J. Guibas, T.~Birdal, Learning multiview
  3d point cloud registration, in: CVPR, 2020.

\bibitem{zhang2019learning}
J.~Zhang, D.~Sun, Z.~Luo, A.~Yao, L.~Zhou, T.~Shen, Y.~Chen, L.~Quan, H.~Liao,
  Learning two-view correspondences and geometry using order-aware network, in:
  CVPR, 2019.

\bibitem{wang2023robust}
H.~Wang, Y.~Liu, Z.~Dong, Y.~Guo, Y.-S. Liu, W.~Wang, B.~Yang, Robust multiview
  point cloud registration with reliable pose graph initialization and history
  reweighting, in: CVPR, 2023.

\bibitem{yew2021learning}
Z.~J. Yew, G.~H. Lee, Learning iterative robust transformation synchronization,
  in: 3DV, 2021.

\bibitem{richter2016playing}
S.~R. Richter, V.~Vineet, S.~Roth, V.~Koltun, Playing for data: Ground truth
  from computer games, in: ECCV, Springer, 2016, pp. 102--118.

\bibitem{shafaei2016play}
A.~Shafaei, J.~J. Little, M.~Schmidt, Play and learn: Using video games to
  train computer vision models, arXiv preprint arXiv:1608.01745.

\bibitem{hu2021sail}
Y.-T. Hu, J.~Wang, R.~A. Yeh, A.~G. Schwing, Sail-vos 3d: A synthetic dataset
  and baselines for object detection and 3d mesh reconstruction from video
  data, in: CVPR, 2021, pp. 1418--1428.

\bibitem{handa2015scenenet}
A.~Handa, V.~Patraucean, V.~Badrinarayanan, S.~Stent, R.~Cipolla, Scenenet:
  understanding real world indoor scenes with synthetic data. arxiv preprint
  (2015), arXiv preprint arXiv:1511.07041.

\bibitem{noichl2021bim}
F.~Noichl, A.~Braun, A.~Borrmann, " bim-to-scan" for scan-to-bim: Generating
  realistic synthetic ground truth point clouds based on industrial 3d models,
  in: Proceedings of the 2021 European Conference on Computing in Construction,
  2021.

\bibitem{handa2012real}
A.~Handa, R.~A. Newcombe, A.~Angeli, A.~J. Davison, Real-time camera tracking:
  When is high frame-rate best?, in: ECCV, Springer, 2012, pp. 222--235.

\bibitem{handa2014benchmark}
A.~Handa, T.~Whelan, J.~McDonald, A.~J. Davison, A benchmark for rgb-d visual
  odometry, 3d reconstruction and slam, in: ICRA, IEEE, 2014, pp. 1524--1531.

\bibitem{mccormac2017scenenet}
J.~McCormac, A.~Handa, S.~Leutenegger, A.~J. Davison, Scenenet rgb-d: Can 5m
  synthetic images beat generic imagenet pre-training on indoor segmentation?,
  in: ICCV, 2017, pp. 2678--2687.

\bibitem{roberto2017procedural}
C.~Roberto~de Souza, A.~Gaidon, Y.~Cabon, A.~Manuel~Lopez, Procedural
  generation of videos to train deep action recognition networks, in: CVPR,
  2017, pp. 4757--4767.

\bibitem{kundu2020virtual}
A.~Kundu, X.~Yin, A.~Fathi, D.~Ross, B.~Brewington, T.~Funkhouser,
  C.~Pantofaru, Virtual multi-view fusion for 3d semantic segmentation, in:
  ECCV, Springer, 2020, pp. 518--535.

\bibitem{qiu2016unrealcv}
W.~Qiu, A.~Yuille, Unrealcv: Connecting computer vision to unreal engine, in:
  ECCV, Springer, 2016, pp. 909--916.

\bibitem{song2017semantic}
S.~Song, F.~Yu, A.~Zeng, A.~X. Chang, M.~Savva, T.~Funkhouser, Semantic scene
  completion from a single depth image, in: CVPR, 2017, pp. 1746--1754.

\bibitem{zhang2017}
Y.~Zhang, S.~Song, E.~Yumer, M.~Savva, J.-Y. Lee, H.~Jin, T.~Funkhouser,
  Physically-based rendering for indoor scene understanding using convolutional
  neural networks, arXiv preprint arXiv:1612.07429.

\bibitem{wang2020tartanair}
W.~Wang, D.~Zhu, X.~Wang, Y.~Hu, Y.~Qiu, C.~Wang, Y.~Hu, A.~Kapoor, S.~Scherer,
  Tartanair: A dataset to push the limits of visual slam, in: IROS, IEEE, 2020,
  pp. 4909--4916.

\bibitem{biswasa2015planning}
H.~K. Biswasa, F.~Bosch{\'e}a, M.~Suna, Planning for scanning using building
  information models: A novel approach with occlusion handling, in: Symposium
  on Automation and Robotics in Construction and Mining (ISARC 2015), Vol.~15,
  2015, p.~18.

\bibitem{diaz2018scan}
L.~D{\'\i}az-Vilari{\~n}o, E.~Fr{\'\i}as, J.~Balado, H.~Gonz{\'a}lez-Jorge,
  Scan planning and route optimization for control of execution of as-designed
  bim., International Archives of the Photogrammetry, Remote Sensing \& Spatial
  Information Sciences.

\bibitem{frias2019bim}
E.~Fr{\'\i}as, L.~D{\'\i}az-Vilari{\~n}o, J.~Balado, H.~Lorenzo, From bim to
  scan planning and optimization for construction control, Remote Sensing
  11~(17) (2019) 1963.

\bibitem{barron2013intrinsic}
J.~T. Barron, J.~Malik, Intrinsic scene properties from a single rgb-d image,
  in: CVPR, 2013, pp. 17--24.

\bibitem{gschwandtner2011blensor}
M.~Gschwandtner, R.~Kwitt, A.~Uhl, W.~Pree, Blensor: Blender sensor simulation
  toolbox, in: International Symposium on Visual Computing, Springer, 2011, pp.
  199--208.

\bibitem{matterport}
{Matterport}, \url{https://matterport.com/}, accessed: 2023-06-29.

\bibitem{cloudcompare}
{Cloud Compare} 3d point cloud and mesh processing software,
  \url{https://www.danielgm.net/cc/}, accessed: 2023-06-29.

\bibitem{icp}
P.~J. Besl, N.~D. McKay, Method for registration of 3-d shapes, in: Sensor
  fusion IV: control paradigms and data structures, Vol. 1611, International
  Society for Optics and Photonics, 1992, pp. 586--606.

\bibitem{primesense}
{Primesense sensor}, \url{http://xtionprolive.com/primesense-carmine-1.09},
  accessed: 2023-06-29.

\bibitem{blender}
{Blender.org - Home of the Blender Project}, \url{https://www.blender.org},
  accessed: 2023-06-29.

\bibitem{zeng20163dmatch}
A.~Zeng, S.~Song, M.~Nie{\ss}ner, M.~Fisher, J.~Xiao, T.~Funkhouser, {3DMatch}:
  learning local geometric descriptors from {RGB-D} reconstructions, in: CVPR,
  2017.

\bibitem{weinmann2013feature}
M.~Weinmann, B.~Jutzi, C.~Mallet, Feature relevance assessment for the semantic
  interpretation of 3d point cloud data, ISPRS Annals of the Photogrammetry,
  Remote Sensing and Spatial Information Sciences 2 (2013) 313--318.

\bibitem{bigun1987optimal}
J.~Bigun, Optimal orientation detection of linear symmetry, link{\"o}ping
  University Electronic Press (1987).

\end{thebibliography}
\end{document}